\documentclass[preprint,11pt]{elsarticle}

\usepackage{lmodern}

\usepackage{amssymb}
\usepackage{amsmath}

\usepackage{lineno}
\usepackage{xcolor}

\usepackage{hyperref}
\hypersetup{
    colorlinks=true,
    linkcolor=blue,
    filecolor=magenta,      
    urlcolor=cyan,
}
\renewcommand{\sectionautorefname}{\ifnum\spacefactor>1000 Section\else section\fi}
\usepackage{subfigure}
\usepackage{booktabs}
\usepackage{multirow}
\usepackage{glossaries}
\usepackage[nobiblatex]{xurl} 

\usepackage{array}
\usepackage{amsfonts}       
\newcolumntype{P}[1]{>{\raggedright\arraybackslash}p{#1}}

\usepackage{algorithm}
\usepackage{algpseudocode}
\usepackage{nicefrac}       

\usepackage[
  top     = 30mm, bottom     = 20mm,
  outer   = 25mm, inner      = 25mm,
  headsep = 10mm, headheight = 15pt,
  marginparwidth = 15mm ]{geometry}

\journal{ISPRS Journal of Photogrammetry and Remote Sensing}

\begin{document}

\begin{frontmatter}

\title{FLORO: A Multimodal Geospatial Foundation Model for Ecological Remote Sensing Across Sensors and Scales}

\author[l1]{Jorge L. Rodr\'iguez\corref{cor1}} 
\ead{jorge.rodriguezgalvis@kaust.edu.sa}
\author[l1]{Victor Angulo-Morales}
\author[l1]{Areej Alwahas}
\author[l1]{Mariana El\'ias-Lara}
\author[l2]{Fida Mohammad Thoker}
\author[l1]{Kasper Johansen}
\author[l2]{Bernard Ghanem}
\author[l1]{Fernando T. Maestre}
\author[l1]{Matthew F. McCabe}

\affiliation[l1]{organization={Biological and Environmental Science and Engineering Division,\\
  King Abdullah University of Science and Technology},
            city={Thuwal},
            postcode={23955}, 
            state={Makkah},
            country={Saudi Arabia}}
\affiliation[l2]{organization={Computer, Electrical and Mathematical Science and Engineering Division,\\
  King Abdullah University of Science and Technology},
            city={Thuwal},
            postcode={23955}, 
            state={Makkah},
            country={Saudi Arabia}}

\cortext[cor1]{Corresponding author}

\begin{abstract}
The proliferation of remote sensing platforms has generated an unprecedented diversity of Earth observation data, creating both the opportunity and the challenge of extracting transferable knowledge across sensors, scales, and environments. Foundation models have emerged as a promising framework for this challenge, leveraging large datasets to learn representations that generalise across tasks and sensing conditions in real-world Earth observation. However, many current approaches rely on very large pretraining datasets and remain constrained by fixed sensor configurations, limited observation-type flexibility, or narrow evaluation settings. These limitations are particularly relevant for ecological and environmental applications, where data are often heterogeneous across platforms,  resolutions, and data sources. We introduce FLORO, a multimodal geospatial foundation model designed to learn transferable representations from a comparatively small, heterogeneous remote-sensing pretraining dataset. FLORO is pretrained using masked autoencoding on a heterogeneous combination of Sentinel-1, Sentinel-2, SkySAT imagery, terrain data, and UAV-derived products. To accommodate real-world sensor variability, FLORO incorporates availability-aware inputs that indicate which spectral bands and auxiliary observations are present in each sample, enabling a unified input space across heterogeneous sensor configurations. We evaluated FLORO on the PANGAEA benchmark under a frozen-encoder protocol across scene classification, segmentation, and regression tasks. Despite being pretrained on a smaller dataset than competing foundation models, FLORO achieved strong and stable transfer across optical, optical-SAR, and optical-elevation benchmarks spanning medium-resolution satellite, airborne, and ultra-high-resolution UAV imagery. FLORO obtained the second-best average segmentation performance across six PANGAEA benchmarks, the third-best average rank across three scene-classification benchmarks, and excelled in regression tasks, while qualitative results showed improved preservation of spatial structure in flood, urban, biomass, and canopy-height prediction settings.  Overall, these results position FLORO as a practical framework for ecological and environmental monitoring under heterogeneous sensing conditions. To support benchmarking and reproducibility, the source code and PANGAEA evaluation integration, including access to the benchmark datasets, are available at \url{https://github.com/jorlrodriguezg/floro}, with pretrained model checkpoints at \url{https://huggingface.co/jorlrodriguezg/floro}.
\end{abstract}

\begin{graphicalabstract}
\includegraphics[width=\linewidth]{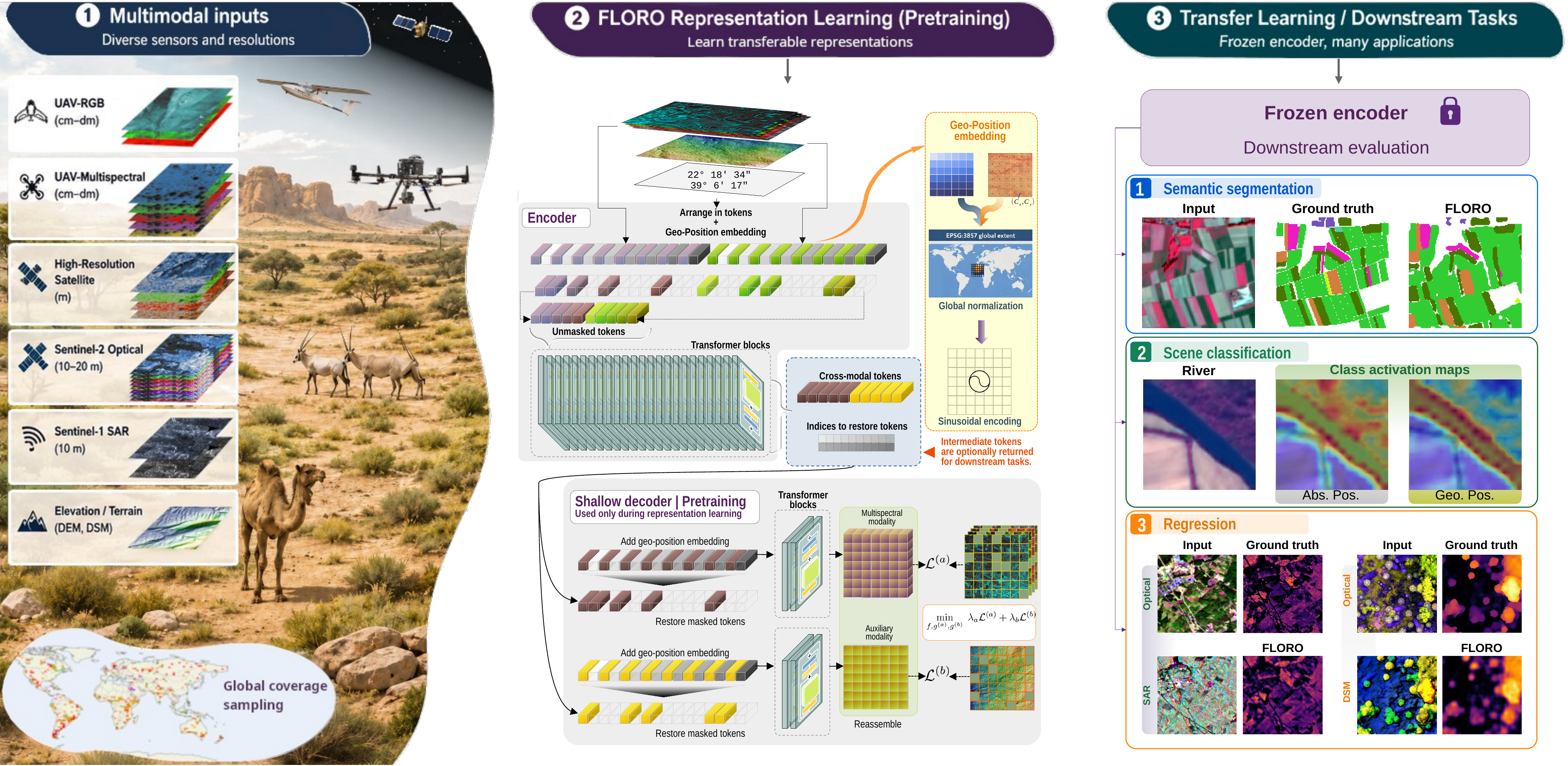}
\end{graphicalabstract}

\begin{highlights}
    \item FLORO learns transferable representations from a comparatively small, heterogeneous pretraining dataset.
    \item Availability-aware inputs support flexible transfer across optical, SAR, elevation, and UAV-derived data.
    \item FLORO achieves strong cross-task transfer across classification, segmentation, and regression under frozen-encoder evaluation.
\end{highlights}


\end{frontmatter}

\section*{Glossary of selected technical terms}
\addcontentsline{toc}{section}{Glossary of machine-learning terms}

\begingroup
\small
\setlength{\LTpre}{6pt}
\setlength{\LTpost}{12pt}
\renewcommand{\arraystretch}{1.0}
\begin{longtable}{>{\raggedright\arraybackslash}p{0.25\textwidth}>{\raggedright\arraybackslash}p{0.69\textwidth}}
\caption{Selected technical terms used in this study.}
\label{tab:ml_glossary}\\
\toprule
\textbf{Term} & \textbf{Definition in the context of this study} \\
\midrule
\endfirsthead
\multicolumn{2}{l}{\textit{Table \thetable{} continued}}\\
\toprule
\textbf{Term} & \textbf{Definition in the context of this study} \\
\midrule
\endhead
\midrule
\multicolumn{2}{r}{\textit{Continued on the next page}}\\
\endfoot
\bottomrule
\endlastfoot

Patch, patch token, and mask token & A patch is a fixed image region; its token is a vector representation. A mask token marks a removed patch for reconstruction. \\

Encoder and frozen encoder & The encoder produces reusable representations. A frozen encoder is not updated during downstream evaluation. \\

Decoder: pretraining vs. downstream & The pretraining decoder reconstructs masked inputs and is then discarded. A downstream decoder or prediction head produces task outputs. \\

Linear evaluation & Evaluation using a learned linear prediction layer on top of a frozen encoder. \\

Modality & A distinct observation type, such as optical reflectance, SAR backscatter, elevation or text. \\

Masking curriculum & Training schedule that progressively increases the proportion of hidden patch tokens. \\

Dense annotation and dense prediction & Dense means per-pixel targets and outputs: classes in semantic segmentation or continuous values in regression. Scene classification instead gives one label per tile; object detection gives sparse localized outputs. \\
\end{longtable}
\endgroup


\section{Introduction}
Recent advances in artificial intelligence, particularly in deep learning, have created new opportunities to address long-standing limitations of ecological monitoring \citep{Besson2022TowardsCommunities, Borowiec2022DeepEvolution}, including the restricted spatial extent of field campaigns, the labor required for manual data collection, and the difficulty of scaling observations across heterogeneous landscapes \citep{Cavender-Bares2022IntegratingConservation}. In remote sensing, deep neural networks have demonstrated strong capacity to learn complex relationships between imagery and ecological variables from large volumes of data, enabling more efficient and spatially extensive monitoring workflows \citep{Ma2019DeepReview, Wu2023DeepLandscape}. These developments are especially relevant in ecology, where the variables of interest often emerge from interactions amongst vegetation structure, spectral response, topography, and environmental context.

The rapid increase of remote sensing platforms \citep{McCabe2017TheHydrology} has led to an increasingly diverse landscape of Earth observation data, spanning multiple sensors, spatial resolutions, and acquisition conditions \citep{Zhu2026OnModels}. Such heterogeneity poses a fundamental challenge for traditional approaches, which often rely on task-specific models trained on limited and narrowly distributed datasets \citep{Cong2023SatMAE:Imagery}. Foundation models have recently emerged as a promising paradigm for remote sensing, in part because they aim to learn transferable representations from large collections of unlabeled data that can later be adapted to a broad range of downstream tasks, including classification, segmentation, and regression \citep{Bommasani2021OnModels, Mall2024REMOTEALIGNMENT,Wang2024MTP:Pretraining,Cha2024AImages}. Their potential is particularly attractive in ecology, where labeled datasets are typically scarce, expensive to produce, and geographically limited \citep{Morera2024FoundationEcology}. However, despite this promise, current remote sensing foundation models still face important limitations. In many cases, gains in representation quality do not consistently translate into robust downstream performance across diverse sensors, resolutions, and task types. Prior work has also noted that geospatial learning can benefit from explicitly leveraging geographic context and from aligning complementary observation types through their spatial relationships \citep{Mai2024OnPaper,Wang2024MTP:Pretraining}. These observations suggest that progress in geospatial foundation modeling depends not only on scale, but also on how effectively a model incorporates multimodal and spatially grounded information. In this study, a modality denotes a distinct type of remote sensing observation, such as optical reflectance, SAR backscatter, or elevation.

Multimodal learning is particularly relevant in remote sensing, where different sensors offer the capacity for complementary descriptions of the Earth surface \citep{McCabe2008HydrologicalStudies,Gao2025SFA-Net:Matching}. Optical and multispectral imagery capture reflectance patterns related to vegetation condition, material composition, and land cover, whereas elevation products provide structural and topographic information, and Synthetic Aperture Radar (SAR) contributes sensitivity to surface roughness, moisture, and imaging conditions less affected by cloud cover. A large body of prior work has shown that combining such sources can improve performance across land-use mapping, environmental monitoring, agricultural assessment, hydrological consistency, and other Earth observation tasks \citep{Benediktsson2007MultipleDevelopments,DallaMura2015ChallengesSensing,Lopez2017EvaluatingData,Sturari2017IntegratingMapping,Wang2023DecouplingLearning}. The integration of spectral and topographic information has been shown to enhance classification in heterogeneous environments \citep{Harston1992IMPROVEDDATA,Sturari2017IntegratingMapping}, while broader forms of multimodal fusion have improved land-cover discrimination, environmental characterization, and predictive mapping \citep{Khanal2018IntegrationYield,Brell20193DExtraction}. However, in practical remote sensing workflows, multimodal learning must also account for the fact that not all sensors provide the same spectral bands or spatially compatible auxiliary data, making flexibility to partial and heterogeneous inputs a central requirement for broadly transferable foundation models \citep{Li2022DeepReview,Hong2026FoundationMultimodality}. For ecological applications, where subtle differences in vegetation structure and condition may be difficult to detect from a single sensing modality, such complementarity is especially valuable.

Ecological monitoring rarely relies on a single sensor, spatial resolution, or acquisition setting. Observations may include medium-resolution satellite imagery, high-resolution commercial imagery, elevation models, and products derived from unmanned aerial vehicles (UAV) \citep{Cavender-Bares2020ApplyingScience,vanderLeeuw2024UsingModels}, often with only limited reference labels. A desirable property of remote sensing foundation models should therefore be the ability to transfer across sensors, spatial resolutions, and acquisition settings without requiring large-scale retraining for each new domain \citep{Mai2024OnPaper,Lu2024AISurvey,Jakubik2023FoundationIntelligence}. Such a need is particularly acute in ecological monitoring, where data may come from medium-resolution satellite platforms, high-resolution commercial imagery, elevation models, or UAV derived products, and often with limited accompanying annotation. In such settings, the central challenge is not merely to pretrain a large encoder, but to learn representations that remain useful across heterogeneous data regimes. Models that can exploit unlabeled multimodal data and transfer effectively to downstream tasks under a common evaluation protocol are therefore of particular interest.

In this study, we investigate whether an availability-aware, large-capacity encoder can learn competitive transferable representations from a comparatively small pretraining dataset and apply them across heterogeneous downstream tasks. We introduce FLORO, a multimodal geospatial foundation model trained on remote sensing data sources that differ in sensor, platform, spatial resolution, radiometry, and available inputs. In contrast to recent foundation models that rely on extremely large pretraining datasets \citep{Xiong2025NeuralObservation,Jakubik2025TerraMind:Observation}, FLORO is trained on a comparatively smaller but highly heterogeneous cross-platform dataset composed of Sentinel-1 GRD SAR \citep{EuropeanSpaceAgency2026Sentinel-1Documentation}, Sentinel-2 L2A surface reflectance imagery \citep{EuropeanSpaceAgency2021CopernicusProduct}, SkySAT imagery processed to surface reflectance \citep{Collison2025PlanetPaper} and paired with high-resolution digital terrain models \citep{IGN2026RGETerrain}, radiometrically calibrated UAV multispectral reflectance orthomosaics and uncalibrated UAV RGB orthomosaics, and UAV-derived structural photogrammetric products, including digital surface models. The design of FLORO introduces flexibility in both spectral composition and modality combination, allowing the encoder to process inputs from heterogeneous sensors without requiring a fixed band configuration. Such flexibility is particularly relevant in Earth observation, where datasets differ widely in their spectral definitions and where some modalities are not spatially compatible across acquisition platforms. By explicitly encoding modality and band availability, FLORO can process heterogeneous training and downstream data within a unified framework while remaining adaptable to diverse downstream sensing scenarios. The resulting architecture combines availability-aware input modeling, cross-scale multimodal pretraining, and geographic grounding through geo-positional encoding, as described in \autoref{sec:floro_architecture}.

Specifically, this work makes four contributions. First, we introduce a unified multimodal pretraining framework that can ingest heterogeneous optical, SAR, and elevation inputs while explicitly modeling modality availability (\autoref{sec:floro_architecture}). Second, we assemble a comparatively small cross-platform pretraining dataset spanning Sentinel-1, Sentinel-2, high-resolution satellite imagery, elevation products, and UAV-derived photogrammetric observations, with variability in spatial resolution, sensor characteristics, and modality availability (\autoref{sec:experiments} and \ref{app:ap1}). Third, we evaluate FLORO under a standardized frozen-encoder benchmark across segmentation, scene classification, and regression tasks to assess transfer under heterogeneous sensing regimes (\autoref{sec:experiments}). Fourth, we use controlled ablations to examine pretraining-dataset composition, the masking curriculum, availability and validity channels, geo-positional encoding, and selected auxiliary modalities  (\autoref{sec:ablations} and \ref{app:ablations}).

The limitations identified in prior work on remote-sensing foundation models and multimodal learning (\autoref{sec:related_work}), together with the masked-reconstruction principles summarized in \autoref{sec:mae_background}, motivate FLORO's availability-aware architecture and pretraining strategy (\autoref{sec:floro_architecture}). We test the resulting representations through the pretraining dataset, downstream datasets, standardized evaluation protocols, and experimental analyses presented in \autoref{sec:experiments}. The Discussion interprets these findings in relation to multimodal transfer, geographic encoding, pretraining-data diversity, and the limitations of the current framework (\autoref{sec:discussion}), while the Conclusion considers their broader implications for ecological and heterogeneous Earth observation applications (\autoref{sec:conclusion}).

\section{Related Work}
\label{sec:related_work}

This section reviews the literature that motivates FLORO and, for each line of work, identifies the limitation that FLORO is designed to address. We first summarise self-supervised and masked pretraining in computer vision and Earth observation, then examine multimodal and cross-sensor fusion and the emerging class of geospatial foundation models. We then discuss geographic grounding and spatial context, and close by stating how FLORO differs from these bodies of work.

\subsection{Self-supervised and masked pretraining in vision}

Masked autoencoding (MAE) reconstructs hidden patches from a subset of visible tokens and has become a reference paradigm for self-supervised representation learning in vision \citep{He2022MaskedLearners}. By encouraging the encoder to capture contextual and spatial relationships without requiring labeled data, such a label-free pretraining objective is particularly attractive for remote sensing, where large volumes of imagery are available but task-specific reference data are costly to obtain. Such reference data include dense annotations, which assign a class or continuous value to each valid pixel, and scene-level labels, which assign a single class to an entire image tile. The MAE paradigm builds on the transformer architecture and on vision transformers that apply it directly to image patches \citep{Vaswani2017AttentionNeed,Dosovitskiy2020AnScale}, and it has been generalised to other modalities, including audio \citep{Huang2023MaskedListen} and video \citep{Vandeghen2026TrackMAE:Predict}. Curriculum-based variants that progressively increase the masking ratio during training have also reported improved transfer relative to a fixed ratio \citep{Madan2024CL-MAE:Autoencoders,Yoon2025CurriMAE:Classification}, which informs the masking schedule adopted in FLORO.

{An important distinction arises, however, when conventional masked pretraining is applied to heterogeneous Earth observation data. The canonical MAE formulation assumes a predefined input configuration in which masking represents the deliberate removal of otherwise available information. In heterogeneous Earth observation, entire spectral groups or auxiliary modalities may instead be genuinely unavailable because different sensors and acquisition platforms provide different observations. Distinguishing deliberately masked information from genuinely absent observations therefore becomes important when extending masked pretraining across sensors and acquisition configurations.

\subsection{Multimodal masked pretraining and Earth observation}

Several works have extended masked autoencoding to multimodal settings. MultiMAE \citep{Bachmann2022MultiMAE:Target}, for example, jointly models multiple input modalities by encoding modality-specific tokens within a shared transformer and reconstructing them through separate decoders. MultiMAE enables cross-modal learning by leveraging complementary information across modalities. However, such approaches often assume relatively consistent modality availability and homogeneous data distributions, whereas remote sensing data are frequently incomplete, sensor-dependent, and acquired across substantially different spatial resolutions and observation geometries.

In Earth observation, this line of work has been extended through spectral, temporal, and scale-aware masked autoencoders. SatMAE \citep{Cong2023SatMAE:Imagery} introduces temporal and multi-spectral positional embeddings so that a masked autoencoder can exploit multi-date, multi-band imagery, and Scale-MAE \citep{Reed2022Scale-MAE:Learning} treats scale as an explicit factor by reconstructing imagery across resolutions. CROMA \citep{Fuller2023CROMA:Autoencoders} combines contrastive and masked objectives to align optical and synthetic aperture radar (SAR) representations, whereas SpectralGPT \citep{Hong2024SpectralGPT:Model} adapts generative pretraining to spectral data. Curated datasets such as SSL4EO-S12 \citep{Wang2023SSL4EO-S12:Observation} support multi-modal, multi-temporal self-supervised learning, SatlasPretrain \citep{Bastani2023SatlasPretrain:Understanding} demonstrates the value of large supervised pretraining across many geospatial tasks, and lightweight pretraining has been shown to be practical for satellite time series \citep{Garnot2020LightweightTimeseries,Tseng2024LightweightTimeseries,SainteFareGarnot2022Multi-modalSeries}.

These approaches establish the effectiveness of self-supervised learning for Earth observation, but address different dimensions of observational heterogeneity. SatMAE primarily models spectral and temporal structure within predefined observation configurations, Scale-MAE explicitly addresses spatial resolution, CROMA exploits complementary optical and SAR information, and SpectralGPT emphasizes spectral representation learning. Although these capabilities are complementary, their evaluation provides more limited evidence about how a shared encoder behaves when sensor characteristics, spectral availability, auxiliary modalities, and spatial resolution vary jointly across datasets. This combined setting, rather than any one of these sources of variability independently, is the regime targeted by FLORO.

\subsection{Multimodal fusion and cross-sensor learning in remote sensing}

Prior work in remote sensing has long demonstrated the value of combining spectral imagery with auxiliary geospatial information. The integration of multispectral data with elevation-derived variables such as slope and aspect has been shown to improve land use and land cover classification in heterogeneous environments \citep{Harston1992IMPROVEDDATA,Sturari2017IntegratingMapping}, while broader multimodal fusion strategies have improved classification robustness, environmental characterization, and predictive mapping \citep{Benediktsson2007MultipleDevelopments,Khanal2018IntegrationYield,DallaMura2015ChallengesSensing,Wang2023DecouplingLearning,Hong2026FoundationMultimodality}. Nevertheless, many of these approaches are designed for specific tasks, study regions, or fixed combinations of input data. As a result, they do not fully address the problem of learning a single transferable representation that remains usable when spectral bands, auxiliary modalities, or spatial resolutions vary across datasets \citep{Gao2025SFA-Net:Matching}.

Related fusion studies report consistent gains from combining sensors: fusing laser scanning with hyperspectral data improves object-based information extraction \citep{Brell20193DExtraction}, and multi-sensor fusion has been shown to benefit environmental characterisation, soil and yield prediction, and hydrological consistency \citep{Lopez2017EvaluatingData,McCabe2008HydrologicalStudies,McCabe2017TheHydrology}. Learned fusion has moved from stacked-feature designs towards objectives that separate shared from modality-specific information and towards attention-based matching across sensors, and comprehensive reviews repeatedly identify the handling of heterogeneous, incomplete, or partially incompatible modalities as an open problem \citep{Li2022DeepReview,Samadzadegan2025AProspects}.

Recent developments have begun addressing these limitations more explicitly. For example, MetaRS \citep{Zhou2026RemoteMethod} introduces meta-modal representation learning to separate shared and modality-specific information for land cover mapping under missing-modality conditions. Such approaches demonstrate that incomplete observations can be accommodated through learned representations rather than requiring all modalities to be present. Nevertheless, many fusion pipelines remain developed for specific applications and predefined combinations of co-registered modalities, and accommodating variable observations across heterogeneous sensors and acquisition platforms remains challenging. Task-specific fusion also does not necessarily produce representations that transfer to unrelated datasets or prediction objectives. FLORO instead represents modality availability explicitly, so that an absent modality is treated as unobserved rather than as a reconstruction target or a zero-valued band.

\subsection{Geospatial foundation models and geographic context}

Recent remote sensing foundation models seek to overcome these limitations by pretraining transferable encoders on large collections of unlabeled Earth observation data \citep{Jakubik2023FoundationIntelligence,Mai2024OnPaper,Lu2024AISurvey}. These models have advanced representation learning for downstream tasks such as classification, segmentation, and regression, but many remain tied to specific sensor configurations, fixed spectral inputs, or large-scale pretraining datasets. In parallel, recent work has highlighted the importance of geographic context and spatial relationships for geospatial learning \citep{Mai2024OnPaper,Wang2024MTP:Pretraining}, suggesting that transferability depends not only on pretraining scale but also on how models encode modality availability, spatial grounding, and cross-sensor variability.

Recent developments have substantially expanded this landscape. Prithvi established large-scale geospatial pretraining for transferable Earth observation representations, while subsequent models have broadened the range of sensors and modalities that can be accommodated. TerraMind \citep{Jakubik2025TerraMind:Observation} extends generative pretraining across multiple Earth observation modalities and supports cross-modal learning and generation, neural-plasticity-inspired models such as DOFA \citep{Xiong2025NeuralObservation} accommodate varying spectral wavelengths, multi-task and continual pretraining \citep{Mendieta2023TowardsPretraining} broaden the set of transferable tasks, and vision-language pretraining such as RemoteCLIP \citep{Liu2024RemoteCLIP:Sensing} couples imagery with text.

Other recent architectures explicitly target sensor and resolution heterogeneity. AnySat \citep{Astruc2025AnySat:Modalities} employs a joint-embedding predictive architecture with scale-adaptive encoders to learn shared representations from heterogeneous datasets spanning multiple sensors, resolutions, and modalities. SkySense++ \citep{Wu2025AObservation} introduces a factorized multimodal architecture and progressive pretraining to support transfer across diverse Earth observation tasks. More specifically addressing spectral flexibility, Any-Optical-Model \citep{Li2026Any-Optical-Model:Sensing} combines spectrum-independent tokenization, adaptive patch embeddings, and channel-wise masked reconstruction to accommodate variable optical band configurations and spatial resolutions. Similarly, SpecAware \citep{Ji2026SpecAware:Mapping} incorporates spectral content and sensor characteristics to support cross-sensor representation learning in hyperspectral imagery.

Together, these models demonstrate substantial progress towards flexible geospatial representations, although they emphasize different dimensions of observational heterogeneity. AnySat directly addresses cross-sensor, cross-modality, and cross-resolution learning, while SkySense++ emphasizes multimodal representation learning and semantic transfer. DOFA, Any-Optical-Model, and SpecAware focus more specifically on accommodating spectral variability through different architectural mechanisms. These developments overlap with important aspects of FLORO and establish that sensor flexibility is not, by itself, a unique contribution. However, differences in pretraining objectives, modality-availability handling, and downstream evaluation leave room for complementary approaches to learning transferable representations from heterogeneous observations.

Standardised benchmarking has begun to keep pace through frameworks such as PANGAEA \citep{Marsocci2026PANGAEA:Benchmark}, recent reviews articulate the opportunities and open challenges of foundation models for GeoAI and for ecological applications \citep{Zhu2026OnModels,Morera2024FoundationEcology}, and parameter-efficient adaptation can reduce downstream cost \citep{Marti-Escofet2025Fine-tuneModels}.

Recent controlled comparisons of flexible geospatial foundation-model architectures further highlight the trade-offs between spectral flexibility, modality alignment, and downstream task performance \citep{Dias2026EmergingModels}. Such studies reinforce the importance of standardized evaluation protocols, since architectural flexibility does not necessarily translate into uniform improvements across downstream tasks.

Pretraining scale represents a related consideration. Many of the strongest systems \citep{Cha2024AImages} rely on very large pretraining corpora, which can improve representation quality but also require substantial computational and data-processing resources. This motivates the complementary question of how much transfer can be obtained from a comparatively small but deliberately heterogeneous, cross-platform pretraining dataset, as examined here.

Dedicated location encoders such as Sphere2Vec \citep{Mai2023Sphere2Vec:Predictions} and spherical-harmonic representations \citep{Ruwurm2024GeographicNetworks} learn smooth functions of coordinates that improve large-scale geospatial prediction, but they are developed largely for point and location-regression tasks, and their integration with masked multimodal reconstruction is less explored. Because coordinates can also encode regional priors, apparent gains may reflect spatial proximity between pretraining and evaluation data rather than transferable representations; we therefore treat geographic grounding as a design choice to be tested rather than assumed, and examine it in a controlled ablation while auditing geographic overlap between pretraining and evaluation (see \ref{app:geographic_overlap}).

\subsection{Positioning of FLORO}

FLORO builds upon masked autoencoding and multimodal transformer architectures, but introduces several modifications tailored to heterogeneous geospatial data. In contrast to approaches that assume fixed input configurations, FLORO explicitly models modality availability, supports optical, SAR, and elevation inputs across different sensing conditions, spatial scales, and acquisition platforms. These design choices allow one encoder to be pretrained and evaluated across variable input configurations without requiring every source to provide the same bands or auxiliary modalities.

Recent developments demonstrate substantial progress towards flexible geospatial representations, including cross-sensor and cross-resolution learning, multimodal integration, and explicit handling of missing observations. However, important differences remain in how these capabilities are combined, the pretraining objectives used, and the downstream tasks through which transferability is evaluated. FLORO investigates a complementary approach based on availability-aware masked autoencoding, integrating optical, SAR, and elevation observations across heterogeneous satellite and UAV datasets. Its contribution lies not in introducing sensor or modality flexibility independently, but in combining explicit availability representation, cross-platform pretraining, and systematic evaluation across classification, segmentation, and continuous biophysical regression tasks within a unified framework.

Relative to the masked-pretraining literature, FLORO reconstructs deliberately masked tokens while marking genuinely missing channels and invalid pixels and excluding them from the loss, so that a single objective applies to partial inputs. Relative to multimodal fusion, it removes the requirement that modalities be co-observed by treating availability as part of the input. Relative to large geospatial foundation models, it examines how much transfer can be obtained from a comparatively small but deliberately heterogeneous, cross-platform pretraining dataset. Relative to geographic-grounding work, it evaluates a geo-positional branch through a controlled ablation instead of assuming a uniform benefit.

The contribution of FLORO therefore lies in the combined treatment of heterogeneous observations and in the systematic evaluation of the resulting representations. These distinctions are made concrete by the architecture in \autoref{sec:floro_architecture} and tested under a standardised frozen-encoder protocol in \autoref{sec:experiments}, spanning semantic segmentation, scene classification, and continuous biophysical regression, including tasks relevant to ecological monitoring.

\section{Masked reconstruction for self-supervised pretraining}
\label{sec:mae_background}

In a vision transformer, the image is first divided into fixed-size patches. Each patch is projected into a token, which is a learned vector representation of the corresponding spatial region. Masked autoencoding (MAE) therefore is a self-supervised learning strategy in which a model learns representations by reconstructing hidden image patches \citep{He2022MaskedLearners}. Let $\mathbf{X}\in\mathbb{R}^{H\times W\times C}$ denote an image tile with height $H$, width $W$, and $C$ channels. Dividing the tile into non-overlapping patches of size $P\times P$ produces $N=(H/P)(W/P)$ flattened patch vectors $\mathbf{x}_i\in\mathbb{R}^{D}$, where $D=P^2C$ and $i\in\{1,\ldots,N\}$. Each patch is projected into a $d$-dimensional token and combined with a positional encoding:

\begin{equation}
\mathbf{e}_i = \mathbf{W}_{p}\mathbf{x}_i+\mathbf{p}_i,
\label{eq:mae_patch_embedding}
\end{equation}

where $\mathbf{W}_{p}\in\mathbb{R}^{d\times D}$ is the patch-projection matrix, $\mathbf{p}_i\in\mathbb{R}^{d}$ is the positional encoding, and $\mathbf{e}_i\in\mathbb{R}^{d}$ is the resulting token.

A random mask partitions the patch indices into disjoint visible and masked sets,

$\mathcal{V}$ and $\mathcal{M}$, whose union is
$\{1,\ldots,N\}$. The encoder processes only the visible tokens:
\begin{equation}
\mathbf{Z}_{\mathcal{V}}
=
f_{\theta}
\left(
\left(\mathbf{e}_i\right)_{i\in\mathcal{V}}
\right),
\label{eq:mae_encoder}
\end{equation}

where $f_{\theta}$ is the encoder with trainable parameters $\theta$, and
$\mathbf{Z}_{\mathcal{V}}$ contains the latent representations of the visible patches.

A lightweight decoder predicts the masked patches from the visible representations and learned mask tokens:

\begin{equation}
\left(
\hat{\mathbf{x}}_i
\right)_{i\in\mathcal{M}}
=
g_{\phi}
\left(
\mathbf{Z}_{\mathcal{V}},
\mathbf{m}_{\mathcal{M}}
\right),
\label{eq:mae_decoder}
\end{equation}

where $g_{\phi}$ is the decoder with trainable parameters $\phi$, $\mathbf{m}_{\mathcal{M}}$ denotes the mask tokens and their positional encodings, and $\hat{\mathbf{x}}_i\in\mathbb{R}^{D}$ is the reconstruction of patch $i$. Training minimizes the mean squared reconstruction error over the masked patches:

\begin{equation}
\mathcal{L}_{\mathrm{MAE}}
=
\frac{1}{|\mathcal{M}|D}
\sum_{i\in\mathcal{M}}
\left\|
\hat{\mathbf{x}}_i-\mathbf{x}_i
\right\|_2^2,
\label{eq:mae_reconstruction_loss}
\end{equation}

where $|\mathcal{M}|$ is the number of masked patches and
$\|\cdot\|_2$ is the Euclidean norm.

By reconstructing only masked content, the encoder must capture contextual information from the visible patches. After pretraining, the decoder is typically discarded and the encoder is transferred to downstream tasks. This formulation provides the background for the masked-reconstruction approach described in \autoref{sec:floro_architecture}.

\section{FLORO Architecture}
\label{sec:floro_architecture}

Building upon the general masked autoencoding framework described in \autoref{sec:mae_background}, FLORO extends MAEs to heterogeneous remote sensing environments where inputs may differ substantially in spectral composition, spatial resolution, modality availability, and geospatial metadata. While standard MAEs are primarily designed for fixed RGB image inputs, FLORO introduces a unified multimodal formulation capable of jointly processing multispectral imagery, SAR observations, elevation products, and heterogeneous optical sensor configurations within a single transformer encoder. The overall architecture is illustrated in \autoref{fig_1}.

\begin{figure}[p]
\centering
\includegraphics[width=0.9\textwidth]{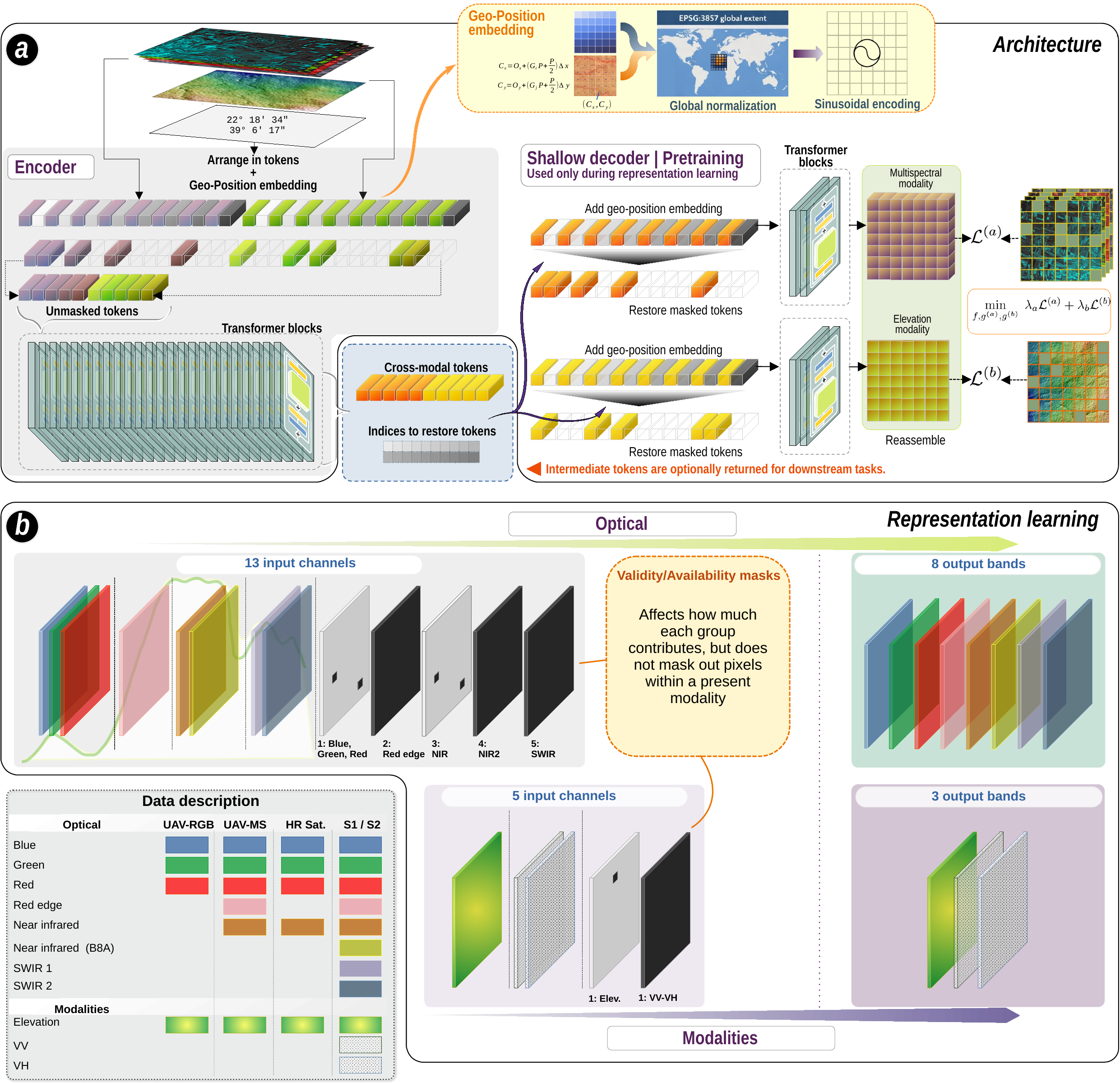}
\caption{Overview of the FLORO architecture for multimodal representation learning and efficient downstream adaptation. \textbf{(a)} During pretraining, heterogeneous remote sensing inputs are converted into patch tokens, augmented with positional information, randomly masked, and processed by a shared Vision Transformer encoder. Lightweight modality-aware decoders reconstruct masked optical and auxiliary-modality content under a self-supervised objective. The pretrained encoder is retained for downstream tasks, whereas the decoder is discarded after pretraining. \textbf{(b)} FLORO introduces a unified multimodal input space combining optical channels, auxiliary modalities, validity masks, and modality-availability indicators. Such a formulation enables the same encoder to process heterogeneous inputs from UAV RGB imagery, UAV multispectral imagery, high-resolution satellite imagery, and Sentinel-1/Sentinel-2 observations. The geo-positional branch further augments patch tokens with projected geographic coordinate embeddings when geospatial metadata are available.}
\label{fig_1}
\end{figure}

FLORO adapts the standard MAE formulation to a common remote-sensing setting in which datasets provide different combinations of optical reflectance, SAR backscatter, elevation, and valid spatial coverage. Five design choices address this variability. First, a common input organization allows the same encoder to process different combinations of optical bands and auxiliary data rather than requiring a separate model for each sensor configuration. Second, sample-level availability indicators and pixel-level validity channels distinguish absent input groups and unusable pixels from physically meaningful measurements close to zero. Third, the positional encoding combines the location of each patch within an image tile with its projected Earth coordinates when georeferencing information is available. Fourth, optical and auxiliary patches are hidden independently during pretraining, exposing the model to different combinations of visible information. Finally, separate reconstruction branches and losses account for the different physical quantities represented by optical reflectance, SAR backscatter, and elevation, while excluding unavailable or invalid targets. Together, these choices are intended to support transfer across the heterogeneous sensors and data configurations commonly encountered in ecological and natural-resource monitoring.

\subsection{Multimodal Input Representation}

FLORO generalizes the standard MAE input formulation to heterogeneous multimodal remote sensing inputs composed of optical spectral groups and auxiliary geospatial modalities. The optical branch may contain visible bands, red-edge channels, near-infrared observations, and shortwave infrared bands depending on sensor availability, while the auxiliary branch may include elevation products and SAR backscatter measurements.

To accommodate heterogeneous sensor configurations, FLORO introduces explicit validity and availability channels indicating which spectral groups and modalities are present in each input sample. Availability and validity indicators serve distinct but complementary purposes. Availability indicators identify which spectral or auxiliary input groups are present in an image tile, whereas validity indicators identify which pixels contain usable measurements. The latter allow the encoder to distinguish a physically meaningful value close to zero from a pixel set to zero because its original observation was missing or invalid. They also support consistent preprocessing across data sources that encode no-data pixels using different values, such as $-3.4028235e+38$, $-65{,}537$ or $-999$. Source-specific no-data values are identified and excluded before normalization; invalid pixels are subsequently set to zero, assigned a validity value of zero, and excluded from the masked-reconstruction loss. Without these operations, extreme values could distort the normalization and generate spurious reconstruction gradients, encouraging the model to learn source-specific no-data patterns rather than valid physical structure. This distinction is also important for dense regression tasks, such as biomass and canopy-height mapping, in which invalid regions can otherwise influence many pixel-level predictions.

Formally, each modality input is represented as

\begin{equation}
\mathbf{X}^{(s)}
\in
\mathbb{R}^{B\times C_s\times H\times W},
\qquad
s\in\{\mathrm{opt},\mathrm{aux}\},
\label{eq:floro_stream_input}
\end{equation}

where $s$ identifies the optical ($\mathrm{opt}$) or auxiliary ($\mathrm{aux}$) stream, $B$ is the batch size, $C_s$ is the number of measurement and indicator channels assigned to stream $s$, and $H$ and $W$ are the tile height and width. The packed channel layout is fixed within each stream, while the indicators encode which measurements are available for each sample.

Each modality stream is independently divided into non-overlapping patches of size $P\times P$, with $P=16$. Patch embedding follows the standard MAE formulation using a convolutional projection layer with kernel size and stride equal to $P$. Following \autoref{eq:mae_patch_embedding}, the stream-specific token sequence is

\begin{equation}
\mathbf{T}^{(s)}
=
E_s\!\left(\mathbf{X}^{(s)}\right)
\in
\mathbb{R}^{B\times N\times d},
\qquad
N=
\left(\frac{H}{P}\right)
\left(\frac{W}{P}\right),
\label{eq:floro_patch_embedding}
\end{equation}

where $E_s$ is the patch-embedding layer for stream $s$; $N$ is the number of patches; $d$ is the encoder token dimension; and $\mathbf{T}^{(s)}$ is the resulting token sequence. The optical and auxiliary inputs are projected by separate patch-embedding layers because they have different channel layouts. Measurement and validity channels are embedded jointly, so each patch token contains both the observed values and information about which groups are available. When auxiliary observations are unavailable, only the optical tokens are processed.

This grouping does not make the spectral response functions of different sensors identical. Instead, it provides a common coarse organization for partially overlapping band sets. The current experiments do not independently quantify modality-specific bias in the shared latent representation, robustness to controlled spatial misregistration or resolution changes, or stability across a systematic range of missing-modality patterns. These questions require targeted perturbation studies and representation-level diagnostics and are left for future research.

\subsection{Hybrid Positional Encoding}

Conventional MAEs typically employ fixed 2D sine-cosine positional embeddings encoding only image-space patch locations. Because ecological and land-surface variables are often geographically structured, explicit location representations can complement image content in large-scale geospatial and remote-sensing prediction \citep{Mai2023Sphere2Vec:Predictions,Ruwurm2024GeographicNetworks,Tseng2024LightweightTimeseries}. FLORO therefore combines local image-grid positions with projected Earth coordinates derived from georeferencing metadata, allowing patches at the same image-grid position but in different geographic regions to receive different location information. The resulting encoding provides approximate geographic context and is not intended to represent precise geodetic distances.

Each georeferenced sample is associated with a geotransform tensor (i.e., the information that maps image pixels to real-world geographic coordinates), denoted by $\mathbf{GT}\in\mathbb{R}^{B\times6}$ for a batch of $B$ image tiles. For each tile, the six coefficients comprise the projected origin $(O_x,O_y)$, horizontal and vertical pixel spacings $(D_x,D_y)$, and two rotation terms. Using the patch grid indices $G_i$ and $G_j$, the projected center coordinates of each patch are computed as

\begin{equation}
\label{eq:centroid_x}
C_x = O_x + \left(G_i \cdot P + \frac{P}{2}\right)\cdot D_x
\end{equation}

\begin{equation}
\label{eq:centroid_y}
C_y = O_y + \left(G_j \cdot P + \frac{P}{2}\right)\cdot D_y.
\end{equation}

Here, $O_x$ and $O_y$ are the projected coordinates of the image origin; $D_x$ and $D_y$ are the horizontal and vertical pixel spacings; $G_i$ and $G_j$ index the patch column and row; $P$ is the patch width in pixels; and $C_x$ and $C_y$ are the projected coordinates of the patch center. The rotation terms are zero for the north-up rasters used here and therefore do not appear in \autoref{eq:centroid_x} and \autoref{eq:centroid_y}.

Web Mercator (EPSG:3857) was selected as a single global projected coordinate frame so that all pretraining samples could be normalized with fixed bounds without discontinuities at UTM-zone boundaries. This choice supports consistent encoding across a geographically distributed dataset, but Web Mercator is not geodetically uniform and increasingly distorts distance and area toward high latitudes. The coordinates are therefore used only as approximate location features rather than as metric distances. The coordinates are normalized using \autoref{eq:normcentroid_x} and \autoref{eq:normcentroid_y}.

\begin{equation}
\label{eq:normcentroid_x}
N_x =
\frac{C_x-\text{min}_x}
{\text{max}_x-\text{min}_x}
\end{equation}

\begin{equation}
\label{eq:normcentroid_y}
N_y =
\frac{C_y-\text{min}_y}
{\text{max}_y-\text{min}_y}.
\end{equation}

where constants $\mathrm{min}_x = -20037508.34$, $\mathrm{max}_x = 20037508.34$, $\mathrm{min}_y = -20048966.10$, and $\mathrm{max}_y = 20048966.10$ are the fixed global Web Mercator coordinate bounds, and $N_x$ and $N_y$ are the corresponding normalized patch-center coordinates.

Following \citet{Vaswani2017AttentionNeed}, sinusoidal embeddings are applied to the normalized coordinates. The sinusoidal mapping itself is not introduced as a new positional-encoding function. Rather, it provides a deterministic multiscale feature map that lifts the two normalized geographic coordinates into the $d$-dimensional token space required by the transformer.

\begin{equation}
\label{eq:encoding_alternates}
E_{b\ell k}^{\mathrm{geo}} =
\begin{cases}
\sin\!\left(N_{x,b\ell}\omega_k\right), & k \bmod 4 = 0, \\
\cos\!\left(N_{x,b\ell}\omega_k\right), & k \bmod 4 = 1, \\
\sin\!\left(N_{y,b\ell}\omega_k\right), & k \bmod 4 = 2, \\
\cos\!\left(N_{y,b\ell}\omega_k\right), & k \bmod 4 = 3,
\end{cases}
\qquad
\omega_k=10000^{-k/(d/2)}.
\end{equation}

Here, $\mathbf{E}_{\mathrm{geo}}\in\mathbb{R}^{B\times N\times d}$ is the geographic embedding tensor; $E_{b\ell k}^{\mathrm{geo}}$ is its component for tile $b\in\{1,\ldots,B\}$, patch $\ell\in\{1,\ldots,N\}$, and embedding dimension $k\in\{0,\ldots,d-1\}$; $N_{x,b\ell}$ and $N_{y,b\ell}$ are the corresponding normalized geographic coordinates; and $\omega_k$ is the frequency associated with dimension $k$. The notation $k\bmod4$ denotes the remainder after dividing $k$ by four.

The resulting geospatial embedding is combined with a standard 2D sine-cosine embedding as

\begin{equation}
\mathbf{T}_{\mathrm{pos}}^{(s)}
=
\mathbf{T}^{(s)}
+
\mathbf{PE}_{\mathrm{abs}}
+
\mathbf{E}_{\mathrm{geo}},
\label{eq:hybrid_positional_encoding}
\end{equation}

where $\mathbf{T}_{\mathrm{pos}}^{(s)}\in\mathbb{R}^{B\times N\times d}$ is the position-augmented token sequence and $\mathbf{PE}_{\mathrm{abs}}\in\mathbb{R}^{N\times d}$ is the fixed image-grid embedding, broadcast across the batch. When geographic metadata are unavailable, $\mathbf{E}_{\mathrm{geo}}$ is omitted.

The distinction between the two positional components lies in their spatial reference. The image-grid embedding $\mathbf{PE}_{\mathrm{abs}}$ encodes the relative arrangement of patches within a tile and is therefore identical for the same patch index across different tiles. In contrast, $\mathbf{E}_{\mathrm{geo}}$ is derived from each tile's geotransform and varies with its Earth-referenced location. Consequently, patches occupying the same image-grid position but located in different geographic regions receive the same local positional encoding but different geographic encodings. The methodological contribution therefore lies not in the sinusoidal basis functions themselves, but in combining image-relative and Earth-referenced positional information within the shared multimodal encoder.

\subsection{Masking Strategy}
\label{sec:masking_strategy}

FLORO performs masking independently across optical and auxiliary modality streams following the MultiMAE formulation \citep{Bachmann2022MultiMAE:Target}. Because each stream receives a separately sampled mask, the encoder is exposed to different combinations of visible optical and auxiliary information during pretraining. FLORO additionally uses a three-stage masking curriculum consistent with the easy-to-hard training principle explored in curriculum-based MAEs \citep{Madan2024CL-MAE:Autoencoders,Yoon2025CurriMAE:Classification}. The ratios were selected a priori rather than through a hyperparameter search. Masking 25\% of the patches during the first stage retains substantial spatial and cross-modal context while the model begins learning reconstruction. The ratio is then increased to 50\% as an intermediate stage and finally to 75\%, the high masking ratio commonly used in standard MAE pretraining \citep{He2022MaskedLearners}. The intended effect is to increase reconstruction difficulty gradually by reducing the information available to the encoder. Further details are reported in \autoref{sec:pretraining_implementation}.

\subsection{Transformer Encoder}

The visible patch tokens are processed using a Vision Transformer encoder $f_{\theta}$ \citep{Dosovitskiy2020AnScale}. FLORO uses a ViT-L backbone comprising 24 transformer blocks with token dimension $d=1024$ and 16 attention heads. While the transformer architecture itself follows the standard ViT formulation, the encoder differs from conventional MAEs in that it jointly processes heterogeneous multispectral and auxiliary modality tokens within a unified latent representation space. The design enables the encoder to model complex cross-modal spatial relationships between optical, elevation, and SAR observations despite varying modality availability. Here, $\theta$ denotes the trainable encoder parameters and $\mathbf{Z}$ denotes the resulting joint latent sequence (see \autoref{eq:mae_encoder}). When auxiliary observations are unavailable, the encoder processes only the visible optical tokens. The encoder outputs latent representations together with modality-specific masking information and restoration indices required for reconstruction.

\subsection{Modality-Aware Decoder}

Unlike standard MAEs employing a single reconstruction branch, FLORO introduces modality-aware decoder streams adapted to heterogeneous remote sensing data. Encoded tokens are first projected to the decoder dimension and then separated into modality-specific branches corresponding to multispectral and auxiliary modalities.

The decoder consists of two transformer blocks with embedding dimension 768, 16 attention heads, and MLP ratio 4. Modality-specific self-attention and cross-attention layers allow information exchange between visible optical and auxiliary modality tokens during reconstruction. Separate reconstruction heads are then used to predict multispectral and auxiliary modality targets. The modality-aware reconstruction design encourages the encoder to capture semantically aligned cross-modal representations while remaining robust to incomplete modality coverage.

\subsection{Validity-Aware Reconstruction Objective}

FLORO extends the standard MAE reconstruction loss to heterogeneous multimodal remote sensing inputs with variable spectral availability. Let $\mathbf{x}^{(s)}$ and $\hat{\mathbf{x}}^{(s)}$ denote the reference and reconstructed targets for stream $s\in\{\mathrm{opt},\mathrm{aux}\}$, respectively, with $\hat{\mathbf{x}}^{(s)}=g_{\phi_s}^{(s)}(\mathbf{Z})$. Here, $g_{\phi_s}^{(s)}$ is the decoder branch for stream $s$, $\phi_s$ denotes its trainable parameters, and $\mathbf{Z}$ is the shared encoder output.

FLORO further excludes unavailable or invalid targets from the masked reconstruction loss in \autoref{eq:mae_reconstruction_loss}. For a given stream, the validity-aware masked mean squared error is

\begin{equation}
\mathcal{E}
\left(\hat{\mathbf{x}},\mathbf{x};\mathbf{m},\mathbf{v}\right)
=
\begin{cases}
\displaystyle
\frac{\sum_q m_qv_q \left(\hat{x}_q-x_q \right)^2}{\sum_q m_qv_q},
&
\text{if }
\displaystyle\sum_qm_qv_q>0,
\\[10pt]
0,
&
\text{otherwise},
\end{cases}
\label{eq:validity_masked_mse}
\end{equation}

where $q$ indexes the scalar reconstruction targets, $x_q$ and $\hat{x}_q$ are the reference and reconstructed values of target $q$, $m_q\in\{0,1\}$ indicates whether the patch containing target $q$ was masked, and $v_q\in\{0,1\}$ indicates whether that target is available and valid. The vectors $\mathbf{m}$ and $\mathbf{v}$ collect the masking and validity indicators, respectively. When all targets are valid, \autoref{eq:validity_masked_mse} reduces to the masked mean squared error in \autoref{eq:mae_reconstruction_loss}.

Unlike standard MAEs that reconstruct fixed RGB channels, FLORO performs reconstruction separately for heterogeneous spectral groups including BGR, red-edge, near-infrared, SWIR, elevation, and SAR modalities. Each reconstruction term is weighted using validity-aware gates derived from modality-availability masks, ensuring that missing modalities do not contribute to the optimization objective.

The stream-specific losses are

\begin{equation}
\mathcal{L}_{s}
=
\mathcal{E}\left(\hat{\mathbf{x}}^{(s)},\mathbf{x}^{(s)};\mathbf{m}^{(s)},\mathbf{v}^{(s)}\right),
\qquad
s\in\{\mathrm{opt},\mathrm{aux}\},
\label{eq:stream_reconstruction_loss}
\end{equation}

where $\mathcal{L}_{s}$, $\mathbf{m}^{(s)}$, and $\mathbf{v}^{(s)}$ are the reconstruction loss, masking indicators, and validity indicators for stream $s$, respectively. The final pretraining objective is

\begin{equation}
\min_{\theta,\phi_{\mathrm{opt}},\phi_{\mathrm{aux}}}
\left[
\lambda_{\mathrm{opt}}\mathcal{L}_{\mathrm{opt}}
+
\lambda_{\mathrm{aux}}\mathcal{L}_{\mathrm{aux}}
\right],
\label{eq:floro_pretraining_objective}
\end{equation}

where $\theta$ denotes the encoder parameters; $\phi_{\mathrm{opt}}$ and $\phi_{\mathrm{aux}}$ denote the optical and auxiliary decoder parameters; and $\lambda_{\mathrm{opt}}$ and $\lambda_{\mathrm{aux}}$ are non-negative weights controlling the contributions of the two reconstruction losses. In all experiments, we used equal weighting by setting $\lambda_{\mathrm{opt}}=\lambda_{\mathrm{aux}}=1.0$.

\section{Experiments}
\label{sec:experiments}

The experimental section separates the data used for self-supervised pretraining from the datasets used for downstream evaluation. We first describe the heterogeneous pretraining data, then report pretraining implementation and computational details, and finally summarize the downstream evaluation data and adaptation protocol before presenting task-specific results.

\subsection{Pretraining data}

To expose FLORO to diverse sensing conditions, spatial resolutions, and modality combinations, we assembled a heterogeneous pretraining dataset composed of 90{,}618 image patches of size $256 \times 256$ pixels. The dataset includes four main data sources: Sentinel-1/Sentinel-2 pairs, high-resolution SkySAT satellite imagery, UAV multispectral imagery with corresponding digital surface models (DSMs), and UAV RGB imagery with corresponding DSMs, as shown in \autoref{fig_2}.

\begin{figure}[ph]
\centering
\includegraphics[width=0.75\textwidth]{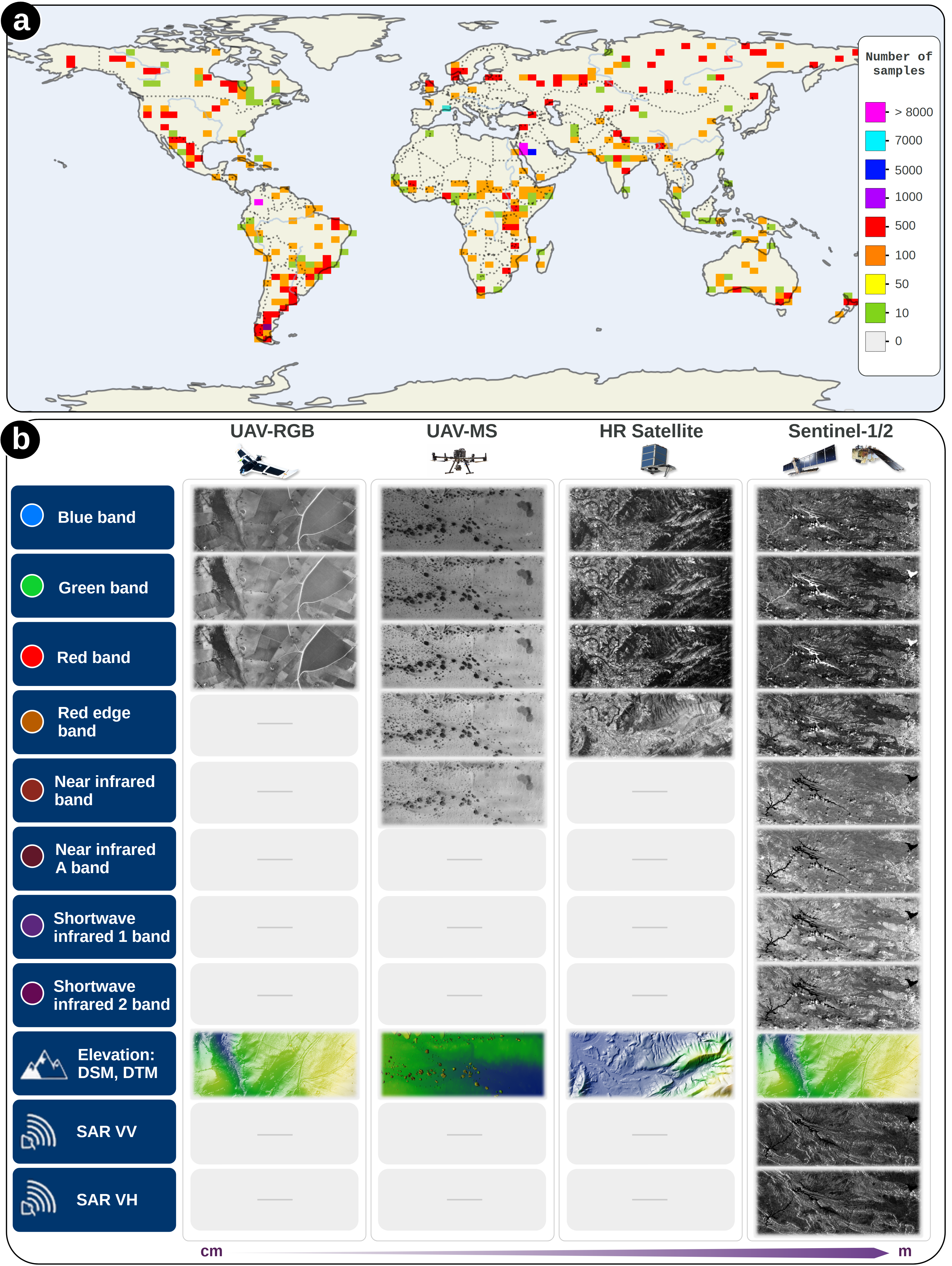}
\caption{Composition and geographic distribution of the heterogeneous pretraining dataset used for FLORO. \textbf{(a)} Global distribution of the 256 $\times$ 256 pixel pretraining patches used in this study. Colors indicate local sample density, from sparsely sampled regions to local concentrations of up to approximately 8,000 patches. \textbf{(b)} Representative examples of the four main source groups included in pretraining: UAV RGB imagery, UAV multispectral imagery, high-resolution satellite imagery, and Sentinel-1/Sentinel-2 data. Rows show the optical spectral groups and auxiliary modalities available for each source, including visible bands, red-edge and near-infrared channels where available, shortwave infrared bands for Sentinel-2, elevation products (DSM/DTM), and SAR backscatter (VV and VH). The figure illustrates the strong heterogeneity in spatial resolution, spectral composition, and modality availability that FLORO is designed to accommodate.}
\label{fig_2}
\end{figure}

The dataset was split into 81{,}282 training samples, 7{,}524 validation samples, and 1{,}812 test samples. The training split contains 29{,}267 Sentinel-1/Sentinel-2 samples, 5{,}638 high-resolution satellite samples, 24{,}816 UAV multispectral-DSM samples, and 21{,}561 UAV RGB-DSM samples. Further dataset details are provided in \ref{app:ap1}. The dataset was designed to cover substantial variation in modality availability, spatial resolution, sensor characteristics, and ecological scene structure. The Sentinel-1/Sentinel-2-based subset has the broadest modality coverage, with all FLORO spectral groups as well as elevation and SAR inputs. The high-resolution SkySAT subset contributes a different satellite sensor, fine spatial detail and radiometric diversity while the UAV-based subsets add airborne platforms, ultra-high-resolution imagery and locally aligned structural information from photogrammetric DSMs. The latter sources therefore broaden sensor, platform, resolution, and source-domain coverage, but do not constitute a simple increase in the number of available modalities. The UAV data were collected across ecologically varied sites in Saudi Arabia and Colombia, enabling the model to encounter fine-scale vegetation structure and heterogeneous dryland scenes during pretraining.

Rather than maximizing dataset size alone, the pretraining dataset combines sources that vary in sensors, platforms, spatial resolutions, radiometry, and input availability. The Sentinel-1/2 subset already supplies the full FLORO modality layout, while the SkySAT and UAV sources broaden sensor-source coverage. The design supports the objective of learning representations that can be used under variable remote sensing input configurations, but it does not assume that every added source improves every downstream task. Because the pretraining dataset is comparatively small relative to those used by recent geospatial foundation models, we used an approximately 90:8:2 split to maximize the number of samples available for representation learning while retaining held-out data for monitoring reconstruction generalization. The validation and test subsets were used only to assess pretraining reconstruction behavior, whereas transfer performance was evaluated independently using downstream PANGAEA benchmarks.

\subsection{Pretraining implementation and computational reporting}
\label{sec:pretraining_implementation}

FLORO is pretrained using a ViT-L encoder with patch size $16 \times 16$ on $256 \times 256$ input tiles. The encoder contains 24 transformer blocks with embedding dimension 1024 and 16 attention heads. The decoder is intentionally lightweight, consisting of two transformer blocks with embedding dimension 768, 16 attention heads, and an MLP ratio of 4. Pretraining is conducted for 150 epochs on 81{,}282 training samples. Each tile produces a $16 \times 16$ grid of projected patch tokens, so one pass through the training set contains 20M (20{,}808{,}192) full-grid patch tokens. Across 150 epochs, this corresponds to 3B (3{,}121{,}228{,}800) nominal full-grid patch-token presentations before accounting for masking. The 150-epoch training schedule was held fixed across the reported pretraining and ablation experiments. Optimization is performed using AdamW \citep{Loshchilov2017DecoupledRegularization} with betas of $(0.90,0.95)$, learning rate $10^{-4}$, and weight decay $0.01$.

Training is performed on 8 NVIDIA V100 GPUs using mixed precision and gradient accumulation. The complete pretraining run required approximately 2.5 days of wall-clock time, corresponding to 480 aggregate V100 GPU-hours. To optimize memory usage, the local batch size and accumulation schedule are adjusted across training stages. During the first 50 epochs, each GPU processes a local batch size of 16 with 4 accumulation steps, resulting in an effective batch size of 512. From epochs 51 to 100, the local batch size is increased to 22 with 3 accumulation steps, resulting in an effective batch size of 528. After epoch 100, the local batch size is increased to 32 with 2 accumulation steps, yielding an effective batch size of 512.

Because masked patches are removed before the encoder, higher masking ratios reduce the number of visible tokens and consequently lower encoder memory and computational requirements. The use of three discrete stages allowed the local batch size and gradient-accumulation schedule to be adjusted at fixed boundaries while maintaining an approximately constant effective global batch size. During epochs 1-50, with 25\% masking, each of the eight GPUs used a local batch size of 16 and four accumulation steps, giving an effective batch size of 512. During epochs 51-100, with 50\% masking, the local batch size was 22 with three accumulation steps, giving an effective batch size of 528. During epochs 101-150, with 75\% masking, the local batch size was 32 with two accumulation steps, giving an effective batch size of 512. A continuously varying or automatically learned masking schedule was not evaluated. Masks for optical and auxiliary modality inputs are generated independently for each batch while preserving the stage-specific masking ratio. Data augmentation is intentionally limited to Gaussian noise, Gaussian blur, and random rotations in fixed $90^\circ$ increments. The maximum Gaussian noise magnitude is set to $0.2$, and the Gaussian blur sigma is set to $1.1$. Rotational augmentation is used to reduce the risk that the model memorizes absolute spatial positions instead of learning transferable visual representations. To simulate variable sensor and modality availability band-drop was applied with a probability of 0.7, conditional on its application, the BGR group was dropped with probability 0.05, elevation with 0.15, and each red-edge, NIR, NIR-A, SWIR, and SAR group with 0.10, while retaining at least one optical group.

Before being passed to the model, modality-specific clipping is applied to normalize input ranges. Optical values are clipped to $[0,1]$, elevation values to $[-500,9000]$, and SAR values to $[-60,20]$, based on the empirical pretraining data distribution. Mixed precision training is used throughout, with selected operations executed in half precision 16-bit floating-point (FP16) and numerically sensitive operations retained in single precision 32-bit floating-point (FP32). Gradient scaling is applied to stabilize backpropagation under reduced precision \citep{Micikevicius2017MixedTraining}, and distributed data loading ensures balanced, non-overlapping subsets across GPUs \citep{Goyal2017AccurateHour}.

\paragraph{Model scale and computational reporting}
To avoid conflating pretraining data volume with computational efficiency, we explicitly report the model scale and inference cost of FLORO and the evaluated baselines. Parameter counts were computed from the instantiated PANGAEA encoder wrappers used for downstream evaluation. The evaluated variants were determined by the pretrained encoder weights and model implementations available through PANGAEA at the time of the experiments; the comparison therefore reflects the models that could be evaluated rather than an exhaustive or architecture-matched set of variants from each model family. The reported parameter counts represent encoder-wrapper sizes and exclude pretraining-only heads or decoders and task-specific downstream modules. Inference cost was profiled on EuroSAT-MS with batch size 1 using each model's required optical input shape and the same lightweight linear classifier. We report estimated floating-point operations as twice the multiply-accumulate count ($\mathrm{FLOPs}=2\times\mathrm{MACs}$). Because native input sizes and numbers of input channels differ among models, these measurements represent the cost of the evaluated PANGAEA pipelines rather than architecture-only FLOPs under an identical input tensor. FLORO contains 307.343M encoder-wrapper parameters and requires 162.920 GFLOPs per sample under this protocol. Its parameter count is comparable to TerraMind-L (305.401M), Scale-MAE (303.098M), and CROMA (303.025M), while its measured inference cost is higher than those models (123.370, 119.380, and 141.293 GFLOPs, respectively) and lower than SpectralGPT (213.075 GFLOPs). TerraMind-L provides the closest reported training-scale reference among these parameter-comparable models. It was pretrained on 500 billion tokens using 32 NVIDIA A100 GPUs for 10 days, corresponding to 7{,}680 aggregate A100 GPU-hours \citep{Jakubik2025TerraMind:Observation}. FLORO used 8 NVIDIA V100 GPUs for 2.5 days, corresponding to 480 aggregate V100 GPU-hours. Its training set contains 20M full-grid patch tokens per pass and 3B nominal full-grid patch-token presentations across 150 epochs before accounting for masking (see \autoref{tab:closest_model} in \ref{app:model_scale}). These device-hour and token figures are not a hardware-normalized compute ratio because the two studies differ in accelerator generation, token definition, input representation, and pretraining objective. Thus, FLORO should be interpreted as compact with respect to pretraining data volume and reported training device-time, not as a small-parameter or low-FLOP model. Detailed encoder-wrapper parameter counts, required input shapes, and inference-cost estimates for all evaluated models are reported in \autoref{tab:encoder_parameter_counts} in \ref{app:model_scale}.

\subsection{Downstream evaluation data and adaptation protocol}

Downstream evaluation separates the benchmark datasets, prediction targets, and adaptation protocol from the task-specific results reported below.

To contextualize FLORO's performance, we compared it against 13 recent geospatial foundation models spanning multimodal, multispectral, vision-language, and scale-aware pretraining paradigms using the PANGAEA benchmark \citep{Marsocci2026PANGAEA:Benchmark}. PANGAEA is a standardized evaluation framework designed to assess the transferability of geospatial foundation models across diverse Earth observation tasks using a unified protocol and consistent downstream decoders. These include multimodal and sensor-fusion models such as CROMA \citep{Fuller2023CROMA:Autoencoders}, DOFA \citep{Xiong2025NeuralObservation}, and TerraMind-L \citep{Jakubik2025TerraMind:Observation}; general-purpose geospatial encoders such as GFM \citep{Mendieta2023TowardsPretraining}, Prithvi \citep{Jakubik2023FoundationIntelligence}, and SatlasNet \citep{Bastani2023SatlasPretrain:Understanding}; spectral and scale-aware approaches such as SpectralGPT \citep{Hong2024SpectralGPT:Model}, and Scale-MAE \citep{Reed2022Scale-MAE:Learning}; the vision-language model RemoteCLIP \citep{Liu2024RemoteCLIP:Sensing}; and four SSL4EO-S12 variants trained with MoCo, DINO, MAE, and Data2Vec \citep{Wang2023SSL4EO-S12:Observation}. In addition to these pretrained foundation models, we included two from-scratch reference baselines: a U-Net \citep{Ronneberger2015U-Net:Segmentation}, representing a conventional convolutional encoder-decoder architecture for dense prediction, and a ViT baseline \citep{Dosovitskiy2020AnScale}, representing a transformer architecture trained without pretraining. 

\begin{figure}[htbp]
\centering
\includegraphics[width=\linewidth]{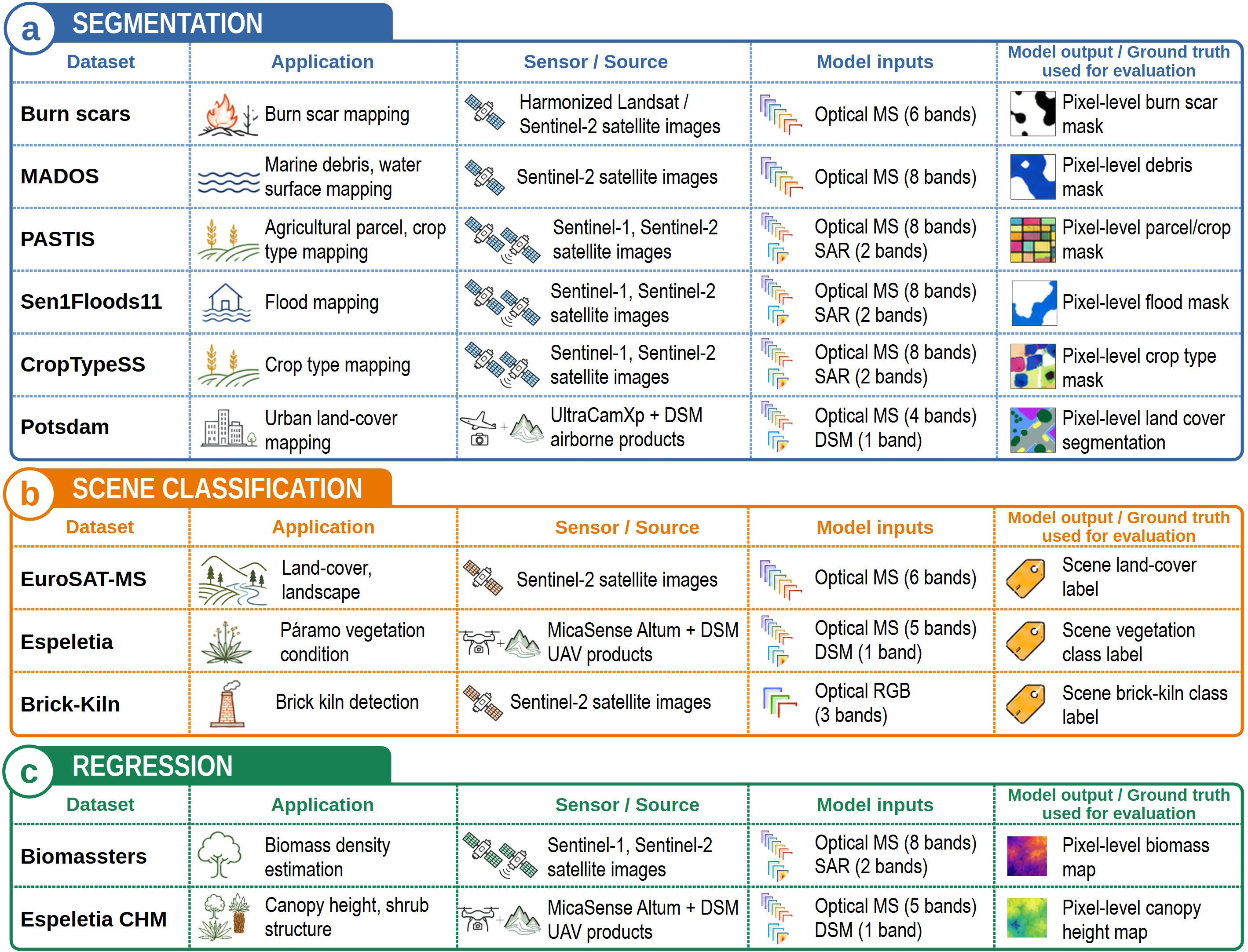}
\caption{Summary of downstream datasets used to evaluate FLORO and the competing models. Datasets are grouped by task type: \textbf{(a)} semantic segmentation, \textbf{(b)} scene classification, and \textbf{(c)} regression. For each dataset, the figure reports the application context, sensor/source, FLORO input groups used for evaluation, and prediction target or ground-truth data. The number of optical, SAR, and DSM bands refers to the channels provided to FLORO after mapping each dataset into the FLORO grouped input representation and excludes validity channels. The benchmark suite spans optical-only, optical-SAR, and optical-elevation settings; scene-level and pixel-level prediction targets; and applications ranging from land-cover and flood mapping to ecological plant-condition assessment, brick-kiln detection, biomass estimation, and canopy-height reconstruction.}
\label{fig:downstream_datasets}
\end{figure}

Together, these baselines cover a broad range of architectural designs, sensing assumptions, and pretraining strategies, providing a strong basis for evaluating the transferability of FLORO. Following the PANGAEA protocol, model performance was assessed separately for segmentation, scene classification, and regression. For segmentation, we report mean Intersection over Union (mIoU), together with dataset-wise comparisons and average rank across benchmarks. For scene classification, we report overall accuracy (OA) and F1-score. For regression, we report $R^2$, RMSE, relative RMSE (rRMSE) and percentile-range-normalized RMSE ($\mathrm{NRMSE}_{P{95}-P_{5}}$) at both pixel and aggregated image-tile levels where applicable. The rRMSE is calculated as $\mathrm{RMSE}/\bar{y}\times100$, where $\bar{y}$ is the mean of the reference values. To provide a complementary normalization for right-skewed target distributions concentrated at low values, $\mathrm{NRMSE}_{P{95}-P_{5}}$ is calculated as $\mathrm{RMSE}/[P_{95}(y)-P_{5}(y)]\times100$, where $P_{95}(y)$ and $P_{5}(y)$ are the 95th and 5th percentiles of the reference distribution, respectively. The percentile-based denominator represents the central 90\% of the reference distribution and is less sensitive to extreme observations than the full minimum-to-maximum range. At each evaluation level, the normalization statistics are calculated from the corresponding pixel-level or image-tile-level reference distribution. Unless otherwise noted, higher values indicate better performance, except for RMSE, rRMSE and $\mathrm{NRMSE}_{P{95}-P_{5}}$, where lower values are preferred. \autoref{fig:downstream_datasets} summarizes the downstream datasets, applications, sensor or data source, FLORO input groups, model outputs, and ground-truth data used for evaluation.

The ViT and U-Net baselines were trained from scratch, whereas pretrained encoders were kept frozen and only the downstream heads were trained. The protocol isolates the transferability of the pretrained representations under limited downstream adaptation. Improvements over from-scratch baselines therefore indicate that the frozen encoders already encode spatial, spectral, or multimodal features useful for downstream prediction, rather than relying solely on task-specific end-to-end optimization. For semantic segmentation and regression tasks, all models except U-Net, which uses its native encoder-decoder architecture, were evaluated with the standardized decoder configurations provided by PANGAEA, based on UPerNet-style dense prediction heads \citep{Xiao2018UnifiedUnderstanding}. For multi-temporal datasets, models that operate on single images were adapted using the linear mapping and lightweight temporal attention encoder (L-TAE) \citep{Garnot2020LightweightTimeseries} provided by PANGAEA.

Accordingly, the purpose of this protocol is to compare FLORO with state-of-the-art foundation-model encoders under matched, deliberately constrained adaptation conditions. The reported values measure the transfer quality of the learned representations when the same general decoder design is used, rather than the performance ceiling of a model specifically optimized for each downstream dataset. Dataset-specific decoder redesign, end-to-end fine-tuning, and extensive task-specific optimization could improve absolute performance, but constitute a separate research question beyond the scope of this study.

Pretraining, semantic segmentation, and scene-classification experiments were performed on NVIDIA V100-SXM2-32GB GPUs, whereas regression experiments were performed on an NVIDIA A100-SXM4-80GB GPU due to hardware availability on Ibex. Within each downstream task, all compared models were evaluated under the same training and hardware configuration.

\subsubsection{Image segmentation}

For semantic segmentation, we evaluated FLORO on six PANGAEA datasets spanning diverse sensing conditions, modalities, and semantic targets. These include two optical benchmarks (HLS Foundation Burnscars \citep{Phillips2023HLSDataset} and MADOS \citep{Kikaki2024DetectingImagery}), multimodal optical-SAR datasets (PASTIS \citep{Garnot2021PanopticNetworks, SainteFareGarnot2022Multi-modalSeries}, Sen1Floods11 \citep{Bonafilia2020Sen1Floods11:Sentinel-1b}, and CropTypeMapping South Sudan (CropTypeSS) \citep{Rustowicz2019SemanticMethods}), as well as one multimodal optical-elevation benchmark (ISPRS Potsdam \citep{ISPRSISPRSDataset}). These datasets cover burned-area mapping, marine debris detection, agricultural and land-cover discrimination, flood mapping, and high-resolution urban segmentation. Together, they provide a broad test of transferability under frozen-encoder adaptation across different spatial resolutions, scene structures, and modality combinations, ranging from medium-resolution satellite imagery to very high-resolution aerial data, and from optical-only inputs to settings where SAR or elevation provide complementary information.

Semantic segmentation provided the clearest evidence of FLORO's transferability across heterogeneous remote sensing conditions. Across the six PANGAEA segmentation benchmarks, which span optical-only, optical-SAR, and optical-elevation settings, FLORO consistently ranked among the strongest models and achieved the second-best average rank among all evaluated models despite being pretrained on fewer than 100{,}000 samples (\autoref{tab:floro_pangaea_segmentation}). FLORO's pretraining dataset is approximately 113 times smaller than that of TerraMind-L, the strongest-performing segmentation model in our comparison, and approximately 144 times smaller than the largest reported pretraining dataset among the evaluated models. It is also about 4.5 times smaller than the smallest competing foundation model pretraining dataset considered here. The average-rank result is particularly informative because it summarizes consistency across datasets rather than performance on a single benchmark; FLORO's average rank of 3.667 indicates that it remained near the top of the model set across diverse segmentation tasks, even when individual datasets favored different architectures or pretraining regimes. Although TerraMind-L obtained the strongest aggregate segmentation performance, FLORO remained among the top-performing models across the benchmark suite and showed particularly robust behavior across datasets with markedly different spatial resolutions, scene structures, and modality combinations. 

\begin{table}[htb]
\caption{Semantic-segmentation performance of FLORO and competing foundation models on six datasets from the PANGAEA benchmark \citep{Marsocci2026PANGAEA:Benchmark}. Results are reported as mean Intersection over Union (mIoU, \%), together with the average mIoU across datasets and the average rank across benchmarks. Avg.\ Rank is computed by ranking the models separately within each dataset and averaging these ranks across the six segmentation benchmarks; lower values indicate more consistently strong performance across datasets. The best model per column is highlighted in bold, the second best is double underlined, and the third best is underlined. Datasets marked with $^{\dagger}$ are optical-only, $^{{\dagger}{\dagger}}$ include SAR data, and $^{{\dagger}{\dagger}{\dagger}}$ include elevation data. For pretrained models, encoders were kept frozen and evaluated under the standardized PANGAEA adaptation protocol; U-Net and ViT baselines were trained from scratch for each dataset. Higher mIoU indicates better segmentation performance.}
\label{tab:floro_pangaea_segmentation}
\centering
\small
\setlength{\tabcolsep}{6pt}
\resizebox{\textwidth}{!}{%
\begin{tabular}{lcccccc|cc}
\toprule
Model & BurnSr$^{\dagger}$ & MADOS$^{\dagger}$ & PASTIS$^{{\dagger}{\dagger}}$ & Sen1Floods11$^{{\dagger}{\dagger}}$ & CropTypeSS$^{{\dagger}{\dagger}}$ & Potsdam$^{{\dagger}{\dagger}{\dagger}}$ & Avg.\ mIoU & Avg.\ Rank \\
\midrule
Scale-MAE          & 76.747 & 52.649 & 24.601 & 71.348 & 24.632 & \underline{\underline{77.230}} & 54.534 & 11.833 \\
CROMA            & 82.545 & 59.217 & 33.133 & 84.513 & 44.621 & 69.900 & 62.321 & 8.000 \\
GFM-Swin       & 76.891 & \underline{63.357} & 20.425 & 70.505 & 46.041 & 75.275 & 58.749 & 10.000 \\
Prithvi 1.0 100M & 80.658 & 42.532 & 30.446 & 86.914 & \underline{51.505} & 54.400 & 57.743 & 10.167 \\
SatlasNet & 79.982 & 55.410 & 17.431 & 82.900 & 47.663 & 69.070 & 58.743 & 11.000 \\
SSL4EO-S12-MoCo    & 81.548 & 50.177 & 33.979 & 87.009 & 38.532 & 67.115 & 59.727 & 9.500 \\
SSL4EO-S12-DINO    & 81.684 & 57.314 & 35.991 & 85.949 & 46.498 & 67.190 & 62.438 & 7.667 \\
SSL4EO-S12-MAE     & 81.911 & 52.673 & 32.711 & 86.119 & 33.551 & 65.084 & 58.675 & 10.167 \\
SSL4EO-S12-Data2Vec  & 82.151 & 52.482 & 34.685 & 85.448 & 37.846 & 66.490 & 59.850 & 10.000 \\
SpectralGPT         & 80.815 & 48.763 & 35.671 & 86.413 & 47.502 & 69.056 & 61.370 & 8.667 \\
RemoteCLIP          & 76.666 & 60.706 & 17.900 & 67.210 & 50.335 & 75.416 & 58.039 & 10.000 \\
DOFA               & 80.670 & \underline{\underline{63.467}} & 30.175 & 84.630 & 24.921 & \textbf{77.926} & 60.298 & 8.500 \\
TerraMind-L         & \underline{\underline{83.151}} & \textbf{66.077} & \textbf{42.646} & \underline{87.128} & \textbf{56.743} & 74.089 & \textbf{68.306} & \textbf{2.500} \\
\midrule
U-Net Baseline & \textbf{85.116} & 59.122 & 27.069 & \textbf{91.102} & 41.485 & \underline{77.217} & \underline{63.519} & \underline{5.500} \\
ViT Baseline  & 82.573 & 42.105 & \underline{\underline{40.052}} & 85.904 & 26.123 & 71.262 & 58.003 & 8.833 \\
\midrule
FLORO& \underline{82.820} & 55.717 & \underline{36.299} & \underline{\underline{87.567}} & \underline{\underline{53.558}} & 75.595 & \underline{\underline{65.259}} & \underline{\underline{3.667}} \\
\bottomrule
\end{tabular}}
\end{table}

Semantic segmentation results suggest that carefully curated multimodal diversity can partially compensate for limited pretraining scale, yielding transferable representations without requiring the massive data volumes used by the largest geospatial foundation models. \autoref{fig:results_segmentation} complements these numerical results with qualitative examples from all six segmentation datasets. Across these examples, FLORO generally produced predictions that are more spatially coherent and more closely aligned with the reference masks, particularly in preserving elongated structures in MADOS, connected flood regions in Sen1Floods11, and sharp urban boundaries in Potsdam. These visual results suggest that FLORO's transfer strength is not limited to aggregated accuracy, but also extends to the recovery of meaningful spatial structure under highly variable sensing conditions, in line with its consistently strong mIoU scores across datasets (\autoref{tab:floro_pangaea_segmentation}).

\begin{figure}[p]
    \centering
    \includegraphics[width=\linewidth]{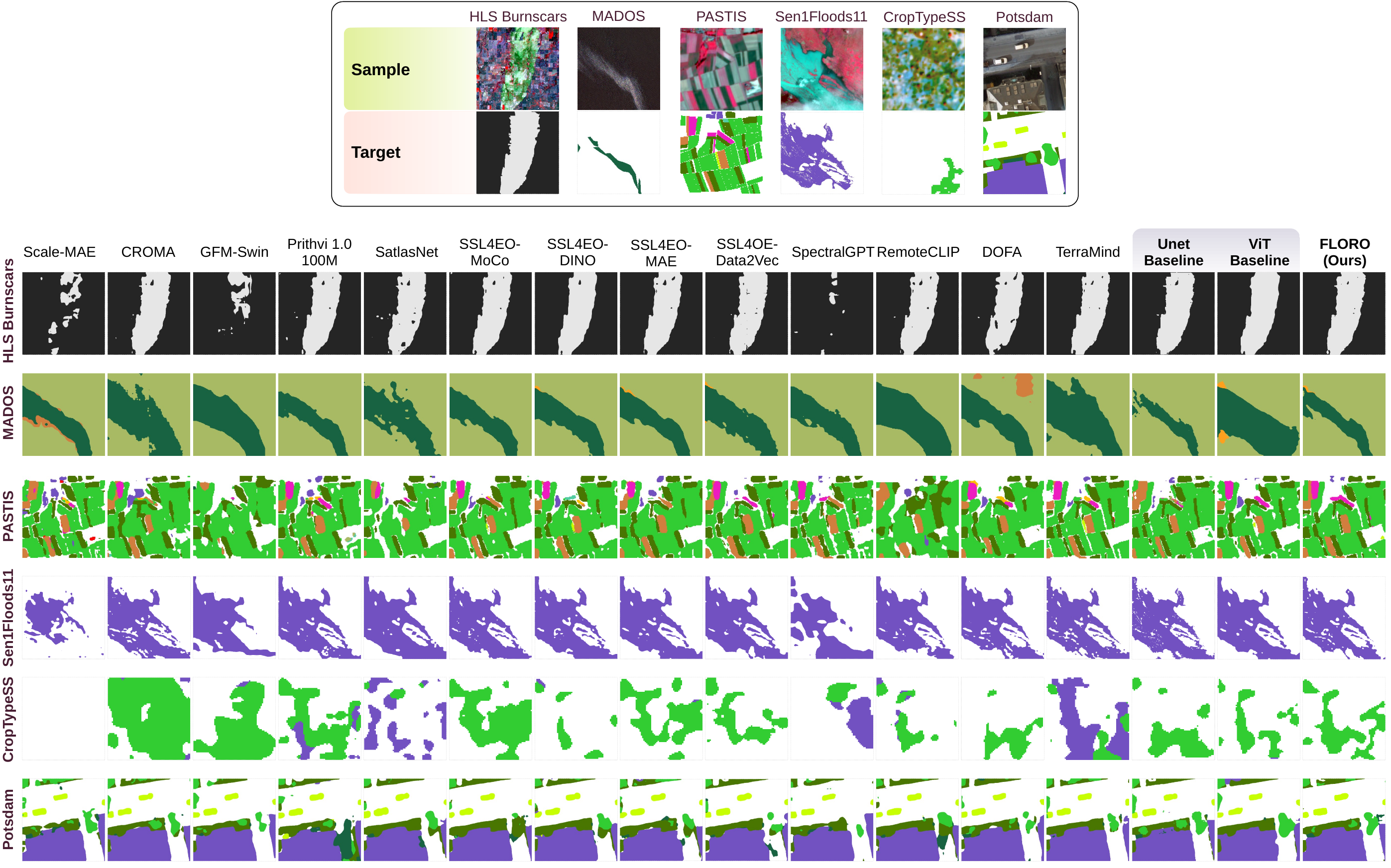}
    \caption{Qualitative segmentation results across six datasets from the PANGAEA benchmark \citep{Marsocci2026PANGAEA:Benchmark}. The top panel shows an input example and the reference target mask, where each column corresponds to one benchmark dataset: HLS Burn Scars, MADOS, PASTIS, Sen1Floods11, CropTypeSS, and ISPRS Potsdam. The bottom panel shows the predictions obtained from each pretrained encoder, each row corresponds to one benchmark dataset. The datasets span optical-only, optical-SAR, and optical-elevation settings and include substantial variation in spatial resolution, scene structure, and semantic classes. Across these heterogeneous conditions, FLORO generally produces predictions that are more spatially coherent and more consistent with the reference annotations, particularly in preserving object continuity, large connected regions, and sharp structural boundaries while reducing scattered false positives.}
    \label{fig:results_segmentation}
\end{figure}

\subsubsection{Scene classification}
\label{sec:scene_class}

Scene classification was evaluated on three benchmarks spanning general land-cover recognition, ecological plant-condition mapping, and environmental monitoring. EuroSAT-MS \citep{Helber2019EuroSAT:Classification} is derived from Sentinel-2 imagery and provides scene-level land-use and land-cover labels. The Espeletia scene-classification benchmark \citep{Rodriguez2026EspeletiaParamo} assigns each image tile to one of three classes describing the presence and health condition of \textit{Espeletiopsis}: no \textit{Espeletiopsis} present, healthy \textit{Espeletiopsis} present, or unhealthy \textit{Espeletiopsis} present. The Brick Kiln Classification benchmark \citep{Lee2021ScalableCapacity} from Bangladesh uses low-resolution Sentinel-2 image tiles prepared with Google Earth Engine and distinguishes tiles with no kiln from tiles where a brick kiln is present. Assessing performance on these datasets provides a broad test of scene-level transfer that includes an ecological plant-health task and an environmental monitoring task.

\begin{table}[htb]
\caption{Scene-classification performance under the frozen-encoder evaluation protocol, reported as accuracy and F1-score for EuroSAT-MS, Espeletia, and Brick Kiln Classification. For each dataset, Accuracy and F1 are first averaged for each model. Models are then ranked by this within-dataset mean, and Avg. Rank is the mean of the ranks across the three datasets; lower values indicate more consistently strong performance. The best model per column is highlighted in bold, the second best is double underlined, and the third best is underlined. Downstream adaptation is restricted to a lightweight linear classification head to emphasize representation quality rather than task-specific fine-tuning. Higher accuracy and F1 values indicate better scene-classification performance. S.-S12 abbreviates SSL4EO-S12.}
\label{tab:scene_classification_results}
\centering
\scriptsize
\setlength{\tabcolsep}{3.5pt}
\begin{tabular}{lccccccc}
\toprule
    &\multicolumn{2}{c}{EuroSAT-MS} & \multicolumn{2}{c}{Espeletia} &\multicolumn{2}{c}{Brick Kiln} & \\
Model &  Acc. & F1 & Acc. &  F1 & Acc. &  F1 & Avg. Rank \\
\midrule
Scale-MAE         & 0.860 & 0.860 & \underline{\underline{0.784}} & \textbf{0.743} & 0.961 & 0.956 & 5.33 \\
CROMA             & 0.900 & 0.900 & 0.703 & 0.677 & 0.934 & 0.925 & 7.67 \\
GFM-Swin          & \textbf{0.962} & \textbf{0.962} & \underline{0.770} & 0.694 & 0.967 & 0.963 & \underline{\underline{2.67}} \\
Prithvi 1.0 100M  & 0.791 & 0.787 & 0.635 & 0.545 & 0.901 & 0.887 & 13.67 \\
SatlasNet         & 0.897 & 0.896 & 0.709 & 0.614 & 0.863 & 0.846 & 10.67 \\
S.-S12-MoCo       & 0.807 & 0.804 & 0.669 & 0.596 & 0.920 & 0.909 & 10.67 \\
S.-S12-DINO       & 0.757 & 0.754 & 0.649 & 0.562 & 0.910 & 0.899 & 13.67 \\
S.-S12-MAE        & 0.791 & 0.789 & 0.703 & 0.614 & 0.919 & 0.909 & 11.33 \\
S.-S12-Data2Vec   & 0.783 & 0.782 & 0.649 & 0.589 & 0.923 & 0.913 & 12.00 \\
SpectralGPT       & 0.887 & 0.888 & 0.736 & 0.687 & 0.950 & 0.944 & 7.33 \\
RemoteCLIP        & \underline{\underline{0.937}} & \underline{\underline{0.937}} & 0.757 & 0.692 & 0.958 & 0.953 & 4.00 \\
DOFA              & 0.901 & 0.902 & \textbf{0.797} & \underline{\underline{0.740}} & \textbf{0.973} & \textbf{0.970} & \textbf{2.33} \\
TerraMind-L       & \underline{0.913} & \underline{0.913} & 0.757 & 0.681 & \underline{0.968} & \underline{0.964} & 4.00 \\
\hline
ViT Baseline      & 0.807 & 0.802 & 0.723 & 0.683 & 0.896 & 0.881 & 11.00 \\
\hline
FLORO             & 0.908 & 0.909 & 0.750 & \underline{0.695} & \underline{\underline{0.970}} & \underline{\underline{0.966}} & \underline{3.67} \\
\bottomrule
\end{tabular}
\end{table}

Under the frozen-encoder protocol, pretrained encoders were adapted using only a lightweight linear classification head, so the comparison emphasizes encoder quality rather than task-specific fine-tuning. FLORO achieved the third-best average rank (3.67) across the three scene-classification benchmarks (\autoref{tab:scene_classification_results}), behind DOFA (2.33) and GFM-Swin (2.67). GFM-Swin obtained the strongest EuroSAT-MS performance, while DOFA achieved the highest Espeletia accuracy and Brick Kiln scores. FLORO obtained the third-highest Espeletia F1 (0.695) and the second-highest Brick Kiln accuracy (0.970) and F1 (0.966). These results indicate that FLORO's frozen representations remain competitive across heterogeneous scene-classification benchmarks spanning different sensors, spatial resolutions, and application domains, placing FLORO among the leading models in the aggregate classification ranking. The practical implication is that useful scene-level transfer can be obtained from a compact, heterogeneous pretraining dataset, while model size and inference cost should be interpreted separately from pretraining data volume.

\autoref{fig:cams} provides complementary qualitative context for the geo-positional experiment summarized with the other ablations in \autoref{fig:ablations}. The figure shows Class Activation Maps (CAMs) \citep{Zhou2016LearningLocalization} after one epoch of training of the lightweight classification head. At this early stage, the activation patterns are primarily shaped by the frozen encoder features rather than extensive task-specific adaptation. Compared with absolute positional encoding, geo-positional encoding produces sharper and more spatially coherent activation patterns for several classes, suggesting that explicit geographic information can alter the spatial organization of the learned representation even when its effect on downstream performance is small.

\begin{figure}[htbp]
    \centering
    \includegraphics[width=\linewidth]{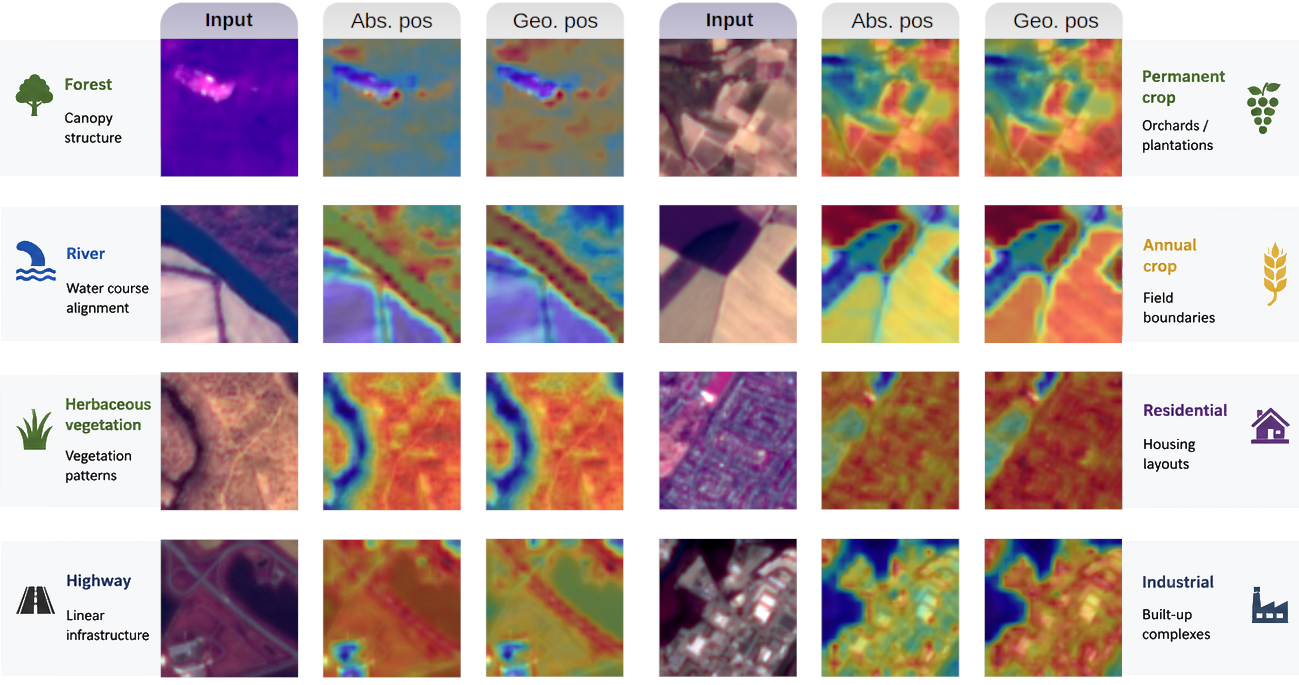}
    \caption{Effect of positional encoding on early scene-discriminative representations in EuroSAT-MS. Class activation maps (CAMs) are shown for selected EuroSAT-MS classes after one epoch of training of a lightweight classification head on frozen FLORO encoder features. For each class, the input image is shown alongside the activation pattern obtained with absolute positional encoding and with geo-positional encoding. At this early stage, the activation maps primarily reflect the structure of the pretrained representations rather than extensive task-specific adaptation. Geo-positional encoding produces sharper and more spatially coherent activation regions for several classes, including rivers, highways, permanent crops, and residential areas. These qualitative differences suggest that explicit geographic information can influence the spatial organization of the learned representation, although the downstream ablation shows only small and task-dependent quantitative effects.}
    \label{fig:cams}
\end{figure}

\subsubsection{Regression tasks}
\label{sec:regression_tasks}

Regression tasks provided a complementary test of FLORO's transferability beyond discrete semantic prediction, focusing instead on continuous ecological variables with strong spatial structure. Unlike segmentation and scene classification, these tasks require the model to preserve both spatial organization and continuous magnitude information. Errors therefore reflect not only whether the correct structures are localized, but also whether their biomass or canopy-height values are numerically calibrated. We evaluated foundation model performance on Biomassters \citep{Nascetti2023BioMassters:Series}, a multimodal biomass-estimation benchmark characterized by parcel geometry, field boundaries, and frequent degradation of optical observations, and on the Espeletia dataset \citep{Rodriguez2026EspeletiaParamo}, which targets canopy-height reconstruction in structurally heterogeneous natural vegetation. Together, these benchmarks assess whether pretrained representations can support not only broad scene understanding, but also the recovery of fine-grained continuous ecological patterns. All models were trained for 80 epochs using a batch size of 8.

\begin{figure}[p]
\centering
\includegraphics[width=0.9\linewidth]{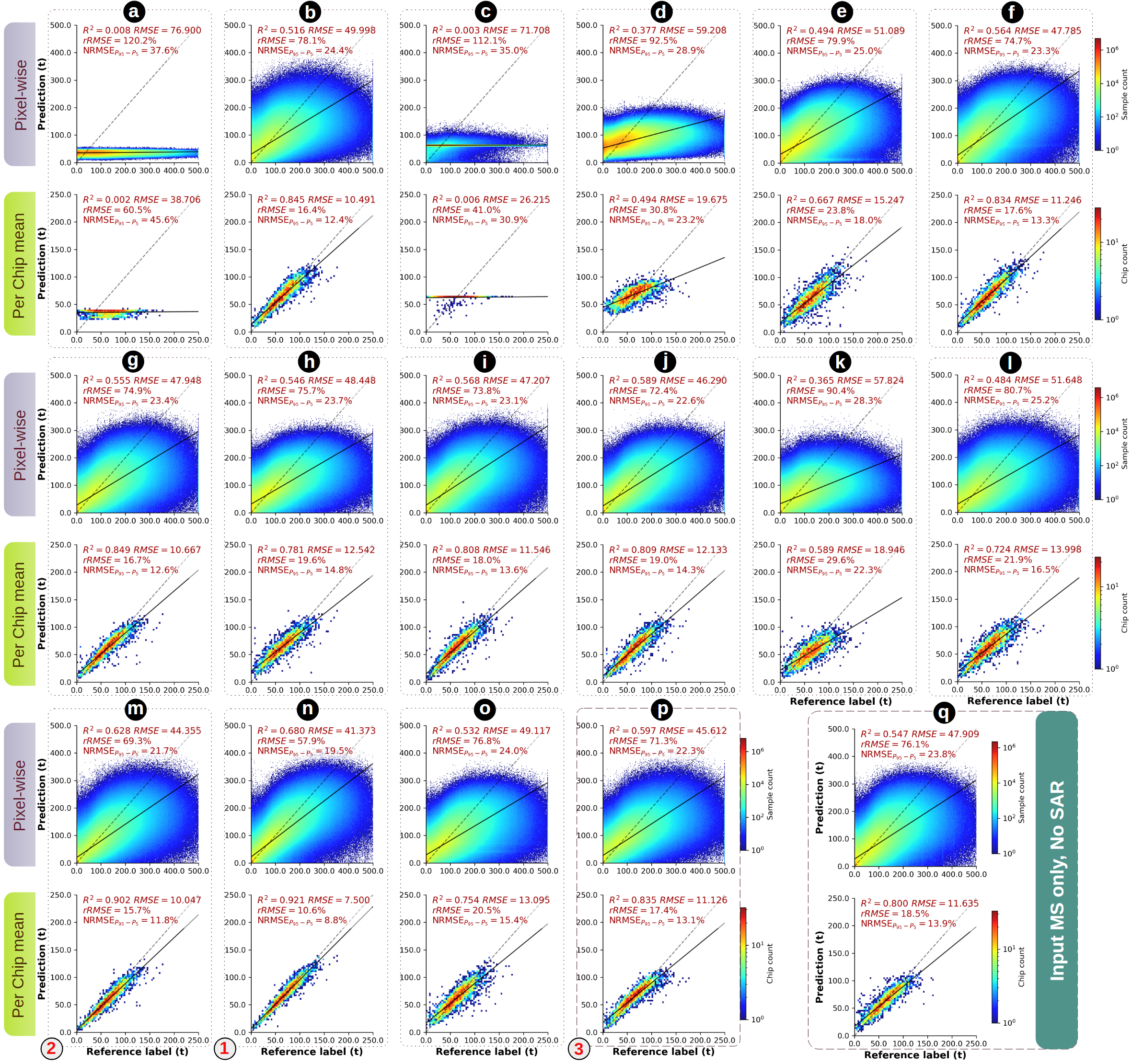}
\caption{Regression performance on the quality-selected, single-month BioMassters benchmark under the PANGAEA evaluation framework \citep{Marsocci2026PANGAEA:Benchmark}. For each model, scatterplots summarize agreement between predictions and reference biomass values at the pixel level and after image-tile-level aggregation, reported using the coefficient of determination ($R^2$), root mean squared error (RMSE), relative RMSE (rRMSE), and $\mathrm{NRMSE}_{P{95}-P_{5}}$. The axis shorthand ``t'' retains the source-documented BioMassters unit, tonnes per cell, and does not denote a conversion to tonnes per hectare. The evaluated models are: \textbf{(a)} Scale-MAE, \textbf{(b)} CROMA, \textbf{(c)} GFM-Swin, \textbf{(d)} Prithvi 1.0 100M, \textbf{(e)} SatlasNet, \textbf{(f)} SSL4EO-S12-MoCo, \textbf{(g)} SSL4EO-S12-DINO, \textbf{(h)} SSL4EO-S12-MAE, \textbf{(i)} SSL4EO-S12-Data2Vec, \textbf{(j)} SpectralGPT, \textbf{(k)} RemoteCLIP, \textbf{(l)} DOFA, \textbf{(m)} TerraMind-L, \textbf{(n)} U-Net Baseline, \textbf{(o)} ViT Baseline, \textbf{(p)} FLORO, and \textbf{(q)} FLORO evaluated with multispectral input only, after removing SAR. The U-Net, ViT, and TerraMind-L comparisons retain the benchmark's default optical-SAR input configuration. \textcircled{\textcolor{red}{\scriptsize 1}}, \textcircled{\textcolor{red}{\scriptsize 2}}, and \textcircled{\textcolor{red}{\scriptsize 3}} indicate the best, second-best, and third-best results, respectively; the same three-model ordering is obtained at the pixel and image-tile levels. These results show that models differ not only in overall numerical accuracy but also in their ability to recover parcel-scale biomass structure, with FLORO achieving among the strongest and most balanced performance across pixel- and image-tile-level evaluation.}
\label{fig:biomassters_evaluation}
\end{figure}

BioMassters comprises approximately 13{,}000 non-overlapping Finnish forest areas observed monthly with Sentinel-1 and Sentinel-2 imagery and paired with airborne-LiDAR-derived above-ground biomass (AGB) maps \citep{Nascetti2023BioMassters:Series}. Each image tile covers $2{,}560 \times 2{,}560$~m and contains a $256 \times 256$ AGB raster sampled at 10~m. The BioMassters documentation uses the term chip for one such spatial patch; in this paper, we use image tile to avoid ambiguity. The AGB targets were derived from forest-inventory attributes originally organized on a $16 \times 16$~m grid using calibrated tree-specific allometric equations and were subsequently upsampled and co-registered to the benchmark grid. To make standardized decoder training tractable across all evaluated encoders, we retained one quality-selected month for each of the 8{,}689 locations in the labeled training collection rather than all 104{,}268 tile-month observations. Following the public BioMassters preprocessing workflow, the month was selected using agreement between standardized nonnegative NDVI and Sentinel-1 VH together with a predefined seasonal preference; the AGB targets were not used for selection. All models used the same selected observations, splits, decoder design, and evaluation procedure. The source documentation denotes the stored AGB values as tonnes per cell (pixel), although it also uses biomass-density terminology without documenting an area-normalized conversion of the upsampled targets. We therefore report results on the native target scale as t/px and apply no conversion to t/ha. Accordingly, the comparison applies to this quality-selected single-month configuration, which may underrepresent poor-quality optical observations, and not to the complete multitemporal BioMassters dataset.

Results on the Biomassters benchmark show that FLORO achieves amongst the strongest quantitative performance in recovering spatially structured biomass patterns. A key factor underlying this performance is its ability to recover plausible biomass structure under multimodal and visually degraded conditions, where optical inputs are partially corrupted. Consistent performance at both pixel and image-tile levels further indicates that the learned representations transfer effectively to parcel-structured biomass estimation (\autoref{fig:biomassters_evaluation}). FLORO ranked third at both evaluation scales. The mean reference value was approximately 64.0 t/px at both scales. At the pixel level, it obtained $R^2 = 0.597$, RMSE $= 45.612$ tonnes per cell, rRMSE $= 71.3\%$ and $\mathrm{NRMSE}_{P{95}-P_{5}} = 22.3\%$. Thus, despite its rank among the evaluated frozen encoders, the pixel-level error remained substantial relative to the reference mean. After image-tile-level mean aggregation, FLORO obtained $R^2 = 0.835$, RMSE $= 11.126$ tonnes per cell, rRMSE $= 17.4\%$ and $\mathrm{NRMSE}_{P{95}-P_{5}} = 13.1\%$. Aggregation therefore reduced sensitivity to local pixel errors by comparing one mean prediction and one mean reference value per $2{,}560 \times 2{,}560$~m tile. The U-Net baseline ranked first and TerraMind-L ranked second at both scales; FLORO was therefore the strongest evaluated pretrained frozen-encoder model after TerraMind-L while preserving plausible parcel-scale structure.

The qualitative comparisons in \autoref{fig:biomassters_results} concern biomass regression and help contextualize the quantitative results in \autoref{fig:biomassters_evaluation}. Scale-MAE and GFM-Swin produce near-uniform low-biomass predictions in several examples, consistent with their very low pixel-level $R^2$ values (0.008 and 0.003) and high rRMSE values (120.2\% and 112.1\%). RemoteCLIP and Prithvi 1.0 100M retain more scene variation but produce visibly over-smoothed predictions with weak parcel boundaries; their pixel-level $R^2$ values are 0.365 and 0.377, their rRMSE values are 90.4\% and 92.5\%, respectively, and their $\mathrm{NRMSE}_{P{95}-P_{5}}$ are 28.3\% and 28.9\%. In contrast, FLORO better preserves parcel geometry, transitions between low- and high-biomass areas, and plausible biomass structure where the optical image is affected by shadow or missing detail. These observations are illustrative model-level comparisons rather than a stratified error analysis.

\begin{figure}[p]
    \centering
    \includegraphics[width=1\linewidth]{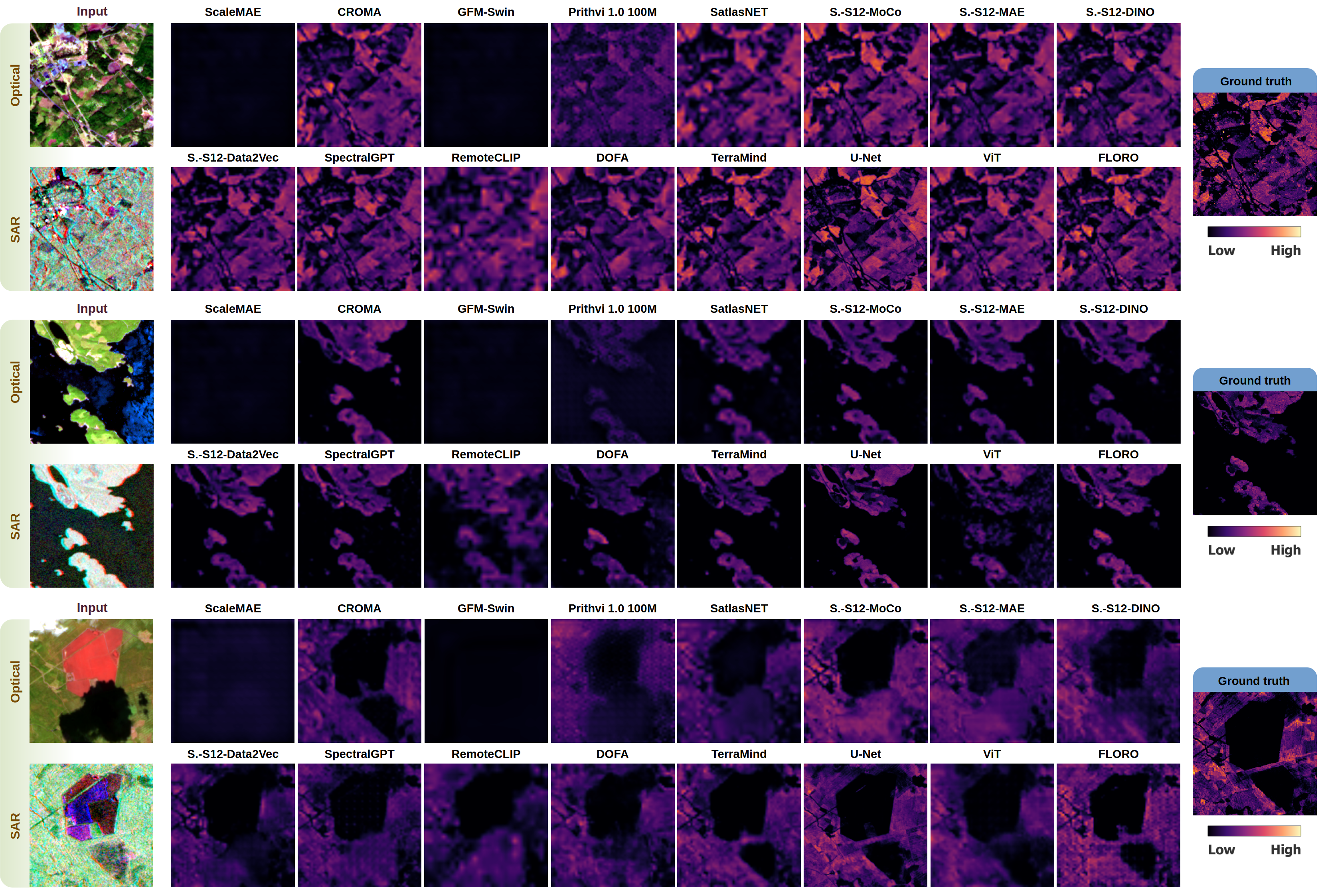}
    \caption{Qualitative comparison of biomass predictions on the Biomassters benchmark. For each example, the optical input, SAR input, reference biomass map, and model predictions are shown for the evaluated methods. Biomassters is characterized by strong parcel geometry, abrupt field boundaries, and heterogeneous within-scene biomass variation, making it a demanding test of structural fidelity in multimodal regression. Scale-MAE and GFM-Swin frequently produce near-uniform low-biomass outputs, whereas RemoteCLIP and Prithvi 1.0 100M show stronger over-smoothing and weaker parcel boundaries. In contrast, stronger methods better preserve parcel organization, transitions between low- and high-biomass areas, and spatial structure under challenging conditions. FLORO produces predictions that are generally closer to the reference maps, particularly in recovering field-scale structure and maintaining plausible biomass patterns where optical information is partially compromised.}
    \label{fig:biomassters_results}
\end{figure}

The Espeletia benchmark provided a more demanding test of localized structural fidelity, as the target variable depends on fine-scale vegetation organization rather than parcel-scale geometry. Across models, performance improved substantially after image tile-level aggregation, indicating that broad canopy-height gradients are easier to recover than exact local magnitudes. Even so, clear differences remained at the pixel level, where FLORO retained better local structural fidelity than many competing methods (\autoref{fig:espeletia_evaluation}). FLORO achieved the strongest pixel-level performance, with $R^2 = 0.912$, RMSE $= 0.234$ m, rRMSE $= 40.9\%$, and $\mathrm{NRMSE}_{P{95}-P_{5}} = 10.3\%$. The U-Net baseline ranked second ($R^2 = 0.851$, RMSE $= 0.306$ m, rRMSE $= 50.3\%$, and $\mathrm{NRMSE}_{P{95}-P_{5}} = 13.2\%$), followed by TerraMind-L in third place ($R^2 = 0.807$, RMSE $= 0.345$ m, rRMSE $= 60.4\%$, and $\mathrm{NRMSE}_{P{95}-P_{5}} = 15.2\%$). After image-tile-level aggregation, FLORO obtained the second-best result, with $R^2 = 0.941$, RMSE $= 0.123$ m, rRMSE $= 21.5\%$, and $\mathrm{NRMSE}_{P{95}-P_{5}} = 7.9\%$; the strongest aggregated result was obtained by the U-Net baseline, with $R^2 = 0.969$, RMSE $= 0.089$ m, rRMSE $= 15.3\%$, and $\mathrm{NRMSE}_{P{95}-P_{5}} = 5.8\%$. 

While quantitative metrics highlight overall performance differences, they do not fully capture how models reconstruct fine-scale canopy structure. \autoref{fig:espeletia_results} provides qualitative context for these differences. Many models capture the broad canopy-height gradient but differ substantially in their ability to preserve compact high-canopy regions, crown-level structure, and coherent local variation. FLORO produces predictions that are generally more faithful to the reference canopy-height model, particularly in preserving compact high-response regions and ecologically meaningful local structure.

To examine whether the aggregate regression scores conceal target-dependent errors, we stratified FLORO's results for the two continuous ecological benchmarks (see details in \autoref{tab:ecological_stratified_errors} in \ref{app:ecological_stratification}). PANGAEA does not natively support stratified evaluation for any dataset, so we adapted its evaluation code to save the prediction and reference rasters for every test image tile. Following the BioMassters protocol, image tiles were assigned to low ($0$-$68.56$ t/px), medium ($68.56$-$112.92$ t/px), and high ($>112.92$ t/px) biomass groups using the published 25th- and 75th-percentile thresholds. Because no externally established ecological canopy-height cutoffs are available for Espeletia CHM, we applied the same quantile procedure to reference image-tile mean canopy height, giving thresholds of $0.228$ and $0.778$ m. Pixel-wise RMSE and signed bias were calculated within each image tile and then summarized across tiles in each group. For BioMassters, mean RMSE increased from $35.175$ t/px in the low group to $51.982$ and $75.240$ t/px in the medium and high groups, while mean bias changed from $+4.739$ to $-4.425$ and $-28.643$ t/px, indicating increasing underestimation at high biomass. For Espeletia CHM, mean RMSE likewise increased from $0.121$ m to $0.195$ and $0.321$ m, while the corresponding biases remained smaller at $-0.012$, $+0.002$, and $-0.034$ m. Thus, FLORO's absolute errors increase with target magnitude in both tasks, with the clearest systematic underestimation occurring in the high-biomass group. This post hoc analysis is limited to BioMassters and Espeletia CHM because their continuous reference rasters permit target-range grouping. It does not establish performance differences among vegetation types, climatic regions, or structural classes not available through the common evaluation protocol.

\begin{figure}[p]
\centering
\includegraphics[width=0.99\linewidth]{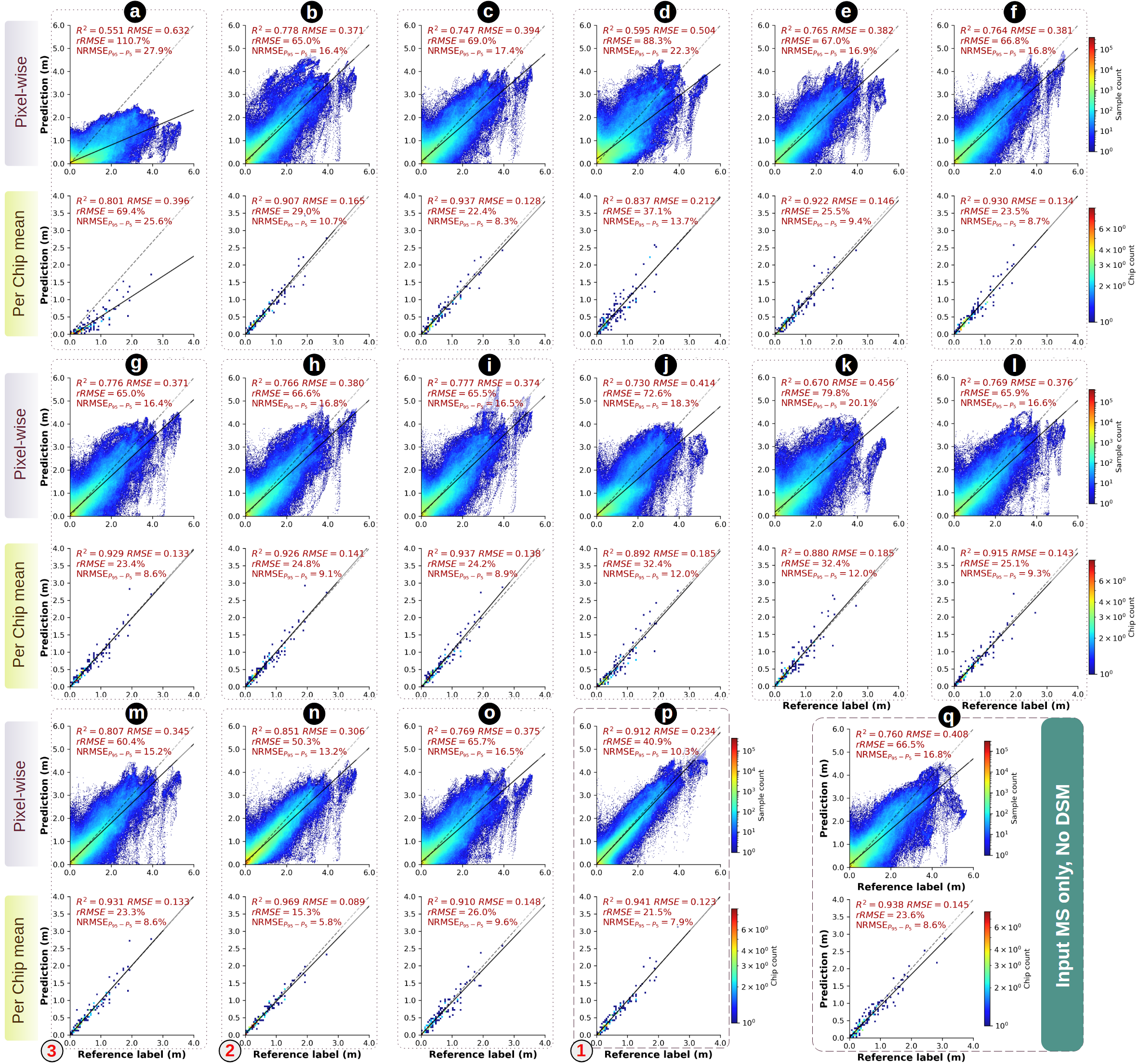}
\caption{Regression performance on the Espeletia canopy-height benchmark under the PANGAEA evaluation framework \citep{Marsocci2026PANGAEA:Benchmark}. For each model, scatterplots compare predicted and reference canopy-height values at the pixel level and after image-tile-level aggregation, summarized using $R^2$, RMSE, and relative RMSE (rRMSE). The evaluated models are: \textbf{(a)} Scale-MAE, \textbf{(b)} CROMA, \textbf{(c)} GFM-Swin, \textbf{(d)} Prithvi 1.0 100M, \textbf{(e)} SatlasNet, \textbf{(f)} SSL4EO-S12-MoCo, \textbf{(g)} SSL4EO-S12-DINO, \textbf{(h)} SSL4EO-S12-MAE, \textbf{(i)} SSL4EO-S12-Data2Vec, \textbf{(j)} SpectralGPT, \textbf{(k)} RemoteCLIP, \textbf{(l)} DOFA, \textbf{(m)} TerraMind-L, \textbf{(n)} U-Net Baseline, \textbf{(o)} ViT Baseline, \textbf{(p)} FLORO, and (q) FLORO evaluated with multispectral input only, after removing DSM. The U-Net, ViT, and TerraMind-L comparisons retain the benchmark's default multispectral-DSM input configuration. \textcircled{\textcolor{red}{\scriptsize 1}}, \textcircled{\textcolor{red}{\scriptsize 2}}, and \textcircled{\textcolor{red}{\scriptsize 3}} indicate the best, second-best, and third-best pixel-wise results, respectively. The figure shows that all models perform better after spatial aggregation than at the pixel level, but important differences remain in local structural fidelity, with FLORO achieving the best pixel-wise performance and remaining highly competitive on image tile-level accuracy.}
\label{fig:espeletia_evaluation}
\end{figure}

\begin{figure}[p]
    \centering
    \includegraphics[width=1\linewidth]{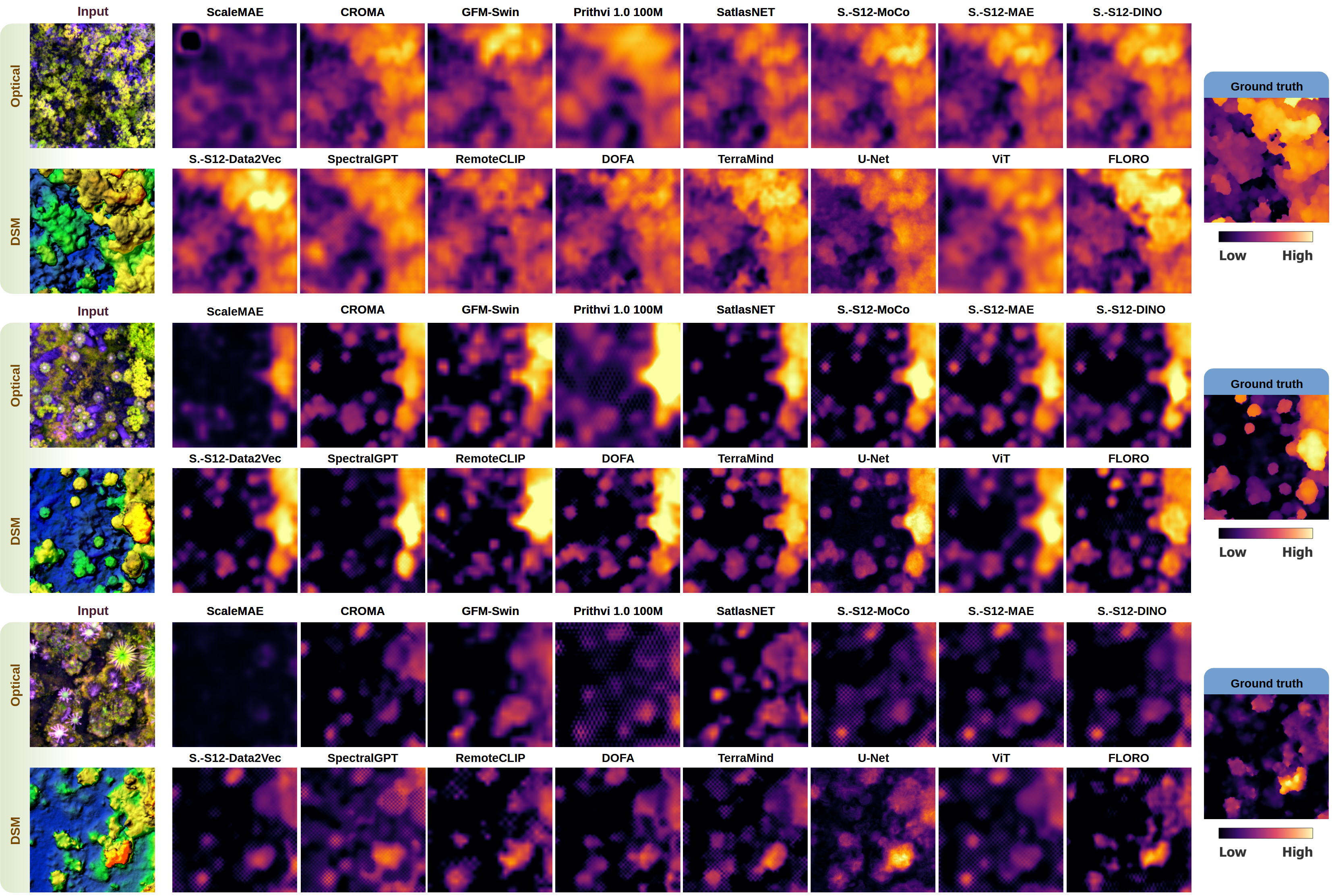}
    \caption{Qualitative comparison of canopy-height predictions on Espeletia. For each sample, the optical input and DSM are shown alongside the reference canopy height model (CHM) and the predictions from the evaluated models. The benchmark emphasizes localized vegetation structure rather than parcel geometry, making it especially informative for assessing the recovery of fine-scale ecological patterns. Many methods reproduce the broad spatial gradient of canopy height but differ markedly in their ability to preserve compact high-canopy regions, crown-level structure, and coherent local variation. Common failure modes include over-smoothing, attenuation of high-response regions, and grid-like artefacts in the predictions. FLORO produces predictions that are generally more faithful to the reference CHM, particularly in retaining compact high-canopy features and ecologically meaningful local structure.}
    \label{fig:espeletia_results}
\end{figure}

Overall, the regression results show that spatial aggregation substantially improves performance, which suggests that many encoders recover the dominant ecological signal at the image tile scale. However, pixel-wise evaluation and qualitative comparisons reveal important differences in local structural fidelity. In Biomassters, these differences are expressed through the preservation of parcel boundaries and biomass gradients under degraded optical conditions, whereas in Espeletia they appear in the recovery of compact high-canopy regions and coherent crown-level variation. FLORO performed among the strongest pretrained models across these settings, achieving balanced pixel- and image tile-level performance on Biomassters and the best pixel-wise performance on Espeletia. These results indicate that its pretrained representations transfer not only to discrete semantic prediction, but also to continuous ecological variables requiring both magnitude calibration and spatially coherent structure.

\subsection{Ablations}
\label{sec:ablations}

To quantify the contribution of specific design choices in FLORO, including pretraining-dataset composition, the masking curriculum, availability and validity channels, geo-positional encoding, and selected auxiliary modalities, we conducted additional ablation experiments in which these components were removed, disabled, replaced, or systematically perturbed. \autoref{fig:ablations} summarizes their effects across the downstream benchmarks, while the corresponding appendix sections provide the exact numerical results and experimental details.

\begin{figure}[p]
    \centering
    \includegraphics[width=1\linewidth]{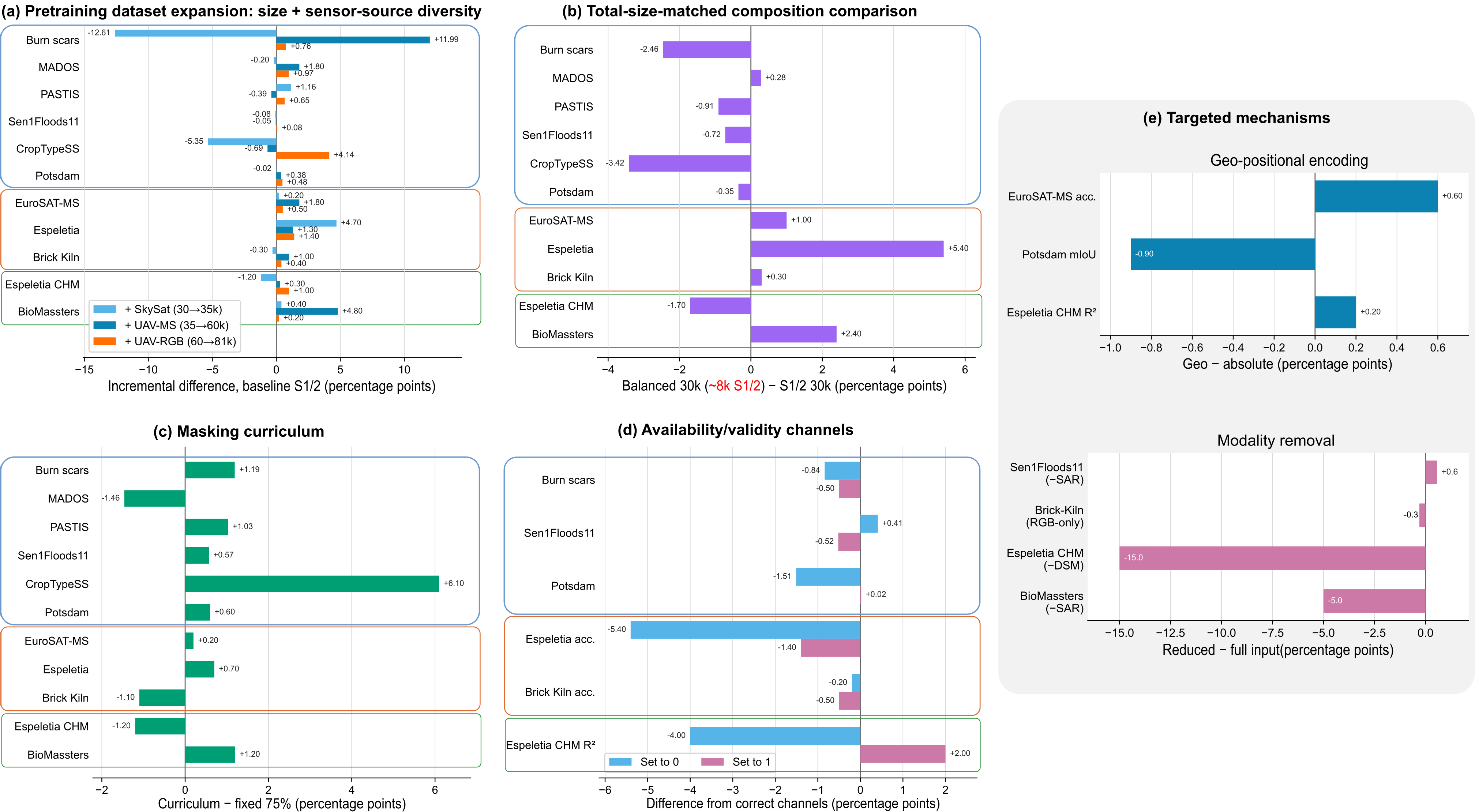}
    \caption{Summary of the single-run ablation effects across the downstream benchmarks. \textbf{(a)} Incremental change after adding SkySat, UAV multispectral (UAV-MS), and UAV RGB data to successively larger pretraining datasets; the sample counts are rounded. \textbf{(b)} Total-size-matched comparison between a balanced multi-sensor 30k-sample subset and the Sentinel-1/2 30k-sample subset. Although both subsets contain approximately 30k samples overall, balancing reduces the Sentinel-1/2 contribution from approximately 30k to approximately 8k samples. Panel (b) is therefore confounded by source-specific sample volume, and performance changes, particularly for downstream datasets that may benefit from Sentinel-1/2 pretraining, cannot be attributed to broader sensor-source coverage alone. The Sentinel-1/2 subset already includes all FLORO spectral groups, elevation, and SAR, so panels (a) and (b) do not represent a monotonic increase in modality coverage. \textbf{(c)} Masking curriculum minus fixed 75\% masking. \textbf{(d)} Change obtained by setting all availability and validity channels to zero or one rather than using their correct values. \textbf{(e)} Targeted mechanism tests: geo-positional minus absolute encoding (top) and reduced-modality minus full-input performance (bottom). Blue, orange, and green outlines group segmentation, classification, and regression benchmarks, respectively. Values are percentage-point (pp) differences; $R^2$ values were multiplied by 100 before differencing. Positive values favor the first configuration named on each horizontal axis. Results are descriptive because one pretrained model was evaluated per configuration and no uncertainty estimates are available.}
    \label{fig:ablations}
\end{figure}

\textbf{Pretraining-dataset composition (\autoref{fig:ablations}a-b).} The Sentinel-1/2 subset already contains the broadest modality coverage, with all FLORO spectral groups plus elevation and SAR, while the SkySat and UAV additions broaden sensors, platforms, resolutions, radiometry, and source domains. The nested expansion produced strongly source- and task-dependent increments. In the total-size-matched comparison, the balanced multi-sensor subset improved all three scene-classification benchmarks, by $+1.0$ pp on EuroSAT-MS, $+5.4$ pp on Espeletia, and $+0.3$ pp on Brick Kiln, as well as MADOS by $+0.28$ pp and BioMassters by $+2.4$ pp. It reduced the other five segmentation benchmarks and Espeletia CHM by $1.7$ pp. However, balancing the subset reduced Sentinel-1/2 data from approximately 30k to approximately 8k samples. The comparison therefore remains confounded by source-specific sample volume, in addition to source identity, geography, resolution, radiometry, and data quality. Performance decreases on downstream datasets that may benefit from Sentinel-1/2 pretraining could partly reflect this reduced exposure rather than a detrimental effect of broader sensor-source coverage. The results show a favorable response to the balanced composition for scene classification and BioMassters, but they do not isolate its cause or establish a general segmentation or regression benefit. Exact scores are reported in \autoref{tab:pretraining_composition_ablation}.

\textbf{Masking curriculum (\autoref{fig:ablations}c).} Relative to fixed 75\% masking, the staged $25\%\rightarrow50\%\rightarrow75\%$ curriculum improved five of six segmentation benchmarks and increased average segmentation mIoU from 63.922\% to 65.262\%. Its largest effect was on CropTypeSS ($+6.10$ pp), whose $32\times32$ inputs require substantial resampling to FLORO's $256\times256$ input, whereas MADOS decreased by 1.46 pp. Classification and regression effects were mixed, including $+0.70$ pp on Espeletia accuracy, $-1.10$ pp on Brick Kiln accuracy, $-1.20$ pp on Espeletia CHM $R^2$, and $+1.20$ pp on BioMassters $R^2$ (\autoref{tab:masking_regime_ablation}).

\textbf{Availability and validity channels (\autoref{fig:ablations}d).} Perturbing the indicator channels produced modest and dataset-dependent segmentation changes, but a clearer response for Espeletia classification: setting the channels to zero or one reduced accuracy by 5.4 and 1.4 pp, respectively. Espeletia CHM behaved differently; setting all channels to zero reduced $R^2$ by 4.0 pp, whereas setting them to one increased it by 2.0 pp when DSM was present. The latter advantage disappeared after DSM removal, indicating an interaction with a strongly predictive structural modality rather than a uniform benefit from forcing the indicators to one (\autoref{tab:availability_validity_ablation} and \ref{tab:chm_dsm_availability_ablation}).

\textbf{Geo-positional encoding (\autoref{fig:ablations}e, top).} In the targeted comparison between FLORO variants using geo-positional and image-grid-only positional encoding, the geo-positional variant changed EuroSAT-MS accuracy and F1 by $+0.6$ pp, Espeletia CHM pixel-level $R^2$ by $+0.2$ pp, and Potsdam mIoU by $-0.92$ pp, from 75.59\% to 74.67\%. These small, mixed, single-run effects do not demonstrate a uniform benefit or statistical significance. Because geographic coordinates may also encode regional priors, we interpret these results together with the footprint-overlap audit in \ref{app:geographic_overlap}. The audit found no positive-area pretraining overlap for Potsdam or Espeletia, but identified overlap for 58 of the 27{,}000 EuroSAT-MS samples (0.215\%). It therefore quantifies direct footprint overlap but cannot exclude broader regional-prior leakage. The early EuroSAT-MS activation maps in \autoref{fig:cams} nevertheless show that geographic coordinates can alter the spatial organization of frozen representations (\autoref{tab:geoposition_ablation}).

\textbf{Targeted modality removals (\autoref{fig:ablations}e, bottom).} Removing SAR slightly increased Sen1Floods11 mIoU ($+0.56$ pp), while using RGB alone reduced Brick Kiln accuracy by 0.30 pp. The ecological regression tasks depended more on their auxiliary structural inputs: BioMassters pixel-level $R^2$ decreased by 5.0 pp without SAR and Espeletia CHM $R^2$ decreased by 15.0 pp without DSM, although both effects were smaller after image-tile aggregation. CropTypeSS further showed an interaction with pretraining composition: full input exceeded SAR removal by 1.354 pp after Sentinel-1/2 pretraining, but not after balanced 30k pretraining ($-0.127$ pp). The detailed metrics and interaction plot are provided in \autoref{tab:targeted_modality_removal} and \autoref{fig:croptypess_sar_interaction}.

\subsection{Cross-benchmark summary}

\autoref{fig:radar_metrics_FLORO} summarizes the reported downstream results without replacing the task-specific analyses above. Panel (a) shows each model's score relative to the best score on each of the six segmentation, three classification, and two regression benchmarks. Panel (b) plots average normalized performance against reported pretraining-dataset size for the four parameter-comparable models, namely FLORO, Scale-MAE, CROMA, and TerraMind-L, whose evaluated encoder-wrapper sizes are approximately between 303 and 307 million parameters.

\begin{figure}[H]
    \centering
\includegraphics[width=0.7\linewidth]{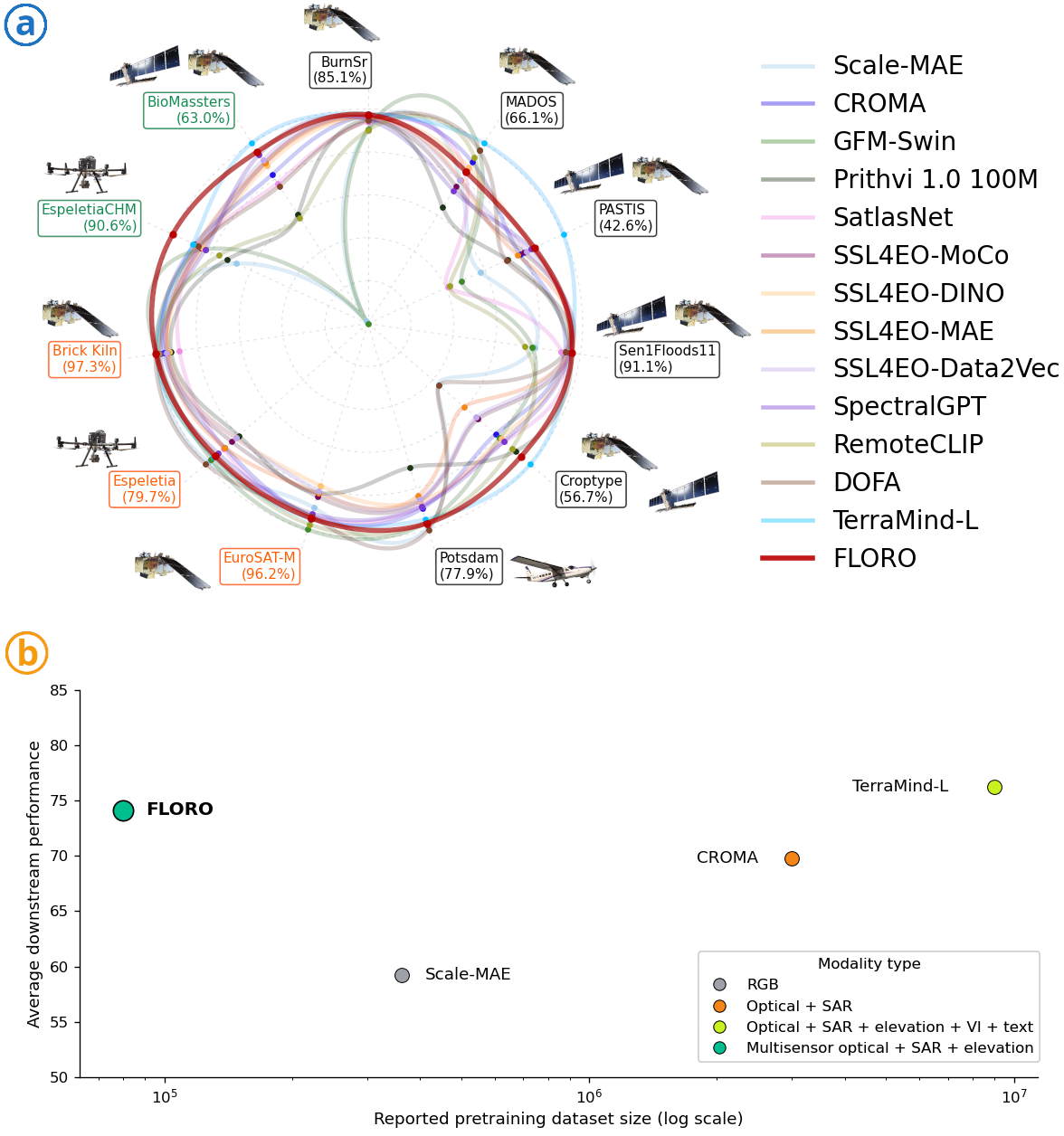}
    \caption{Cross-benchmark summary comparison of the evaluated foundation models. \textbf{(a)} Benchmark-wise relative performance of all evaluated foundation models, normalized to the best result obtained for each dataset. The comparison spans six segmentation benchmarks (black label), three scene-classification benchmarks (orange label), and two regression benchmarks (green label). Percentages shown next to each dataset indicate the best observed score used as the normalization reference. The benchmark suite covers optical-only, optical-SAR, and optical-elevation settings across satellite, airborne, and UAV data. \textbf{(b)} Average normalized downstream performance plotted against reported pretraining dataset size on a logarithmic scale for the parameter-comparable models only: FLORO, Scale-MAE, CROMA, and TerraMind-L. Their evaluated encoder-wrapper sizes are approximately 303-307 million parameters. Marker color denotes pretraining modality composition. Restricting this panel by model size reduces capacity as a confounding factor, although architecture, training objective, dataset composition, and dataset quality still differ among models.}
    \label{fig:radar_metrics_FLORO}
\end{figure}

\section{Discussion}
\label{sec:discussion}

Our findings indicate that  FLORO achieves strong, stable transfer performance across a highly heterogeneous benchmark suite. FLORO ranked second on average across the segmentation benchmarks, achieved the third-best average rank across three scene-classification benchmarks under a deliberately lightweight linear evaluation protocol, and performed strongly on both biomass estimation and canopy-height regression. Relative to existing geospatial foundation models, FLORO contributes a distinct combination of design choices rather than a single isolated architectural novelty. Its main methodological difference is the integration of availability-aware multimodal input modeling with cross-scale heterogeneous pretraining across satellite and UAV-derived data. The design targets a practical limitation of ecological remote sensing, where data often differ not only in scale and radiometry, but also in which spectral groups and auxiliary modalities are available \citep{Cavender-Bares2022IntegratingConservation}. The cross-benchmark results show that a large-capacity encoder can learn competitive representations from a relatively small number of pretraining samples (approximately 81{,}000 in this study), and transfer them across classification, segmentation, and regression tasks. This is evidence of data-efficient transfer with respect to pretraining data volume, not evidence of a small model, low inference cost, or a universal benefit from increasing sensor-source diversity.

To summarize model behavior across tasks and sensing regimes, \autoref{fig:radar_metrics_FLORO} provides a benchmark-normalized comparison of downstream transfer performance. The benchmark suite is intentionally heterogeneous, spanning medium-resolution satellite imagery, high-resolution airborne data, and ultra-high-resolution UAV products, as well as optical-only, optical-SAR, and optical-elevation settings. Accordingly, \autoref{fig:radar_metrics_FLORO}a should be interpreted as a cross-domain synthesis of transfer behavior rather than as a comparison under a single standardized input regime. Despite this heterogeneity, FLORO remains consistently close to the best-performing model across most benchmarks. This pattern does not indicate uniform dominance on every dataset, but rather stable transfer across markedly different task types, scene structures, and sensing configurations. The consistency is especially notable given the comparatively small size of the FLORO pretraining dataset relative to several competing foundation models.

Average normalized downstream performance is further related to reported pretraining dataset size in \autoref{fig:radar_metrics_FLORO}b. To reduce confounding by model capacity, this panel is restricted to FLORO, Scale-MAE, CROMA, and TerraMind-L, whose evaluated encoder-wrapper parameter counts are closely matched, approximately 303-307 million parameters under the profiling protocol in \autoref{tab:encoder_parameter_counts}. The comparison therefore contrasts pretraining data scale and modality composition among similarly sized models, rather than implying that every model in panel (a) has the same capacity. It still does not treat dataset size as the only determinant of transfer performance, because the selected models differ in architecture, pretraining objective, modality coverage, data sources, and data quality. Within this restricted comparison, FLORO occupies a favorable region of the trade-off between downstream performance and pretraining-data scale. It achieves strong average transfer performance while using a substantially smaller, heterogeneous pretraining dataset. This distinction matters in practice because, for many ecological and environmental applications, assembling, curating, storing, and training on massive pretraining datasets can be difficult \citep{Koldasbayeva2024ChallengesPractice,Yu2025MachineAnalysis}. The result therefore motivates further study of representation learning at modest pretraining-data scales, without implying that heterogeneous source composition alone caused FLORO's transfer performance.

The combined results support a central conclusion: FLORO learned competitive representations from a comparatively small pretraining dataset and transferred them effectively across heterogeneous downstream tasks. The ablations do not identify sensor-source diversity as the general cause of this behavior. In the total-size-matched control, the balanced composition improved all three scene-classification benchmarks and BioMassters, but reduced five of six segmentation benchmarks and Espeletia CHM. However, balancing also reduced Sentinel-1/2 data from approximately 30k to approximately 8k samples, so the comparison remains confounded by source-specific sample volume. The Sentinel-1/2 subset also had the broadest modality coverage, so the comparison tests changes in source proportions, sensors, platforms, resolutions, and source domains rather than a simple increase in multimodality. The evidence therefore supports data-efficient cross-task transfer and task-dependent source selection, but does not isolate the effect of sensor-source diversity. The expanded positional-encoding ablation likewise indicates small, task-dependent changes rather than a uniform benefit from geographic grounding.

The observed behavior of FLORO is also consistent with its architectural design. In particular, the use of validity and availability channels allows the model to operate within a unified input space even when spectral bands and auxiliary modalities differ across sensors. Its objective is to use whatever valid observations are available while remaining operational when particular spectral groups or auxiliary data are absent. The model therefore distinguishes between two forms of missing information. Patches deliberately hidden from an observed input are reconstructed as part of masked-autoencoder pretraining, whereas genuinely unavailable channels and invalid pixels are identified by the availability and validity indicators and excluded from the corresponding grouped reconstruction loss. Thus, FLORO retains masked reconstruction for observed data but does not require the model to generate entire modalities that were never acquired. This choice reflects a practical remote-sensing workflow in which analyses are adapted to the available observations rather than relying on generated substitutes for missing measurements \citep{Schrodt2020IntegratingScales}. Such flexibility is important in remote sensing, where datasets rarely share identical spectral definitions or spatially compatible modality combinations \citep{Li2022DeepReview,Samadzadegan2025AProspects}. Performance is nevertheless expected to change when informative inputs are removed because optical bands, SAR, and elevation provide different and potentially non-redundant information. The desired behavior is therefore graceful, task-dependent degradation rather than invariance to input removal. The channel-perturbation ablation showed its clearest effect on Espeletia classification, while segmentation effects were modest and dataset-dependent and the CHM response depended strongly on elevation-related structural information. 

The masking-regime ablation showed that curriculum pretraining primarily benefited dense segmentation, with its clearest gain on the severely resampled CropTypeSS input; the advantage on Potsdam's airborne UltraCamXp imagery was comparatively small. The channel-perturbation ablation showed its clearest effect on Espeletia scene classification, while segmentation effects were modest and dataset-dependent and the CHM response depended strongly on elevation-related structural information. The targeted modality-removal tests further showed that auxiliary inputs play scale- and task-dependent roles. Removing SAR slightly improved Sen1Floods11 and produced a pretraining-dependent response on CropTypeSS, whereas RGB-only Brick Kiln changed little. Removing SAR from BioMassters caused a modest decline at both evaluation scales, while removing DSM from Espeletia CHM caused a marked reduction in pixel-level accuracy but only a small change after image-tile aggregation. These findings indicate that availability-aware modeling is useful but task- and modality-dependent: validity masks should be constructed carefully when auxiliary structural inputs are consistently available, spatially aligned, and highly predictive of a continuous target. The positional-encoding ablation likewise produced small, mixed effects across classification, segmentation, and regression. Taken together, these results support FLORO's flexible multimodal input handling while cautioning against interpreting any single component as uniformly beneficial across tasks.

At the same time, the results also highlight important benchmark-dependent limitations. FLORO was not the best-performing model across all datasets, and some benchmarks are more naturally aligned with certain competing architectures. For example, the CropTypeSS dataset \citep{Rustowicz2019SemanticMethods} is composed of very small $32 \times 32$ image tiles, requiring substantial resampling for FLORO, whose native input size is $256 \times 256$. The mismatch is less severe for several competing models that operate with smaller native input sizes, such as GFM-Swin ($192 \times 192$) and the SSL4EO MoCo and DINO variants ($224 \times 224$). Scene-classification performance varied across datasets: GFM-Swin remained strongest on EuroSAT-MS, DOFA achieved the highest Espeletia accuracy and Brick Kiln scores, and FLORO ranked third for Espeletia F1 and second for both Brick Kiln metrics. Together with FLORO's stronger segmentation and regression performance, these findings suggest that spatially structured prediction is a relative strength of the model. Nevertheless, its competitive scene-classification results show that the learned representations also support semantic scene-level prediction. These considerations do not weaken the value of the benchmark: rather, they emphasize that cross-model comparison in remote sensing must be interpreted in the context of differences in native resolution, modality assumptions, decoder complexity, and dataset characteristics.

The qualitative results further reinforce this interpretation. In segmentation, FLORO produced predictions that were generally more spatially coherent and topologically faithful across different target types, particularly for elongated structures, connected flood regions, and urban boundaries. In the regression tasks, FLORO more effectively preserved parcel-level organization in Biomassters and local canopy structure in Espeletia, while avoiding some of the over-smoothing and artifact patterns observed in other models. These qualitative differences are important because they indicate that strong transfer is not only reflected in aggregate metrics, but also in the recovery of ecologically and geographically meaningful spatial structure: something particularly relevant for ecological applications, where downstream usefulness often depends as much on structural fidelity as on average numerical agreement \citep{Ploton2020SpatialModels,Wolff2025InterannualCommunities}.

More broadly, our work contributes to the ongoing discussion on how remote sensing foundation models should be designed and evaluated \citep{Zhu2026OnModels}. The present results motivate closer study of how much transferable representation quality can be obtained from modest pretraining-data volumes, especially when models must handle partial and heterogeneous inputs. For ecological and environmental monitoring, this is an encouraging result, since assembling extremely large pretraining datasets and accessing large-scale compute may be impractical. FLORO serves as a multimodal geospatial foundation model and as a case study of competitive cross-task transfer from a comparatively small pretraining dataset. The composition ablations further show that source selection should be evaluated by downstream task family rather than summarized as a universal diversity benefit.

Several directions remain for future work. First, although frozen-encoder evaluation is useful for isolating representation quality, it does not fully capture the performance ceiling attainable through end-to-end adaptation \citep{Marti-Escofet2025Fine-tuneModels}. Second, the current benchmark framework necessarily relies on standardized decoders, which is appropriate for comparison but may understate the potential of some encoders under task-optimized downstream heads \citep{Chen2026LightFormer:Segmentation}. Third, the present pretraining setup does not yet include additional sources such as text supervision or broader temporal context, both of which may further improve transfer \citep{Jakubik2025TerraMind:Observation}. The current experiments also do not independently quantify dominance or representation collapse within the shared latent representation, robustness to controlled spatial misregistration or resolution changes, or stability across a systematic range of missing-modality patterns. These questions require targeted perturbation studies and representation-level diagnostics and are left for future research. Additional architecture-level controls also remain open. A shared-mask variant requires redefining how masks are coupled across streams under heterogeneous modality availability. A pretraining decoder variant with a single shared transformer for masked reconstruction requires a new common output parameterization for spectral and auxiliary targets, while naive validity handling and alternative reconstruction weighting require new loss definitions and the evaluation of multiple weighting schemes. Each alternative changes the pretraining architecture or objective and requires a complete pretraining and downstream-validation workflow, so these comparisons are treated as future research rather than simple component-removal ablations. Future work should therefore examine how FLORO behaves under controlled fine-tuning, whether its availability-aware design remains advantageous in more specialized adaptation settings, and how it can be extended to ecological applications that require finer semantic granularity, such as species-level vegetation analysis, carbon mapping, or long-term ecosystem monitoring.

A further consideration concerns the evaluation of FLORO's geo-positional branch. The geo-positional branch should therefore be regarded as exploratory rather than as an established source of improved transfer. Its small and task-dependent effects indicate that explicit location information may provide useful context in some settings but can also introduce regional priors that do not generalize across datasets. The audit in \ref{app:geographic_overlap} addresses direct footprint duplication and proximity, but it cannot exclude broader spatial dependence: a small number of EuroSAT-MS samples overlap the pretraining data, and the Espeletia evaluation tiles are adjacent to downstream training tiles. Moreover, the single-run experiments do not establish statistical significance or robustness to coordinate, projection, and orthomosaic uncertainty. Future work should exclude overlapping samples, use spatially disjoint and cross-region evaluations, repeat pretraining across random seeds, and perturb geographic metadata to determine whether the observed effects are reproducible and genuinely attributable to location encoding.

A separate limitation concerns the environmental and geographic coverage of the pretraining data. The pretraining dataset was not constructed as a balanced representation of dense urban, industrial, and polar environments. The Potsdam experiment provides evidence of transfer to one dense-urban benchmark and one naturalistic out-of-distribution evaluation involving imagery acquired with an airborne UltraCamXp camera system not represented during pretraining, where FLORO obtained 75.595\% mIoU \citep{ISPRSISPRSDataset}. Brick Kiln Classification \citep{Lee2021ScalableCapacity} provides a narrower test involving detection of an industrial activity, where FLORO obtained 97\% accuracy and 0.966 F1. These results support transfer to two underrepresented settings but do not establish generalization across the wider diversity of urban and industrial landscapes. No polar benchmark was evaluated, and the increasing high-latitude distortion of the Web Mercator coordinates used by the geo-positional branch introduces an additional limitation for polar applications. Generalization to polar environments therefore remains unknown and requires dedicated high-latitude pretraining data, suitable coordinate representations, and geographically independent downstream evaluation.

\section{Conclusion}
\label{sec:conclusion}

Under the frozen-encoder evaluation protocol of the PANGAEA benchmark, FLORO demonstrated strong and stable transfer performance across scene classification, semantic segmentation, and regression tasks. In segmentation, it remained among the top-performing pretrained models across six highly heterogeneous benchmarks spanning optical, optical-SAR, and optical-elevation settings. In scene classification, FLORO achieved the third-best average rank across EuroSAT-MS, Espeletia, and Brick Kiln Classification under a deliberately lightweight linear evaluation protocol, indicating that its encoder learned scene-discriminative representations that can support useful downstream classification with limited task-specific training. In regression, FLORO showed strong performance on both biomass estimation and canopy-height reconstruction, while qualitative results indicated improved preservation of parcel structure, local canopy organization, and ecologically meaningful spatial detail. Targeted auxiliary-modality removals showed that multispectral-only FLORO retained useful transfer performance: SAR provided a modest benefit for biomass estimation, whereas DSM was especially important for dense canopy-height reconstruction but had little effect on image-tile-level canopy-height means. Taken together, these results support effective representation learning and competitive cross-task transfer from a comparatively small pretraining dataset. 

\section*{Acknowledgments}
The authors acknowledge support from the King Abdullah University of Science and Technology and the KAUST Center of Excellence in Generative Artificial Intelligence.

\section*{Code and Data Availability}

The code implementing the FLORO framework, including data preprocessing routines, model architecture, training scripts, and evaluation pipelines, is publicly available at \url{https://github.com/jorlrodriguezg/floro} under the MIT License. The repository includes an integration with the PANGAEA evaluation framework, through which the downstream benchmark datasets used in this study can be downloaded and the reported evaluations reproduced. Pretrained model checkpoints are publicly available at \url{https://huggingface.co/jorlrodriguezg/floro}, and additional documentation is provided at \url{https://jorlrodriguezg.github.io/floro/}.

The pretraining corpus comprises four subsets: (i) Sentinel-1 and Sentinel-2 imagery paired with Copernicus elevation data; (ii) commercially acquired SkySat imagery paired with the publicly available IGN RGE ALTI 1 m digital elevation model of France, resampled to 50 cm; (iii) UAV multispectral imagery acquired by the research team; and (iv) UAV RGB imagery acquired by the authors.

The assembled pretraining corpus is not publicly hosted due to its size and applicable redistribution restrictions. Shareable components may be made available by the corresponding author upon reasonable request, subject to licensing and data-sharing permissions. Sentinel and Copernicus products are accessible through their original data providers, and the IGN elevation data are publicly available at \url{https://geodatafr.github.io/IGN/RGE_Alti_1m_HR_Digital_elevation_model/}. Commercially acquired SkySat imagery cannot be redistributed under the applicable licensing conditions.

\newpage
\appendix
\section{Pretraining data preparation}
\label{app:ap1}

The pretraining dataset combines medium-resolution multisensor satellite observations with high-resolution satellite and UAV data. The intent is to expose the model to a broad range of spatial resolutions, viewing geometries, and sensor characteristics while retaining a consistent tensor layout for masked reconstruction. The dataset includes: (i) global Sentinel-2 surface reflectance and Sentinel-1 backscatter retrieved through Google Earth Engine (GEE) \citep{Gorelick2017GoogleEveryone}, (ii) high-resolution SkySAT imagery paired with 1\,m terrain products from France, produced by the Institute National De L'Information Géographique et Forestière (IGN), (iii) UAV RGB orthomosaics acquired with a fixed-wing (flying wing) platform, and (iv) UAV multispectral orthomosaics acquired using a DJI Matrice 100 equipped with a MicaSense sensor.

Across all platforms, preprocessing follows three shared principles. First, each data source is converted to physically interpretable units whenever possible (e.g., Sentinel-2 surface reflectance, Sentinel-1 backscatter in dB). Second, all data are spatially subdivided into fixed-size image tiles to enable scalable pretraining and consistent batching. Third, image tile extraction preserves co-registration among modalities available for the same scene (e.g., optical with terrain) and records validity masks when observations are incomplete (e.g., cloud masking or nodata regions).

\subsection{Sentinel-1/2 global data acquisition and preparation}

Sentinel-1 and Sentinel-2 data were acquired globally using workflows implemented in Google Earth Engine (GEE) \citep{Gorelick2017GoogleEveryone}. Sentinel-2 surface reflectance was retrieved from the harmonized surface-reflectance collection and cloud-masked using the corresponding cloud-probability product. Pixels with cloud probability greater than or equal to the selected threshold were masked before compositing. Sentinel-1 backscatter was retrieved from GRD products in dual-polarization VV/VH mode, composited over the same temporal windows as Sentinel-2, and converted to decibels as
\[
\sigma^{0}_{\mathrm{dB}} = 10 \log_{10}\left(\sigma^{0}_{\mathrm{linear}}\right).
\]

where $\sigma^{0}_{\mathrm{linear}}$ is the radar cross-section per unit area (dimensionless ratio) and $\sigma^{0}_{\mathrm{dB}}$ is the same value expressed in decibels (dB), which compresses the wide dynamic range of radar data for easier analysis.

To promote global diversity in land-cover conditions, image tile centers were selected using a stratified sampling design based on ESA WorldCover classes and latitude bands. Sampling locations were exported as Earth Engine assets with stored metadata, including the temporal window, latitude band, land-cover class, and random key. This allowed deterministic reuse of the same sampling locations across export batches and preprocessing revisions.

For each sampled location, square regions were exported as co-registered Sentinel-2/Sentinel-1 stacks at 10\,m spatial resolution. Sentinel-2 bands B2, B3, B4, and B8 were used at native 10\,m resolution, while B5, B8A, B11, and B12 were resampled from 20\,m to 10\,m. Sentinel-2 reflectance values were rescaled by \(10000\), and Sentinel-1 VV and VH backscatter were stored in decibel units. Exports were written as Cloud-Optimized GeoTIFFs and subsequently used for offline image tile extraction. When external elevation data were included, the DEM was warped to the exported image tile grid and inserted into a fixed auxiliary band position to preserve the common FLORO tensor layout.

\subsection{High-resolution SkySAT imagery and 1 m terrain data (France)}

High-resolution SkySAT imagery is incorporated to expose the model to fine spatial detail and textural variability not present in medium-resolution satellite observations. For SkySAT scenes, the preprocessing pipeline (i) converts imagery to a consistent radiometric representation (using the provider-delivered calibrated products where available), (ii) spatially aligns imagery with corresponding 1\,m terrain products for the same region in France, and (iii) extracts paired image tiles from the optical and terrain layers using identical spatial windows to preserve pixel-level correspondence. Because spatial resolution differs substantially from Sentinel-2, image tile sizes are selected to maintain a comparable physical footprint where feasible, while keeping the tensor layout consistent at the model input stage via standardized resizing or patchification during training.

\subsection{UAV RGB data (fixed-wing platform)}

UAV RGB orthomosaics acquired with a flying wing platform were included to capture fine-scale structure and local land-surface heterogeneity. Preprocessing comprised: (i) orthomosaic generation with geometric consistency checks, (ii) masking of nodata regions introduced during mosaicking, and (iii) extraction of fixed-size image tiles either on a regular grid or using a sampling design matched to the study-area extent. Digital surface models (DSMs) were produced as part of the same photogrammetric adjustment used to generate the orthomosaic, and digital terrain models (DTMs) were subsequently derived from the DSM products. Because the orthomosaic and terrain layers shared a common georeferencing solution and were exported on the same spatial grid, no additional co-registration was required. Image tiles were extracted using identical spatial windows across RGB and terrain modalities.

\subsection{UAV multispectral data (DJI Matrice 100 with MicaSense)}

UAV multispectral orthomosaics acquired using a DJI Matrice 100 equipped with a MicaSense sensor are incorporated to provide high-resolution spectral information complementary to both Sentinel-2 and RGB UAV imagery. Platform-specific preprocessing includes radiometric calibration to reflectance, band alignment across the multispectral channels, and orthomosaic generation. Image tiles are extracted from the multispectral orthomosaics using the same windowing strategy as other high-resolution datasets. The orthomosaic and terrain layers were produced within the same photogrammetric workflow and exported on a common grid, so additional co-registration was not required beyond standard resampling to the image tile grid.

\subsection{Tensor harmonization across data sources}

After source-specific preprocessing, all samples were converted to the common FLORO input representation used during masked-autoencoding pretraining. Optical observations were mapped to the available spectral groups, while auxiliary information such as SAR and elevation was mapped to the corresponding modality slots when present. Validity and availability indicators were used to distinguish observed, missing, and invalid inputs. This harmonization allowed Sentinel-1/2, SkySAT-terrain, UAV multispectral, and UAV RGB samples to be combined within a unified pretraining framework despite differences in spatial resolution, spectral coverage, radiometric calibration, and modality availability. During pretraining, band and modality dropping were additionally applied to simulate incomplete sensor configurations, exposing the model to cases in which only a subset of spectral groups or auxiliary modalities was available.

\section{Model scale and inference cost for the foundation model variants evaluated through PANGAEA}\label{app:model_scale}

To support interpretation of the cross-benchmark results in \autoref{fig:radar_metrics_FLORO}b, we report the parameter count and per-sample inference compute of each encoder wrapper evaluated through PANGAEA. All variants were profiled on EuroSAT-MS with batch size 1, using the ATen backend, each model's required optical input shape, and a lightweight linear classification head. FLOPs are estimated as twice the multiply-accumulate count. Because the models use different input resolutions and numbers of spectral channels, these measurements represent their instantiated use in PANGAEA rather than compute normalized to a common input configuration. They do not measure latency, memory consumption, energy use, or pretraining compute.

\begin{table}[H]
\centering
\caption{Reported pretraining scale for FLORO and TerraMind-L, the closest parameter-matched variant evaluated through PANGAEA. Token counts follow each model's reporting convention. Aggregate GPU-hours are the unadjusted product of training duration and accelerator count and are not normalized for accelerator throughput, numerical precision, utilization, or energy consumption.}
\label{tab:closest_model}
\resizebox{\linewidth}{!}{%
\begin{tabular}{l l l l l r}
\toprule
\textbf{Model} & \textbf{Backbone scale} & \textbf{Training-token reporting} & \textbf{Schedule} & \textbf{Hardware} & \textbf{Aggregate GPU-hours} \\
\midrule
TerraMind-L & Large & 500B training tokens & 10 days & 32 NVIDIA A100 & 7{,}680 A100 GPU-h \\
FLORO & ViT-L & 20M (20{,}808{,}192) per pass; 3B (3{,}121{,}228{,}800) across 150 epochs, before masking & 2.5 days & 8 NVIDIA V100 & 480 V100 GPU-h \\
\bottomrule
\end{tabular}}
\end{table}

\begin{table}[H]
\centering
\small
\setlength{\tabcolsep}{4pt}
\renewcommand{\arraystretch}{1.10}
\caption{Model scale and inference cost for the variants evaluated through PANGAEA. Parameter counts refer to the instantiated encoder wrappers and exclude pretraining-only and downstream modules. Inference cost was profiled on EuroSAT-MS with batch size 1 using the ATen backend, each model's required optical input shape, and a lightweight linear classifier. FLOPs are estimated as twice the multiply-accumulate count.}
\label{tab:encoder_parameter_counts}
\resizebox{\textwidth}{!}{%
\begin{tabular}{P{4.2cm} c r r}
\toprule
\textbf{Model} &
\shortstack{\textbf{Optical input}\\\textbf{shape ($C\times H\times W$)}} &
\shortstack{\textbf{Encoder-wrapper}\\\textbf{parameters (M)}} &
\shortstack{\textbf{Inference cost}\\\textbf{(GFLOPs/sample)}} \\
\midrule
CROMA & $12\times120\times120$ & 303.025 & 141.293 \\
DOFA & $4\times224\times224$ & 111.312 & 34.167 \\
GFM-Swin & $3\times192\times192$ & 86.725 & 22.595 \\
Prithvi & $6\times224\times224$ & 86.387 & 33.960 \\
RemoteCLIP & $3\times224\times224$ & 87.454 & 5.898 \\
SatlasNet & $9\times128\times128$ & 87.943 & 11.116 \\
Scale-MAE & $3\times224\times224$ & 303.098 & 119.380 \\
SpectralGPT & $12\times128\times128$ & 85.402 & 213.075 \\
SSL4EO-S12 MoCo & $13\times224\times224$ & 22.648 & 8.884 \\
SSL4EO-S12 DINO & $13\times224\times224$ & 22.635 & 9.593 \\
SSL4EO-S12 MAE & $13\times224\times224$ & 22.648 & 8.884 \\
SSL4EO-S12 Data2Vec & $13\times224\times224$ & 22.582 & 9.599 \\
TerraMind-L & $12\times224\times224$ & 305.401 & 123.370 \\
FLORO & $8\times256\times256$ & 307.343 & 162.920 \\
ViT baseline & $13\times224\times224$ & 87.765 & 34.499 \\
\bottomrule
\end{tabular}
}
\end{table}
It is important to note that parameter count and inference compute are not interchangeable measures of model scale. TerraMind-L is the closest evaluated variant to FLORO in encoder-wrapper size, with 305.401M and 307.343M parameters, respectively, a difference of approximately 0.6\%. Under the common profiling protocol, FLORO requires 162.920 GFLOPs per sample, compared with 123.370 GFLOPs for TerraMind-L, or approximately 1.32 times as much compute. The wider variation in FLOPs among models with similar parameter counts further reflects differences in architecture, input dimensions, and tokenization.

\section{Geographic overlap audit for georeferenced downstream datasets}
\label{app:geographic_overlap}

We audited the three downstream datasets for which geotransform metadata were retained, EuroSAT-MS, Potsdam, and Espeletia, against all 81{,}282 pretraining footprints. Each raster extent was represented as a polygon. Positive-area intersection and edge-to-edge distance were measured in the local WGS84 UTM zone of each downstream footprint. Boundary contact without shared area was recorded separately from positive-area overlap, and proximity was summarized at 10~m, 100~m, 1~km, and 10~km thresholds.

EuroSAT-MS contained 58 positive-area overlaps among 27{,}000 samples (0.215\%); 41 were at least 99.9\% covered, and every overlapping pretraining footprint came from the Sentinel-1/Sentinel-2/elevation source group. The counts within 10~m, 100~m, 1~km, and 10~km were 59, 62, 71, and 303, respectively. Potsdam had no positive-area overlap and none of its 38 footprints was within 10~km of a pretraining footprint. Espeletia also had no positive-area overlap; its 985 samples were approximately 3.62-3.96~km from the nearest UAV-RGB/DSM pretraining footprint. In a separate downstream-split audit, none of the 296 Espeletia validation or test tiles shared positive area with a downstream training tile, but all touched at least one training-tile boundary. This adjacency means that the Espeletia split should not be interpreted as a spatially disjoint regional test.

\begin{figure}[htbp]
\centering
\includegraphics[width=\textwidth]{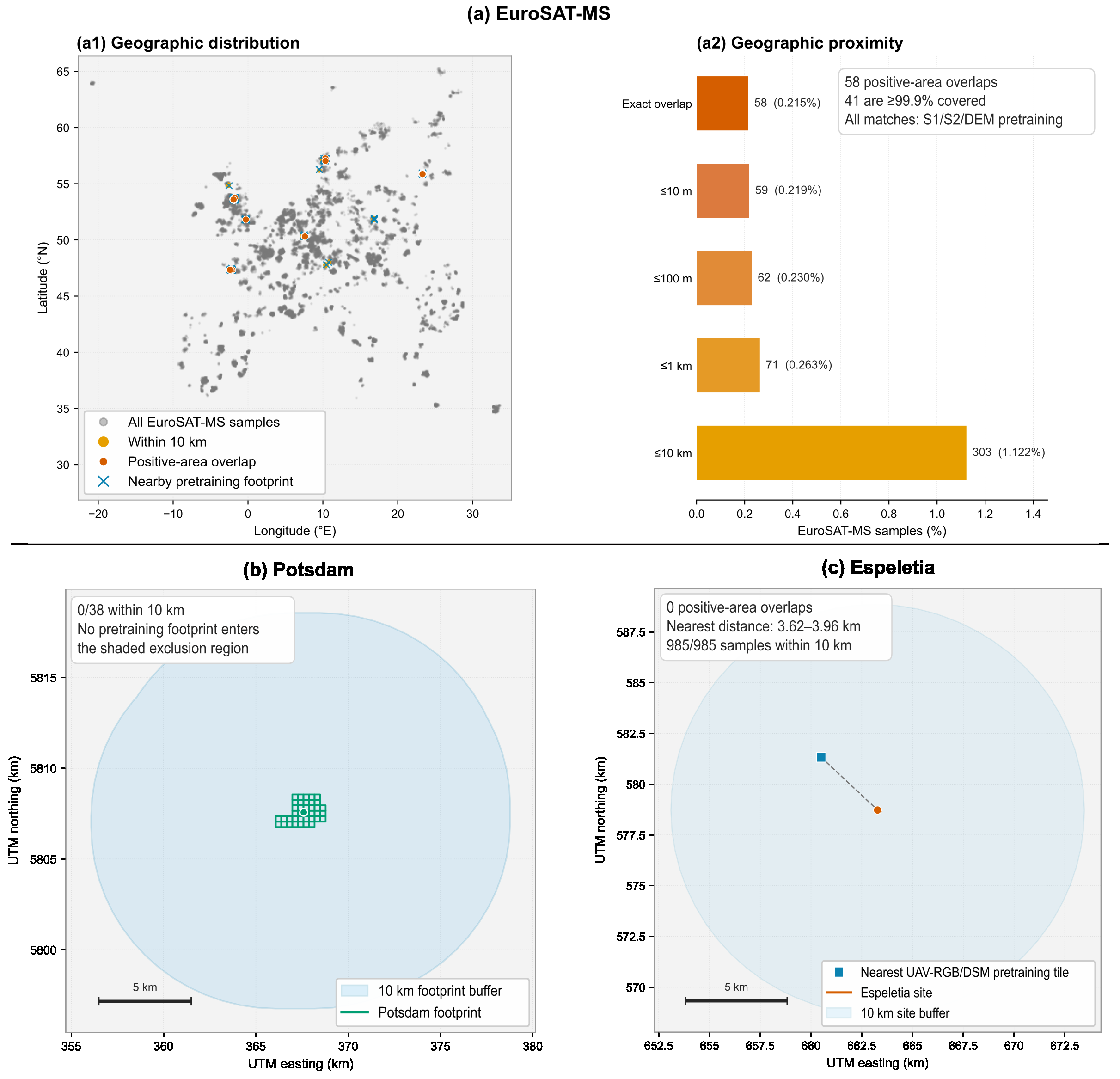}
\caption{Geographic overlap audit for the three downstream datasets with retained geotransform metadata. \textbf{(a)} EuroSAT-MS, with \textbf{(a1)} the geographic distribution of all samples and nearby or overlapping pretraining footprints and \textbf{(a2)} cumulative proximity counts. Of 27{,}000 samples, 58 (0.215\%) have positive-area overlap with a pretraining footprint, and all such matches are from the Sentinel-1/Sentinel-2/elevation source group. \textbf{(b)} Potsdam, where none of the 38 footprints overlaps or lies within 10~km of a pretraining footprint. \textbf{(c)} Espeletia, where no downstream sample overlaps pretraining and the nearest UAV-RGB/DSM pretraining footprint is approximately 3.62-3.96~km away. Intersections and edge-to-edge distances were computed using polygon footprints in the local WGS84 UTM zone of each downstream sample. The audit quantifies direct footprint overlap and proximity, but does not by itself eliminate broader regional-prior leakage.}
\label{fig:geographic_overlap_audit}
\end{figure}

\newpage

\section{Stratified target-range error analysis}
\label{app:ecological_stratification}

PANGAEA reports aggregate evaluation metrics and does not natively support stratification for any downstream dataset. To enable this post hoc analysis, we adapted its evaluation code to save the prediction and reference rasters for every test image tile. We then performed target-range stratification for BioMassters biomass and Espeletia canopy height because their continuous reference rasters permit grouping by image-tile mean target value. The remaining downstream datasets use discrete scene or segmentation labels and do not provide a common vegetation-type, dryland, biome, or structural-complexity variable under the benchmark protocol. We therefore restrict the analysis to these two regression tasks rather than assigning unsupported ecological groups. Following the BioMassters procedure, each image tile contributes one pixel-wise RMSE and one mean signed error, prediction minus reference. Values in \autoref{tab:ecological_stratified_errors} are the mean and population standard deviation of those image-tile metrics within each reference-target group. Consequently, the Overall mean RMSE is an average of image-tile RMSE values and differs from the pooled-pixel RMSE reported in the main regression figures.

\begin{table}[htbp]
\centering
\small
\setlength{\tabcolsep}{4pt}
\caption{FLORO error analysis stratified by reference image-tile mean biomass or canopy height. BioMassters uses the published 25th- and 75th-percentile thresholds, $68.56$ and $112.92$ t/px. Espeletia CHM uses the corresponding percentiles of its reference image-tile means, $0.228$ and $0.778$ m. RMSE and signed bias are calculated pixel-wise within each image tile and summarized across tiles as mean $\pm$ standard deviation. Positive bias denotes overestimation and negative bias denotes underestimation.}
\label{tab:ecological_stratified_errors}
\begin{tabular}{llrrr}
\toprule
Benchmark & Reference-target group & Image tiles & RMSE mean $\pm$ std. & Bias mean $\pm$ std. \\
\midrule
BioMassters & Low, $0$-$68.56$ t/px & 1,721 & $35.175 \pm 9.627$ t/px & $+4.739 \pm 7.390$ t/px \\
& Medium, $68.56$-$112.92$ t/px & 1,167 & $51.982 \pm 10.486$ t/px & $-4.425 \pm 9.212$ t/px \\
& High, $>112.92$ t/px & 113 & $75.240 \pm 16.498$ t/px & $-28.643 \pm 14.661$ t/px \\
& Overall & 3,001 & $43.219 \pm 14.551$ t/px & $-0.082 \pm 11.126$ t/px \\
\midrule
Espeletia CHM & Low, $<0.228$ m & 37 & $0.121 \pm 0.063$ m & $-0.012 \pm 0.048$ m \\
& Medium, $0.228$-$0.778$ m & 74 & $0.195 \pm 0.055$ m & $+0.002 \pm 0.090$ m \\
& High, $>0.778$ m & 37 & $0.321 \pm 0.121$ m & $-0.034 \pm 0.201$ m \\
& Overall & 148 & $0.208 \pm 0.106$ m & $-0.011 \pm 0.122$ m \\
\bottomrule
\end{tabular}
\end{table}

\newpage

\section{Detailed ablation results}
\label{app:ablations}

This section provides the numerical results underlying the compact ablation summary in \autoref{fig:ablations}. The experiments examine complementary aspects of FLORO, including pretraining-data composition, masking, input indicators, positional encoding, and selected downstream modalities, under otherwise matched protocols. Unless stated otherwise, each configuration corresponds to one pretrained model or one downstream run; the differences are therefore descriptive rather than estimates of statistical significance.

\subsection{Pretraining-dataset composition and CropTypeSS SAR interaction}

The pretraining-data experiment combines two controls (\autoref{tab:pretraining_composition_ablation}). First, four nested configurations expand from approximately 30k Sentinel-1/2 samples to Sentinel-1/2 plus SkySat (34{,}905 samples), then add UAV multispectral data (59{,}721 samples), and finally add UAV RGB data (81{,}282 samples). Second, a balanced multi-sensor subset of approximately 30k samples is compared with the Sentinel-1/2 subset at the same rounded total sample count. In constructing the balanced subset, the Sentinel-1/2 contribution is reduced from approximately 30k to approximately 8k samples and the remaining samples are drawn from the other source groups. The Sentinel-1/2 subset already includes all FLORO spectral groups together with elevation and SAR, while the other sources primarily increase diversity in sensors, platforms, spatial resolutions, radiometry, and source domains. Architecture, pretraining objective, masking curriculum, optimization, and downstream evaluation were held fixed.

The nested increments are heterogeneous. Adding SkySat reduced Burn Scars by 12.61 points and CropTypeSS by 5.35 points but improved Espeletia classification by 4.70 points. Adding UAV multispectral data then recovered 11.99 points on Burn Scars and increased BioMassters by 4.80 points, while adding UAV RGB improved every benchmark relative to the preceding 60k configuration, most strongly CropTypeSS (+4.14 points). At fixed size, the balanced subset improved all three scene-classification benchmarks, MADOS segmentation, and BioMassters regression. It reduced the other five segmentation benchmarks and Espeletia CHM regression. The largest changes were $+5.40$ points for Espeletia classification, $+2.40$ points for BioMassters, $-3.42$ points for CropTypeSS, and $-2.47$ points for Burn Scars (\autoref{fig:ablations}a).

The total-size-matched comparison controls the overall number of samples but remains confounded by source-specific sample volume because Sentinel-1/2 data decrease from approximately 30k to approximately 8k samples in the balanced subset. Performance decreases, particularly for downstream datasets that may benefit from Sentinel-1/2 pretraining, may therefore partly reflect reduced exposure to this source. The comparison also does not isolate sensor-source diversity from source identity, geographic coverage, spatial resolution, radiometry, or source-specific data quality, and it does not represent an increase in modality coverage because the Sentinel-1/2 subset already includes all FLORO spectral groups, elevation, and SAR. We therefore interpret the controls as showing that the balanced composition can benefit scene classification and selected regression settings, especially BioMassters, but does not provide a general benefit for segmentation. These experiments do not establish sensor-source diversity as the cause of FLORO's overall transfer performance.

\begin{table}[htb]
\centering
\caption{Effect of pretraining-dataset composition on downstream transfer. The first four rows form a nested expansion; the final row is a total-size-matched, balanced multi-sensor control. Although the two 30k subsets have approximately the same overall size, the balanced subset contains only approximately 8k Sentinel-1/2 samples. All models use the same FLORO architecture, masking curriculum, optimization setup, and downstream protocol. Segmentation is reported as mIoU, scene classification as accuracy, and regression as pixel-level $R^2$, all in percent. UAV-MS denotes UAV multispectral imagery; sample counts are rounded in the row labels.}
\label{tab:pretraining_composition_ablation}
\setlength{\tabcolsep}{3pt}
\resizebox{\textwidth}{!}{%
\begin{tabular}{lrrrrrrrrrrr}
\toprule
& \multicolumn{6}{c}{Segmentation mIoU} & \multicolumn{3}{c}{Classification accuracy} & \multicolumn{2}{c}{Regression $R^2$} \\
\cmidrule(lr){2-7}\cmidrule(lr){8-10}\cmidrule(lr){11-12}
Pretraining dataset & BurnSr & MADOS & PASTIS & Sen1Floods11 & CropTypeSS & Potsdam & EuroSAT-MS & Espeletia & Brick Kiln & Espeletia CHM & BioMassters \\
\midrule
Sentinel-1/2 (30k) & 82.684 & 53.145 & 34.878 & \textbf{87.620} & \textbf{55.457} & 74.764 & 88.3 & 67.6 & 95.9 & 91.1 & 54.6 \\
+ SkySat (35k) & 70.073 & 52.948 & 36.043 & 87.540 & 50.109 & 74.738 & 88.5 & 72.3 & 95.6 & 89.9 & 55.0 \\
+ UAV-MS (60k) & 82.061 & 54.750 & 35.646 & 87.488 & 49.418 & 75.121 & 90.3 & 73.6 & 96.6 & 90.2 & 59.8 \\
+ UAV-RGB (81k) & \textbf{82.820} & \textbf{55.717} & \textbf{36.299} & 87.567 & 53.558 & \textbf{75.595} & \textbf{90.8} & \textbf{75.0} & \textbf{97.0} & \textbf{91.2} & \textbf{60.0} \\
\midrule
Balanced multi-sensor (30k) & 80.216 & 53.428 & 33.968 & 86.904 & 52.044 & 74.414 & 89.3 & 73.0 & 96.2 & 89.4 & 57.0 \\
\bottomrule
\end{tabular}}
\end{table}

Because CropTypeSS uses both optical and SAR inputs, we also tested whether the downstream contribution of SAR depended on the pretraining dataset (\autoref{tab:croptypess_sar_interaction} and \autoref{fig:croptypess_sar_interaction}). With Sentinel-1/2 pretraining, removing SAR reduced mIoU from 55.457\% to 54.103\%, so the full input was 1.354 points better. With balanced 30k pretraining, removing SAR changed mIoU from 52.044\% to 52.171\%, a difference of only $-0.127$ points in the full-minus-reduced convention. The two lines should not be averaged into a single mechanism effect: they indicate that downstream SAR dependence changes with pretraining composition, while the Sentinel-1/2 configuration retains the higher absolute mIoU.

\begin{table}[htb]
\centering
\caption{Interaction between pretraining-dataset composition and downstream SAR input on CropTypeSS. Difference is full input minus SAR removed; positive values favor retaining SAR. All values are single-run mIoU percentages.}
\label{tab:croptypess_sar_interaction}
\begin{tabular}{lrrr}
\toprule
Pretraining dataset & Full input & SAR removed & Difference (pp) \\
\midrule
Sentinel-1/2 (30k) & 55.457 & 54.103 & $+1.354$ \\
Balanced multi-sensor (30k) & 52.044 & 52.171 & $-0.127$ \\
Mean & 53.751 & 53.137 & $+0.614$ \\
\bottomrule
\end{tabular}
\end{table}

\begin{figure}[H]
    \centering
    \includegraphics[width=0.78\linewidth]{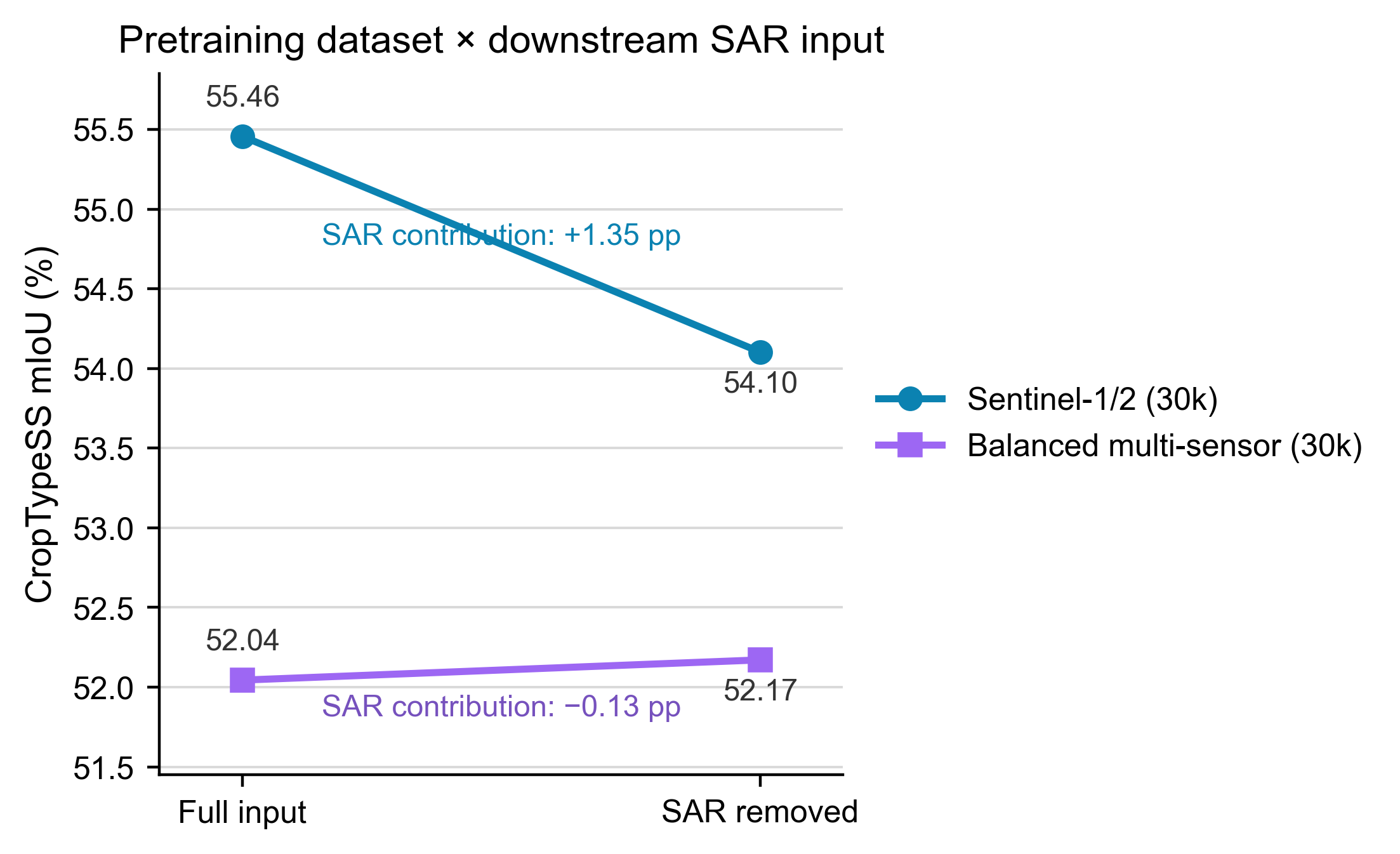}
    \caption{CropTypeSS interaction between pretraining-dataset composition and downstream SAR input. The Sentinel-1/2-pretrained model decreased from 55.457\% to 54.103\% mIoU when SAR was removed, whereas the balanced 30k model changed from 52.044\% to 52.171\%. The nonparallel responses show why a single average SAR-removal effect would obscure the dependence on pretraining composition.}
    \label{fig:croptypess_sar_interaction}
\end{figure}

\subsection{Pretraining masking-regime ablation}

To evaluate the masking curriculum, we compared the standard staged schedule ($25\% \rightarrow 50\% \rightarrow 75\%$) with an otherwise matched model pretrained using a fixed $75\%$ masking ratio throughout training (\autoref{tab:masking_regime_ablation}). Curriculum pretraining improved five of the six segmentation benchmarks and increased average segmentation mIoU from 63.922\% to 65.262\%. The largest gain occurred on CropTypeSS (6.10 percentage points), followed by BurnSr (1.19 points), PASTIS (1.03 points), Potsdam (0.60 points), and Sen1Floods11 (0.57 points); fixed $75\%$ masking improved MADOS by 1.46 points. For scene classification, curriculum pretraining improved EuroSAT-MS and Espeletia accuracy by 0.20 and 0.70 points, respectively, whereas fixed masking improved Brick Kiln accuracy by 1.10 points. Regression effects were also mixed: fixed masking improved Espeletia CHM $R^2$ by 1.20 points, whereas curriculum pretraining improved BioMassters $R^2$ by 1.20 points.

These results suggest that the staged schedule primarily benefits representations used for dense spatial prediction rather than providing uniform gains across tasks. CropTypeSS consists of $32 \times 32$ image tiles that must be resampled to FLORO's $256 \times 256$ input size; its 6.10-point gain is therefore consistent with greater robustness to severe input-scale mismatch. By contrast, the curriculum advantage on Potsdam's previously unseen airborne UltraCamXp domain was only 0.60 points. Because the ablation is based on one pretrained model per regime, small differences should not be interpreted as statistically significant.

\begin{table}[htb]
\centering
\caption{Effect of the pretraining masking regime on downstream transfer. The curriculum progressively increases the masking ratio from $25\%$ to $50\%$ and then $75\%$; the fixed regime uses $75\%$ masking throughout pretraining. Segmentation is reported as mIoU, scene classification as accuracy, and regression as pixel-level $R^2$, all in percent. Improvement is curriculum pretraining minus fixed $75\%$ masking, in percentage points; positive values favor the curriculum.}
\label{tab:masking_regime_ablation}
\setlength{\tabcolsep}{3pt}
\resizebox{\textwidth}{!}{%
\begin{tabular}{lrrrrrrrrrrr}
\toprule
& \multicolumn{6}{c}{Segmentation mIoU} & \multicolumn{3}{c}{Classification accuracy} & \multicolumn{2}{c}{Regression $R^2$} \\
\cmidrule(lr){2-7}\cmidrule(lr){8-10}\cmidrule(lr){11-12}
Pretraining regime & BurnSr & MADOS & PASTIS & Sen1Floods11 & CropTypeSS & Potsdam & EuroSAT-MS & Espeletia & Brick Kiln & Espeletia CHM & BioMassters \\
\midrule
Fixed 75\% masking & 81.63 & \textbf{57.18} & 35.27 & 87.00 & 47.46 & 74.99 & 90.60 & 74.30 & \textbf{98.10} & \textbf{92.40} & 58.80 \\
Curriculum pretraining & \textbf{82.82} & 55.72 & \textbf{36.30} & \textbf{87.57} & \textbf{53.56} & \textbf{75.60} & \textbf{90.80} & \textbf{75.00} & 97.00 & 91.20 & \textbf{60.00} \\
Curriculum improvement (pp) & $\mathbf{+1.19}$ & $-1.46$ & $\mathbf{+1.03}$ & $\mathbf{+0.57}$ & $\mathbf{+6.10}$ & $\mathbf{+0.60}$ & $\mathbf{+0.20}$ & $\mathbf{+0.70}$ & $-1.10$ & $-1.20$ & $\mathbf{+1.20}$ \\
\bottomrule
\end{tabular}}
\end{table}

\subsection{Availability- and validity-channel ablation}

To assess whether the availability and validity channels contribute to downstream transfer, we compared the standard FLORO input representation with two perturbed variants in which all availability and validity channels were set to zero or to one (\autoref{tab:availability_validity_ablation}). The effects on segmentation were modest and dataset-dependent. Relative to the correct setting, setting the channels to zero reduced Burn Scars and Potsdam mIoU by 0.84 and 1.51 percentage points, respectively; setting them to one reduced Burn Scars by 0.50 points but changed Potsdam by only $+0.02$ points. Sen1Floods11 was similarly insensitive. This dataset supplies all FLORO spectral groups together with both SAR channels (VV and VH), so elevation is the only absent input group. Its nearly complete and consistent input configuration likely reduces the information added by the availability channels, although the downstream decoder may also compensate for the perturbation.

The clearest channel effect occurred for Espeletia scene classification. Relative to the correct setting, setting all channels to zero reduced accuracy by 5.4 points and F1 by 7.6 points, while setting them to one reduced accuracy by 1.4 points and F1 by 3.1 points. The effect on Brick Kiln Classification was small. Espeletia CHM regression showed a different pattern: setting all channels to zero degraded performance, whereas setting them to one improved the regression metrics relative to the standard configuration. The channel contribution is therefore task- and modality-dependent rather than uniformly beneficial.

\begin{table}[htb]
\centering
\caption{Effect of availability and validity channels on downstream transfer. Segmentation and classification metrics are reported in percent; regression $R^2$, RMSE, and rRMSE follow their displayed units. The standard setting uses the correct availability and validity channels. Higher is better for mIoU, accuracy, F1, and $R^2$; lower is better for RMSE and rRMSE.}
\label{tab:availability_validity_ablation}
\resizebox{\linewidth}{!}{%
\begin{tabular}{lcccccccccc}
\toprule
Input setting & \multicolumn{3}{c}{Segmentation mIoU} & \multicolumn{4}{c}{Classification} & \multicolumn{3}{c}{Regression} \\
\cmidrule(lr){2-4}\cmidrule(lr){5-8}\cmidrule(lr){9-11}
& Burn Scars & Sen1Floods11 & Potsdam & \multicolumn{2}{c}{Espeletia} & \multicolumn{2}{c}{Brick Kiln} & \multicolumn{3}{c}{Espeletia CHM} \\
\cmidrule(lr){5-6}\cmidrule(lr){7-8}\cmidrule(lr){9-11}
& mIoU & mIoU & mIoU & Acc. & F1 & Acc. & F1 & $R^2$ & RMSE & rRMSE (\%) \\
\midrule
Correct channels & \textbf{82.820} & 87.30 & 75.60 & \textbf{75.00} & \textbf{69.50} & \textbf{97.00} & \textbf{96.60} & 0.91 & 0.26 & 42.80 \\
Channels set to 0 & 81.98 & \textbf{87.71} & 74.09 & 69.60 & 61.90 & 96.80 & 96.40 & 0.87 & 0.30 & 49.40 \\
Channels set to 1 & 82.32 & 86.78 & \textbf{75.62} & 73.60 & 66.40 & 96.50 & 96.00 & \textbf{0.93} & \textbf{0.22} & \textbf{35.90} \\
\bottomrule
\end{tabular}}
\end{table}

We hypothesized that the CHM behavior was linked to the role of DSM/elevation in canopy-height prediction. Removing DSM/elevation substantially reduced overall performance, from approximately $R^2=0.91$ to $R^2=0.760$ with the correct channels (\autoref{tab:chm_dsm_availability_ablation}). More importantly, the advantage of setting all channels to one disappeared: relative to the correct setting, $R^2$ decreased from 0.760 to 0.752 and rRMSE increased from 66.5\% to 67.4\%. The improvement observed with the full input was therefore not a generic consequence of forcing the channels to one; it was associated with the presence of DSM/elevation, which provides strong structural information for canopy-height prediction. Availability channels consequently interact with task-relevant modalities. For continuous structural regression, especially when elevation inputs are consistently available and spatially aligned, overly conservative validity masks may suppress useful information or alter the activation of multimodal encoder pathways.

\begin{table}[H]
\centering
\caption{Interaction between DSM/elevation input and availability- and validity-channel perturbation for Espeletia CHM regression. Higher is better for $R^2$; lower is better for RMSE and rRMSE.}
\label{tab:chm_dsm_availability_ablation}
\begin{tabular}{lccc}
\toprule
Input setting & $R^2$ & RMSE & rRMSE (\%) \\
\midrule
Full input, correct channels & 0.91 & 0.26 & 42.8 \\
Full input, channels set to 0 & 0.87 & 0.30 & 49.4 \\
Full input, channels set to 1 & \textbf{0.93} & \textbf{0.22} & \textbf{35.9} \\
\midrule
No DSM/elevation, correct channels & 0.760 & 0.408 & 66.5 \\
No DSM/elevation, channels set to 0 & 0.761 & 0.409 & 66.5 \\
No DSM/elevation, channels set to 1 & 0.752 & 0.414 & 67.4 \\
\bottomrule
\end{tabular}
\end{table}

\subsection{Geo-positional encoding ablation}

This experiment isolates explicit geographic grounding on the three downstream benchmarks for which geospatial metadata were retained (\autoref{tab:geoposition_ablation}). Geo-positional encoding improved EuroSAT-MS accuracy and F1 by 0.6 percentage points and Espeletia CHM pixel-level $R^2$ by 0.2 points. Potsdam mIoU instead decreased from 75.59\% to 74.67\% ($-0.92$ points). The effects are small and task-dependent, and the single-run comparisons do not establish statistical significance. The activation maps in \autoref{fig:cams} provide qualitative context for the EuroSAT-MS result.

\begin{table}[htb]
\centering
\caption{Effect of absolute and geo-positional encoding on scene classification, semantic segmentation, and regression. All values are reported in percent. Difference is geo-positional minus absolute encoding in percentage points; positive values favor geo-positional encoding.}
\label{tab:geoposition_ablation}
\begin{tabular}{llrrr}
\toprule
Benchmark & Metric & Absolute & Geo-positional & Difference (pp) \\
\midrule
\multirow{2}{8em}{EuroSAT-MS} & Accuracy & 91.9 & 92.5 & $+0.6$ \\
 & F1 & 91.5 & 92.1 & $+0.6$ \\
Potsdam & mIoU & 75.59 & 74.67 & $-0.92$ \\
Espeletia CHM & Pixel-level $R^2$ & 91.2 & 91.4 & $+0.2$ \\
\bottomrule
\end{tabular}
\end{table}

\subsection{Targeted downstream modality removals}

The targeted removal experiments test whether selected auxiliary inputs contribute at transfer time; they do not isolate whether including those modalities during pretraining was beneficial (\autoref{tab:targeted_modality_removal}). Removing SAR slightly improved Sen1Floods11 mIoU by 0.56 points, while RGB-only input reduced Brick Kiln accuracy by 0.30 points. In contrast, removing SAR from BioMassters reduced pixel-level $R^2$ by 5.0 points, and removing DSM from Espeletia CHM reduced it by approximately 15 points. Auxiliary-modal contributions are therefore task-dependent rather than uniformly positive.

\begin{table}[htb]
\centering
\caption{Targeted downstream modality-removal effects. Segmentation is reported as mIoU, classification as accuracy, and regression as pixel-level $R^2$, all in percent. Difference is reduced input minus full input; negative values favor retaining the removed modality.}
\label{tab:targeted_modality_removal}
\begin{tabular}{llrrr}
\toprule
Benchmark & Reduced-input test & Full input & Reduced input & Difference (pp) \\
\midrule
Sen1Floods11 & SAR removed & 87.62 & 88.18 & $+0.56$ \\
Brick Kiln & RGB only & 97.00 & 96.70 & $-0.30$ \\
BioMassters & SAR removed & 59.70 & 54.70 & $-5.00$ \\
Espeletia CHM & DSM removed & 91.20 & 76.00 & $-15.20$ \\
\bottomrule
\end{tabular}
\end{table}

The two ecological regression tasks also show that aggregation changes the apparent modality contribution (\autoref{tab:ecological_modality_removal}). Without SAR, BioMassters $R^2$ decreased from 0.597 to 0.547 at the pixel level and from 0.835 to 0.800 after image-tile aggregation. Without DSM, Espeletia CHM $R^2$ decreased from 0.912 to 0.760 at the pixel level but only from 0.941 to 0.938 after aggregation. SAR therefore provides a modest complementary signal for biomass estimation, whereas DSM is especially important for fine-scale canopy-height reconstruction. These are within-FLORO comparisons; the U-Net, ViT, and TerraMind-L results in \autoref{fig:biomassters_evaluation} and \autoref{fig:espeletia_evaluation} retain their default auxiliary inputs.

\begin{table}[htb]
\centering
\small
\setlength{\tabcolsep}{2.5pt}
\caption{Detailed BioMassters and Espeletia CHM auxiliary-modality removal results at the pixel and image-tile levels. $\mathrm{NRMSE}$ is normalized by $P_{95}-P_{5}$. Higher is better for $R^2$; lower is better for the error metrics.}
\label{tab:ecological_modality_removal}
\resizebox{\textwidth}{!}{%
\begin{tabular}{llrrrrrrrr}
\toprule
& & \multicolumn{4}{c}{Pixel level} & \multicolumn{4}{c}{Image-tile level} \\
\cmidrule(lr){3-6}\cmidrule(lr){7-10}
Benchmark & Input & $R^2$ & RMSE & rRMSE (\%) & NRMSE (\%) & $R^2$ & RMSE & rRMSE (\%) & NRMSE (\%) \\
\midrule
BioMassters & Multispectral + SAR & 0.597 & 45.612 t/px & 71.3 & 22.3 & 0.835 & 11.126 t/px & 17.4 & 13.1 \\
BioMassters & Multispectral only & 0.547 & 47.909 t/px & 76.1 & 23.8 & 0.800 & 11.635 t/px & 18.5 & 13.9 \\
\midrule
Espeletia CHM & Multispectral + DSM & 0.912 & 0.234 m & 40.9 & 10.3 & 0.941 & 0.123 m & 21.5 & 7.9 \\
Espeletia CHM & Multispectral only & 0.760 & 0.408 m & 66.5 & 16.8 & 0.938 & 0.145 m & 23.6 & 8.6 \\
\bottomrule
\end{tabular}}
\end{table}

\newpage
\section*{Declaration of generative AI and AI-assisted technologies in the manuscript preparation process}

During the preparation of this work the authors used Grammarly and OpenAI's ChatGPT for language editing, grammar checking, and improving the clarity and readability of the manuscript. After using these tools, the authors reviewed and edited the content as needed and take full responsibility for the content of the published article.

\bibliographystyle{elsarticle-num-names} 
\bibliography{references_floro}

@article{Besson2022TowardsCommunities,
    title = {{Towards the fully automated monitoring of ecological communities}},
    year = {2022},
    journal = {Ecology Letters},
    author = {Besson, Marc and Alison, Jamie and Bjerge, Kim and Gorochowski, Thomas E. and H{\o}ye, Toke T. and Jucker, Tommaso and Mann, Hjalte M.R. and Clements, Christopher F.},
    number = {12},
    month = {12},
    pages = {2753--2775},
    volume = {25},
    publisher = {John Wiley and Sons Inc},
    doi = {10.1111/ele.14123},
    issn = {14610248},
    pmid = {36264848}
}

@misc{Borowiec2022DeepEvolution,
    title = {{Deep learning as a tool for ecology and evolution}},
    year = {2022},
    booktitle = {Methods in Ecology and Evolution},
    author = {Borowiec, Marek L. and Dikow, Rebecca B. and Frandsen, Paul B. and McKeeken, Alexander and Valentini, Gabriele and White, Alexander E.},
    number = {8},
    month = {8},
    pages = {1640--1660},
    volume = {13},
    publisher = {British Ecological Society},
    doi = {10.1111/2041-210X.13901},
    issn = {2041210X}
}

@article{Cavender-Bares2022IntegratingConservation,
    title = {{Integrating remote sensing with ecology and evolution to advance biodiversity conservation}},
    year = {2022},
    journal = {Nature Ecology and Evolution},
    author = {Cavender-Bares, Jeannine and Schneider, Fabian D. and Santos, Maria João and Armstrong, Amanda and Carnaval, Ana and Dahlin, Kyla M. and Fatoyinbo, Lola and Hurtt, George C. and Schimel, David and Townsend, Philip A. and Ustin, Susan L. and Wang, Zhihui and Wilson, Adam M.},
    number = {5},
    month = {5},
    pages = {506--519},
    volume = {6},
    publisher = {Nature Research},
    doi = {10.1038/s41559-022-01702-5},
    issn = {2397334X},
    pmid = {35332280}
}

@misc{Ma2019DeepReview,
    title = {{Deep learning in remote sensing applications: A meta-analysis and review}},
    year = {2019},
    booktitle = {ISPRS Journal of Photogrammetry and Remote Sensing},
    author = {Ma, Lei and Liu, Yu and Zhang, Xueliang and Ye, Yuanxin and Yin, Gaofei and Johnson, Brian Alan},
    month = {6},
    pages = {166--177},
    volume = {152},
    publisher = {Elsevier B.V.},
    doi = {10.1016/j.isprsjprs.2019.04.015},
    issn = {09242716}
}

@article{Wu2023DeepLandscape,
    title = {{Deep learning enables satellite-based monitoring of large populations of terrestrial mammals across heterogeneous landscape}},
    year = {2023},
    journal = {Nature Communications},
    author = {Wu, Zijing and Zhang, Ce and Gu, Xiaowei and Duporge, Isla and Hughey, Lacey F. and Stabach, Jared A. and Skidmore, Andrew K. and Hopcraft, J. Grant C. and Lee, Stephen J. and Atkinson, Peter M. and McCauley, Douglas J. and Lamprey, Richard and Ngene, Shadrack and Wang, Tiejun},
    number = {1},
    month = {12},
    volume = {14},
    publisher = {Nature Research},
    doi = {10.1038/s41467-023-38901-y},
    issn = {20411723},
    pmid = {37244940}
}

@article{McCabe2017TheHydrology,
    title = {{The future of Earth observation in hydrology}},
    year = {2017},
    journal = {Hydrology and Earth System Sciences},
    author = {McCabe, Matthew F. and Rodell, Matthew and Alsdorf, Douglas E. and Miralles, Diego G. and Uijlenhoet, Remko and Wagner, Wolfgang and Lucieer, Arko and Houborg, Rasmus and Verhoest, Niko E.C. and Franz, Trenton E. and Shi, Jiancheng and Gao, Huilin and Wood, Eric F.},
    number = {7},
    pages = {3879--3914},
    volume = {21},
    publisher = {Copernicus GmbH},
    doi = {10.5194/hess-21-3879-2017},
    issn = {16077938}
}

@article{Zhu2026OnModels,
    title = {{On the foundations of Earth foundation models}},
    year = {2026},
    journal = {Communications Earth {\&} Environment},
    author = {Zhu, Xiao Xiang and Xiong, Zhitong and Wang, Yi and Stewart, Adam J. and Heidler, Konrad and Wang, Yuanyuan and Yuan, Zhenghang and Dujardin, Thomas and Xu, Qingsong and Shi, Yilei},
    month = {1},
    publisher = {Springer Science and Business Media LLC},
    doi = {10.1038/s43247-025-03127-x},
    issn = {26624435}
}

@article{Cong2023SatMAE:Imagery,
    title = {{SatMAE: Pre-training Transformers for Temporal and Multi-Spectral Satellite Imagery}},
    year = {2023},
    journal = {arXiv preprint arXiv:2207.08051},
    author = {Cong, Yezhen and Khanna, Samar and Meng, Chenlin and Liu, Patrick and Rozi, Erik and He, Yutong and Burke, Marshall and Lobell, David B. and Ermon, Stefano},
    month = {1},
    url = {http://arxiv.org/abs/2207.08051},
    doi = {https://doi.org/10.48550/arXiv.2207.08051},
    arxivId = {2207.08051}
}

@article{Bommasani2021OnModels,
    title = {{On the Opportunities and Risks of Foundation Models}},
    year = {2021},
    journal = {arXiv preprint arXiv:2108.07258},
    author = {Bommasani, Rishi and Hudson, Drew A. and Adeli, Ehsan and Altman, Russ and Arora, Simran and von Arx, Sydney and Bernstein, Michael S. and Bohg, Jeannette and Bosselut, Antoine and Brunskill, Emma and Brynjolfsson, Erik and Buch, Shyamal and Card, Dallas and Castellon, Rodrigo and Chatterji, Niladri and Chen, Annie and Creel, Kathleen and Davis, Jared Quincy and Demszky, Dora and Donahue, Chris and Doumbouya, Moussa and Durmus, Esin and Ermon, Stefano and Etchemendy, John and Ethayarajh, Kawin and Fei-Fei, Li and Finn, Chelsea and Gale, Trevor and Gillespie, Lauren and Goel, Karan and Goodman, Noah and Grossman, Shelby and Guha, Neel and Hashimoto, Tatsunori and Henderson, Peter and Hewitt, John and Ho, Daniel E. and Hong, Jenny and Hsu, Kyle and Huang, Jing and Icard, Thomas and Jain, Saahil and Jurafsky, Dan and Kalluri, Pratyusha and Karamcheti, Siddharth and Keeling, Geoff and Khani, Fereshte and Khattab, Omar and Koh, Pang Wei and Krass, Mark and Krishna, Ranjay and Kuditipudi, Rohith and Kumar, Ananya and Ladhak, Faisal and Lee, Mina and Lee, Tony and Leskovec, Jure and Levent, Isabelle and Li, Xiang Lisa and Li, Xuechen and Ma, Tengyu and Malik, Ali and Manning, Christopher D. and Mirchandani, Suvir and Mitchell, Eric and Munyikwa, Zanele and Nair, Suraj and Narayan, Avanika and Narayanan, Deepak and Newman, Ben and Nie, Allen and Niebles, Juan Carlos and Nilforoshan, Hamed and Nyarko, Julian and Ogut, Giray and Orr, Laurel and Papadimitriou, Isabel and Park, Joon Sung and Piech, Chris and Portelance, Eva and Potts, Christopher and Raghunathan, Aditi and Reich, Rob and Ren, Hongyu and Rong, Frieda and Roohani, Yusuf and Ruiz, Camilo and Ryan, Jack and R{\'{e}}, Christopher and Sadigh, Dorsa and Sagawa, Shiori and Santhanam, Keshav and Shih, Andy and Srinivasan, Krishnan and Tamkin, Alex and Taori, Rohan and Thomas, Armin W. and Tram{\`{e}}r, Florian and Wang, Rose E. and Wang, William and Wu, Bohan and Wu, Jiajun and Wu, Yuhuai and Xie, Sang Michael and Yasunaga, Michihiro and You, Jiaxuan and Zaharia, Matei and Zhang, Michael and Zhang, Tianyi and Zhang, Xikun and Zhang, Yuhui and Zheng, Lucia and Zhou, Kaitlyn and Liang, Percy},
    month = {8},
    url = {http://arxiv.org/abs/2108.07258},
    arxivId = {2108.07258}
}

@inproceedings{Mall2024RemoteAlignment,
    title = {{Remote Sensing Vision-Language Foundation Models Without Annotations via Ground Remote Alignment}},
    year = {2024},
    booktitle = {ICLR 2024},
    author = {Mall, Utkarsh and Phoo, Cheng Perng and Liu, Meilin Kelsey and Vondrick, Carl and Hariharan, Bharath and Bala, Kavita},
    month = {5},
    pages = {1--13},
    url = {https://graft.cs.cornell.edu},
    address = {Vienna}
}

@article{Wang2024MTP:Pretraining,
    title = {{MTP: Advancing Remote Sensing Foundation Model via Multitask Pretraining}},
    year = {2024},
    journal = {IEEE Journal of Selected Topics in Applied Earth Observations and Remote Sensing},
    author = {Wang, Di and Zhang, Jing and Xu, Minqiang and Liu, Lin and Wang, Dongsheng and Gao, Erzhong and Han, Chengxi and Guo, Haonan and Du, Bo and Tao, Dacheng and Zhang, Liangpei},
    pages = {11632--11654},
    volume = {17},
    publisher = {Institute of Electrical and Electronics Engineers Inc.},
    doi = {10.1109/JSTARS.2024.3408154},
    issn = {21511535},
    arxivId = {2403.13430}
}

@article{Cha2024AImages,
    title = {{A Billion-scale Foundation Model for Remote Sensing Images}},
    year = {2024},
    journal = {IEEE Journal of Selected Topics in Applied Earth Observations and Remote Sensing},
    author = {Cha, Keumgang and Seo, Junghoon and Lee, Taekyung},
    url = {http://dx.doi.org/10.1109/JSTARS.2024.3401772},
    doi = {10.1109/JSTARS.2024.3401772}
}

@misc{Morera2024FoundationEcology,
    title = {{Foundation models in shaping the future of ecology}},
    year = {2024},
    booktitle = {Ecological Informatics},
    author = {Morera, Albert},
    month = {5},
    volume = {80},
    publisher = {Elsevier B.V.},
    doi = {10.1016/j.ecoinf.2024.102545},
    issn = {15749541}
}

@article{Mai2024OnPaper,
    title = {{On the Opportunities and Challenges of Foundation Models for GeoAI (Vision Paper)}},
    year = {2024},
    journal = {ACM Transactions on Spatial Algorithms and Systems},
    author = {Mai, Gengchen and Huang, Weiming and Sun, Jin and Song, Suhang and Mishra, Deepak and Liu, Ninghao and Gao, Song and Liu, Tianming and Cong, Gao and Hu, Yingjie and Cundy, Chris and Li, Ziyuan and Zhu, Rui and Lao, Ni},
    number = {2},
    month = {6},
    pages = {1--46},
    volume = {10},
    url = {https://dl.acm.org/doi/10.1145/3653070},
    doi = {10.1145/3653070},
    issn = {2374-0353}
}

@article{McCabe2008HydrologicalStudies,
    title = {{Hydrological consistency using multi-sensor remote sensing data for water and energy cycle studies}},
    year = {2008},
    journal = {Remote Sensing of Environment},
    author = {McCabe, M. F. and Wood, E. F. and W{\'{o}}jcik, R. and Pan, M. and Sheffield, J. and Gao, H. and Su, H.},
    number = {2},
    month = {2},
    pages = {430--444},
    volume = {112},
    doi = {10.1016/j.rse.2007.03.027},
    issn = {00344257}
}

@inproceedings{Benediktsson2007MultipleDevelopments,
    title = {{Multiple Classifier Systems in Remote Sensing: From Basics to Recent Developments}},
    year = {2007},
    booktitle = {Multiple Classifier Systems},
    author = {Benediktsson, Jon Atliand and Chanussot, Jocelyn and Fauvel, Mathieu},
    editor = {{Haindl Michaland KittlerJosefand Roli Fabio}},
    pages = {501--512},
    publisher = {Springer Berlin Heidelberg},
    address = {Berlin, Heidelberg},
    isbn = {978-3-540-72523-7}
}

@article{DallaMura2015ChallengesSensing,
    title = {{Challenges and Opportunities of Multimodality and Data Fusion in Remote Sensing}},
    year = {2015},
    journal = {Proceedings of the IEEE},
    author = {Dalla Mura, M. and Prasad, S. and Pacifici, F. and Gamba, P. and Chanussot, J. and Benediktsson, J. A.},
    number = {9},
    month = {9},
    pages = {1585--1601},
    volume = {103},
    publisher = {Institute of Electrical and Electronics Engineers Inc.},
    url = {https://ieeexplore.ieee.org/document/7194740/},
    doi = {10.1109/JPROC.2015.2462751},
    issn = {0018-9219}
}

@article{Lopez2017EvaluatingData,
    title = {{Evaluating the hydrological consistency of evaporation products using satellite-based gravity and rainfall data}},
    year = {2017},
    journal = {Hydrology and Earth System Sciences},
    author = {L{\'{o}}pez, Oliver and Houborg, Rasmus and McCabe, Matthew Francis},
    number = {1},
    month = {1},
    pages = {323--343},
    volume = {21},
    publisher = {Copernicus GmbH},
    doi = {10.5194/hess-21-323-2017},
    issn = {16077938}
}

@article{Sturari2017IntegratingMapping,
    title = {{Integrating elevation data and multispectral high-resolution images for an improved hybrid Land Use/Land cover mapping}},
    year = {2017},
    journal = {European Journal of Remote Sensing},
    author = {Sturari, Mirco and Frontoni, Emanuele and Pierdicca, Roberto and Mancini, Adriano and Malinverni, Eva Savina and Tassetti, Anna Nora and Zingaretti, Primo},
    number = {1},
    volume = {50},
    publisher = {Associazione Italiana di Telerilevamento},
    doi = {10.1080/22797254.2017.1274572},
    issn = {22797254}
}

@article{Wang2023DecouplingLearning,
    title = {{Decoupling Common and Unique Representations for Multimodal Self-supervised Learning}},
    year = {2023},
    journal = {arXiv preprint arXiv:2309.05300},
    author = {Wang, Yi and Albrecht, Conrad M and Braham, Nassim Ait Ali and Liu, Chenying and Xiong, Zhitong and Zhu, Xiao Xiang},
    month = {9},
    url = {http://arxiv.org/abs/2309.05300},
    arxivId = {2309.05300}
}

@article{Harston1992ImprovedData,
    title = {{Improved Image Classification With Neural Networks by Fusing Multispectral Signatures With Topographical Data}},
    year = {1992},
    journal = {Telematics and Informatics},
    author = {Harston, Craig and Schumacher, Chris},
    pages = {157--162},
    volume = {9}
}

@article{Khanal2018IntegrationYield,
    title = {{Integration of high resolution remotely sensed data and machine learning techniques for spatial prediction of soil properties and corn yield}},
    year = {2018},
    journal = {Computers and Electronics in Agriculture},
    author = {Khanal, Sami and Fulton, John and Klopfenstein, Andrew and Douridas, Nathan and Shearer, Scott},
    month = {10},
    pages = {213--225},
    volume = {153},
    publisher = {Elsevier B.V.},
    doi = {10.1016/j.compag.2018.07.016},
    issn = {01681699}
}

@article{Brell20193DExtraction,
    title = {{3D hyperspectral point cloud generation: Fusing airborne laser scanning and hyperspectral imaging sensors for improved object-based information extraction}},
    year = {2019},
    journal = {ISPRS Journal of Photogrammetry and Remote Sensing},
    author = {Brell, Maximilian and Segl, Karl and Guanter, Luis and Bookhagen, Bodo},
    month = {3},
    pages = {200--214},
    volume = {149},
    publisher = {Elsevier B.V.},
    doi = {10.1016/j.isprsjprs.2019.01.022},
    issn = {09242716}
}

@article{Li2022DeepReview,
    title = {{Deep learning in multimodal remote sensing data fusion: A comprehensive review}},
    year = {2022},
    journal = {International Journal of Applied Earth Observation and Geoinformation},
    author = {Li, Jiaxin and Hong, Danfeng and Gao, Lianru and Yao, Jing and Zheng, Ke and Zhang, Bing and Chanussot, Jocelyn},
    month = {8},
    volume = {112},
    publisher = {Elsevier B.V.},
    doi = {10.1016/j.jag.2022.102926},
    issn = {1872826X},
    arxivId = {2205.01380}
}

@article{Hong2026FoundationMultimodality,
    title = {{Foundation Models in Remote Sensing: Evolving from Unimodality to Multimodality}},
    year = {2026},
    journal = {arXiv preprint arXiv:2603.00988},
    author = {Hong, Danfeng and Li, Chenyu and Li, Xuyang and Camps-Valls, Gustau and Chanussot, Jocelyn},
    month = {3},
    url = {http://arxiv.org/abs/2603.00988 http://dx.doi.org/10.1109/MGRS.2026.3669086},
    doi = {10.1109/MGRS.2026.3669086},
    arxivId = {2603.00988}
}

@article{Lu2024AISurvey,
    title = {{AI Foundation Models in Remote Sensing: A Survey}},
    year = {2024},
    journal = {arXiv preprint arXiv:2408.03464},
    author = {Lu, Siqi and Guo, Junlin and Zimmer-Dauphinee, James R and Nieusma, Jordan M and Wang, Xiao and VanValkenburgh, Parker and Wernke, Steven A and Huo, Yuankai},
    month = {8},
    url = {http://arxiv.org/abs/2408.03464},
    arxivId = {2408.03464}
}

@article{Jakubik2023FoundationIntelligence,
    title = {{Foundation Models for Generalist Geospatial Artificial Intelligence}},
    year = {2023},
    journal = {arXiv preprint arXiv:2310.18660},
    author = {Jakubik, Johannes and Roy, Sujit and Phillips, C. E. and Fraccaro, Paolo and Godwin, Denys and Zadrozny, Bianca and Szwarcman, Daniela and Gomes, Carlos and Nyirjesy, Gabby and Edwards, Blair and Kimura, Daiki and Simumba, Naomi and Chu, Linsong and Mukkavilli, S. Karthik and Lambhate, Devyani and Das, Kamal and Bangalore, Ranjini and Oliveira, Dario and Muszynski, Michal and Ankur, Kumar and Ramasubramanian, Muthukumaran and Gurung, Iksha and Khallaghi, Sam and {Hanxi} and {Li} and Cecil, Michael and Ahmadi, Maryam and Kordi, Fatemeh and Alemohammad, Hamed and Maskey, Manil and Ganti, Raghu and Weldemariam, Kommy and Ramachandran, Rahul},
    month = {10},
    url = {http://arxiv.org/abs/2310.18660},
    arxivId = {2310.18660}
}

@article{Xiong2025NeuralObservation,
    title = {{Neural Plasticity-Inspired Multimodal Foundation Model for Earth Observation}},
    year = {2025},
    journal = {arXiv preprint arXiv:2403.15356},
    author = {Xiong, Zhitong and Wang, Yi and Zhang, Fahong and Stewart, Adam J. and Hanna, Joëlle and Borth, Damian and Papoutsis, Ioannis and Saux, Bertrand Le and Camps-Valls, Gustau and Zhu, Xiao Xiang},
    month = {10},
    url = {http://arxiv.org/abs/2403.15356},
    arxivId = {2403.15356}
}

@article{Jakubik2025TerraMind:Observation,
    title = {{TerraMind: Large-Scale Generative Multimodality for Earth Observation}},
    year = {2025},
    journal = {arXiv preprint arXiv:2504.11171},
    author = {Jakubik, Johannes and Yang, Felix and Blumenstiel, Benedikt and Scheurer, Erik and Sedona, Rocco and Maurogiovanni, Stefano and Bosmans, Jente and Dionelis, Nikolaos and Marsocci, Valerio and Kopp, Niklas and Ramachandran, Rahul and Fraccaro, Paolo and Brunschwiler, Thomas and Cavallaro, Gabriele and Bernabe-Moreno, Juan and Long{\'{e}}p{\'{e}}, Nicolas},
    month = {9},
    url = {http://arxiv.org/abs/2504.11171},
    arxivId = {2504.11171}
}

@misc{EuropeanSpaceAgency2026Sentinel-1Documentation,
    title = {{Sentinel-1 Documentation}},
    year = {2026},
    author = {{European Space Agency}},
    publisher = {Copernicus Data Space Ecosystem Documentation},
    url = {https://documentation.dataspace.copernicus.eu/Data/SentinelMissions/Sentinel1.html}
}

@misc{EuropeanSpaceAgency2021CopernicusProduct,
    title = {{Copernicus Sentinel-2 Collection 1 MSI Level-2A Product}},
    year = {2021},
    author = {{European Space Agency}},
    publisher = {Copernicus Sentinel Online},
    url = {https://sentinels.copernicus.eu/sentinel-data-access/sentinel-products/sentinel-2-data-products/collection-1-level-2a}
}

@techreport{Collison2025PlanetPaper,
    title = {{Planet Surface Reflectance Technical White Paper}},
    year = {2025},
    author = {Collison, Alan and Curdoglo, Mariana},
    url = {https://assets.planet.com/marketing/PDF/Planet_Surface_Reflectance_Technical_White_Paper.pdf},
    institution = {Planet Labs PBC}
}

@misc{IGN2026RGETerrain,
    title = {{RGE ALTI Mod{\`{e}}le Num{\'{e}}rique de Terrain}},
    year = {2026},
    booktitle = {https://cartes.gouv.fr/rechercher-une-donnee/dataset/IGNF{\_}RGE-ALTI?redirected{\_}from=geoservices.ign.fr{\#}telechargement1m},
    author = {{IGN}},
    month = {4},
    url = {https://cartes.gouv.fr/rechercher-une-donnee/dataset/IGNF_RGE-ALTI?redirected_from=geoservices.ign.fr#telechargement1m}
}

@inproceedings{He2022MaskedLearners,
    title = {{Masked Autoencoders Are Scalable Vision Learners}},
    year = {2022},
    booktitle = {Proceedings of the IEEE Computer Society Conference on Computer Vision and Pattern Recognition},
    author = {He, Kaiming and Chen, Xinlei and Xie, Saining and Li, Yanghao and Dollar, Piotr and Girshick, Ross},
    pages = {15979--15988},
    volume = {2022-June},
    publisher = {IEEE Computer Society},
    isbn = {9781665469463},
    doi = {10.1109/CVPR52688.2022.01553},
    issn = {10636919},
    arxivId = {2111.06377}
}

@article{Bachmann2022MultiMAE:Target,
    title = {{MultiMAE: Multi-modal Multi-task Masked Autoencoders}},
    year = {2022},
    journal = {arXiv preprint arXiv:2204.01678},
    author = {Bachmann, Roman and Mizrahi, David and Atanov, Andrei and Zamir, Amir},
    url = {http://arxiv.org/abs/2204.01678},
    arxivId = {2204.01678}
}

@article{Dosovitskiy2020AnScale,
    title = {{An Image is Worth 16x16 Words: Transformers for Image Recognition at Scale}},
    year = {2020},
    journal = {arXiv preprint arXiv:2010.11929},
    author = {Dosovitskiy, Alexey and Beyer, Lucas and Kolesnikov, Alexander and Weissenborn, Dirk and Zhai, Xiaohua and Unterthiner, Thomas and Dehghani, Mostafa and Minderer, Matthias and Heigold, Georg and Gelly, Sylvain and Uszkoreit, Jakob and Houlsby, Neil},
    month = {10},
    url = {http://arxiv.org/abs/2010.11929},
    arxivId = {2010.11929}
}

@article{Vaswani2017AttentionNeed,
    title = {{Attention Is All You Need}},
    year = {2017},
    journal = {arXiv preprint arXiv:1706.03762},
    author = {Vaswani, Ashish and Shazeer, Noam and Parmar, Niki and Uszkoreit, Jakob and Jones, Llion and Gomez, Aidan N. and Kaiser, Lukasz and Polosukhin, Illia},
    month = {6},
    url = {http://arxiv.org/abs/1706.03762},
    arxivId = {1706.03762}
}

@article{Loshchilov2017DecoupledRegularization,
    title = {{Decoupled Weight Decay Regularization}},
    year = {2017},
    journal = {arXiv preprint arXiv:1711.05101},
    author = {Loshchilov, Ilya and Hutter, Frank},
    month = {11},
    url = {http://arxiv.org/abs/1711.05101},
    arxivId = {1711.05101}
}

@article{Micikevicius2017MixedTraining,
    title = {{Mixed Precision Training}},
    year = {2017},
    journal = {arXiv preprint arXiv:1710.03740},
    author = {Micikevicius, Paulius and Narang, Sharan and Alben, Jonah and Diamos, Gregory and Elsen, Erich and Garcia, David and Ginsburg, Boris and Houston, Michael and Kuchaiev, Oleksii and Venkatesh, Ganesh and Wu, Hao},
    month = {10},
    url = {http://arxiv.org/abs/1710.03740},
    arxivId = {1710.03740}
}

@article{Goyal2017AccurateHour,
    title = {{Accurate, Large Minibatch SGD: Training ImageNet in 1 Hour}},
    year = {2017},
    journal = {arXiv preprint arXiv:1706.02677},
    author = {Goyal, Priya and Doll{\'{a}}r, Piotr and Girshick, Ross and Noordhuis, Pieter and Wesolowski, Lukasz and Kyrola, Aapo and Tulloch, Andrew and Jia, Yangqing and He, Kaiming},
    month = {6},
    url = {http://arxiv.org/abs/1706.02677},
    arxivId = {1706.02677}
}

@article{Marsocci2026PANGAEA:Benchmark,
    title = {{PANGAEA: Assessing Geospatial Foundation Models Capabilities through a Global and Inclusive Benchmark}},
    year = {2026},
    journal = {IEEE Geoscience and Remote Sensing Magazine},
    author = {Marsocci, Valerio and Jia, Yuru and Bellier, Georges Le and Kerekes, David and Zeng, Liang and Hafner, Sebastian and Gerard, Sebastian and Brune, Eric and Yadav, Ritu and Shibli, Ali and Fang, Heng and Ban, Yifang and Vergauwen, Maarten and Audebert, Nicolas and Nascetti, Andrea},
    number = {1},
    pages = {245--285},
    volume = {14},
    publisher = {Institute of Electrical and Electronics Engineers Inc.},
    doi = {10.1109/MGRS.2025.3628194},
    issn = {21686831}
}

@article{Fuller2023CROMA:Autoencoders,
    title = {{CROMA: Remote Sensing Representations with Contrastive Radar-Optical Masked Autoencoders}},
    year = {2023},
    journal = {arXiv preprint arXiv:2311.00566},
    author = {Fuller, Anthony and Millard, Koreen and Green, James R.},
    month = {11},
    url = {http://arxiv.org/abs/2311.00566},
    arxivId = {2311.00566}
}

@article{Mendieta2023TowardsPretraining,
    title = {{Towards Geospatial Foundation Models via Continual Pretraining}},
    year = {2023},
    journal = {arXiv preprint arXiv:2302.04476},
    author = {Mendieta, Matias and Han, Boran and Shi, Xingjian and Zhu, Yi and Chen, Chen},
    month = {8},
    url = {http://arxiv.org/abs/2302.04476},
    arxivId = {2302.04476}
}

@article{Bastani2023SatlasPretrain:Understanding,
    title = {{SatlasPretrain: A Large-Scale Dataset for Remote Sensing Image Understanding}},
    year = {2023},
    journal = {arXiv preprint arXiv:2211.15660},
    author = {Bastani, Favyen and Wolters, Piper and Gupta, Ritwik and Ferdinando, Joe and Kembhavi, Aniruddha},
    month = {8},
    url = {http://arxiv.org/abs/2211.15660},
    arxivId = {2211.15660}
}

@article{Hong2024SpectralGPT:Model,
    title = {{SpectralGPT: Spectral Remote Sensing Foundation Model}},
    year = {2024},
    journal = {IEEE Transactions on Pattern Analysis and Machine Intelligence},
    author = {Hong, Danfeng and Zhang, Bing and Li, Xuyang and Li, Yuxuan and Li, Chenyu and Yao, Jing and Yokoya, Naoto and Li, Hao and Ghamisi, Pedram and Jia, Xiuping and Plaza, Antonio and Gamba, Paolo and Benediktsson, Jon Atli and Chanussot, Jocelyn},
    number = {8},
    pages = {5227--5244},
    volume = {46},
    publisher = {IEEE Computer Society},
    doi = {10.1109/TPAMI.2024.3362475},
    issn = {19393539},
    pmid = {38568772},
    arxivId = {2311.07113}
}

@article{Reed2022Scale-MAE:Learning,
    title = {{Scale-MAE: A Scale-Aware Masked Autoencoder for Multiscale Geospatial Representation Learning}},
    year = {2022},
    journal = {arXiv preprint arXiv:2212.14532},
    author = {Reed, Colorado J. and Gupta, Ritwik and Li, Shufan and Brockman, Sarah and Funk, Christopher and Clipp, Brian and Keutzer, Kurt and Candido, Salvatore and Uyttendaele, Matt and Darrell, Trevor},
    month = {12},
    url = {http://arxiv.org/abs/2212.14532},
    arxivId = {2212.14532}
}

@article{Liu2024RemoteCLIP:Sensing,
    title = {{RemoteCLIP: A Vision Language Foundation Model for Remote Sensing}},
    year = {2024},
    journal = {arXiv preprint arXiv:2306.11029},
    author = {Liu, Fan and Chen, Delong and Guan, Zhangqingyun and Zhou, Xiaocong and Zhu, Jiale and Ye, Qiaolin and Fu, Liyong and Zhou, Jun},
    month = {4},
    url = {http://arxiv.org/abs/2306.11029},
    arxivId = {2306.11029}
}

@article{Wang2023SSL4EO-S12:Observation,
    title = {{SSL4EO-S12: A Large-Scale Multi-Modal, Multi-Temporal Dataset for Self-Supervised Learning in Earth Observation}},
    year = {2023},
    journal = {arXiv preprint arXiv:2211.07044},
    author = {Wang, Yi and Braham, Nassim Ait Ali and Xiong, Zhitong and Liu, Chenying and Albrecht, Conrad M and Zhu, Xiao Xiang},
    month = {5},
    url = {http://arxiv.org/abs/2211.07044},
    arxivId = {2211.07044}
}

@incollection{Ronneberger2015U-Net:Segmentation,
    title = {{U-Net: Convolutional Networks for Biomedical Image Segmentation}},
    year = {2015},
    author = {Ronneberger, Olaf and Fischer, Philipp and Brox, Thomas},
    publisher = {MICCAI 2015},
    booktitle = {Lecture Notes in Computer Science},
    month = {5},
    pages = {234--241},
    url = {http://link.springer.com/10.1007/978-3-319-24574-4_28},
    doi = {10.1007/978-3-319-24574-4{\_}28}
}

@article{Xiao2018UnifiedUnderstanding,
    title = {{Unified Perceptual Parsing for Scene Understanding}},
    year = {2018},
    journal = {arXiv preprint arXiv:1807.10221},
    author = {Xiao, Tete and Liu, Yingcheng and Zhou, Bolei and Jiang, Yuning and Sun, Jian},
    month = {7},
    url = {https://arxiv.org/abs/1807.10221},
    arxivId = {1807.10221}
}

@inproceedings{Garnot2020LightweightTimeseries,
    title = {{Lightweight temporal self-attention for classifying satellite images time series}},
    year = {2020},
    booktitle = {Lecture Notes in Computer Science (including subseries Lecture Notes in Artificial Intelligence and Lecture Notes in Bioinformatics)},
    author = {Garnot, Vivien Sainte Fare and Landrieu, Loic},
    pages = {171--181},
    volume = {12588 LNAI},
    publisher = {Springer Science and Business Media Deutschland GmbH},
    isbn = {9783030657413},
    doi = {10.1007/978-3-030-65742-0{\_}12},
    issn = {16113349},
    arxivId = {2007.00586}
}

@misc{Phillips2023HLSDataset,
    title = {{HLS Foundation Burnscars Dataset}},
    year = {2023},
    author = {Phillips, Christopher and Roy, Sujit and Ankur, Kumar and Ramachandran, Rahul},
    month = {8},
    url = {https://huggingface.co/ibm-nasa-geospatial/hls_burn_scars}
}

@article{Kikaki2024DetectingImagery,
    title = {{Detecting Marine pollutants and Sea Surface features with Deep learning in Sentinel-2 imagery}},
    year = {2024},
    journal = {ISPRS Journal of Photogrammetry and Remote Sensing},
    author = {Kikaki, Katerina and Kakogeorgiou, Ioannis and Hoteit, Ibrahim and Karantzalos, Konstantinos},
    month = {4},
    pages = {39--54},
    volume = {210},
    publisher = {Elsevier B.V.},
    doi = {10.1016/j.isprsjprs.2024.02.017},
    issn = {09242716}
}

@inproceedings{Garnot2021PanopticNetworks,
    title = {{Panoptic Segmentation of Satellite Image Time Series with Convolutional Temporal Attention Networks}},
    year = {2021},
    booktitle = {Proceedings of the IEEE International Conference on Computer Vision},
    author = {Garnot, Vivien Sainte Fare and Landrieu, Loic},
    pages = {4852--4861},
    publisher = {Institute of Electrical and Electronics Engineers Inc.},
    isbn = {9781665428125},
    doi = {10.1109/ICCV48922.2021.00483},
    issn = {15505499},
    arxivId = {2107.07933}
}

@article{SainteFareGarnot2022Multi-modalSeries,
    title = {{Multi-modal temporal attention models for crop mapping from satellite time series}},
    year = {2022},
    journal = {ISPRS Journal of Photogrammetry and Remote Sensing},
    author = {Sainte Fare Garnot, Vivien and Landrieu, Loic and Chehata, Nesrine},
    month = {5},
    pages = {294--305},
    volume = {187},
    publisher = {Elsevier B.V.},
    doi = {10.1016/j.isprsjprs.2022.03.012},
    issn = {09242716},
    arxivId = {2112.07558}
}

@inproceedings{Bonafilia2020Sen1Floods11:Sentinel-1b,
    title = {{Sen1Floods11: A georeferenced dataset to train and test deep learning flood algorithms for sentinel-1}},
    year = {2020},
    booktitle = {IEEE Computer Society Conference on Computer Vision and Pattern Recognition Workshops},
    author = {Bonafilia, Derrick and Tellman, Beth and Anderson, Tyler and Issenberg, Erica},
    month = {6},
    pages = {835--845},
    volume = {2020-June},
    publisher = {IEEE Computer Society},
    isbn = {9781728193601},
    doi = {10.1109/CVPRW50498.2020.00113},
    issn = {21607516}
}

@InProceedings{Rustowicz2019SemanticMethods,
    author = {M Rustowicz, Rose and Cheong, Robin and Wang, Lijing and Ermon, Stefano and Burke, Marshall and Lobell, David},
    title = {Semantic Segmentation of Crop Type in Africa: A Novel Dataset and Analysis of Deep Learning Methods},
    booktitle = {Proceedings of the IEEE/CVF Conference on Computer Vision and Pattern Recognition (CVPR) Workshops},
    month = {June},
    pages = {75--82},
    year = {2019}
}

@misc{ISPRSISPRSDataset,
    title = {{ISPRS Potsdam Dataset}},
    booktitle = {International Society for Photogrammetry and Remote Sensing},
    author = {{ISPRS}},
    year = {2018},
    publisher = {International Society for Photogrammetry and Remote Sensing},
    url = {https://www.isprs.org/education/benchmarks/UrbanSemLab/2d-sem-label-potsdam.aspx}
}

@article{Helber2019EuroSAT:Classification,
    title = {{EuroSAT: A Novel Dataset and Deep Learning Benchmark for Land Use and Land Cover Classification}},
    year = {2019},
    journal = {arXiv preprint arXiv:1709.00029},
    author = {Helber, Patrick and Bischke, Benjamin and Dengel, Andreas and Borth, Damian},
    month = {2},
    url = {http://arxiv.org/abs/1709.00029},
    arxivId = {1709.00029}
}

@inproceedings{Zhou2016LearningLocalization,
    title = {{Learning Deep Features for Discriminative Localization}},
    year = {2016},
    booktitle = {Proceedings of the IEEE Computer Society Conference on Computer Vision and Pattern Recognition},
    author = {Zhou, Bolei and Khosla, Aditya and Lapedriza, Agata and Oliva, Aude and Torralba, Antonio},
    month = {12},
    pages = {2921--2929},
    volume = {2016-December},
    publisher = {IEEE Computer Society},
    isbn = {9781467388504},
    doi = {10.1109/CVPR.2016.319},
    issn = {10636919},
    arxivId = {1512.04150}
}

@inproceedings{Nascetti2023BioMassters:Series,
    title = {{BioMassters: A Benchmark Dataset for Forest Biomass Estimation using Multi-modal Satellite Time Series}},
    year = {2023},
    booktitle = {Advances in Neural Information Processing Systems},
    author = {Nascetti, Andrea and Yadav, Ritu and Brodt, Kiril and Qu, Qixun and Fan, Hongwei and Shendryk, Yuri and Shah, Isha and Chung, Christine},
    editor = {Oh, A. and Naumann, T. and Globerson, A. and Saenko, K. and Hardt, M. and Levine, S.},
    pages = {20409--20420},
    publisher = {Curran Associates, Inc.},
    url = {https://proceedings.neurips.cc/paper_files/paper/2023/file/40daf2a00278c4bea1b26cd4c8a654f8-Paper-Datasets_and_Benchmarks.pdf},
    doi = {10.57967/hf/1009}
}

@misc{Rodriguez2026EspeletiaParamo,
    title = {{Espeletia Scene Classification and Canopy Height Dataset from UAV Multispectral Imagery in a Colombian P{\'{a}}ramo}},
    year = {2026},
    author = {Rodr{\'{i}}guez, Jorge L. and Angulo-Morales, Victor},
    publisher = {Zenodo},
    url = {https://zenodo.org/records/19205335},
    doi = {https://doi.org/10.5281/zenodo.19205335}
}

@misc{Koldasbayeva2024ChallengesPractice,
    title = {{Challenges in data-driven geospatial modeling for environmental research and practice}},
    year = {2024},
    booktitle = {Nature Communications },
    author = {Koldasbayeva, Diana and Tregubova, Polina and Gasanov, Mikhail and Zaytsev, Alexey and Petrovskaia, Anna and Burnaev, Evgeny},
    number = {1},
    month = {12},
    volume = {15},
    publisher = {Nature Research},
    doi = {10.1038/s41467-024-55240-8},
    issn = {20411723},
    pmid = {39702456},
    arxivId = {2311.11057}
}

@misc{Yu2025MachineAnalysis,
    title = {{Machine learning for ecological analysis}},
    year = {2025},
    booktitle = {Chemical Engineering Journal},
    author = {Yu, Zhengyang and Bu, Chunfeng and Li, Yanjie},
    month = {3},
    volume = {507},
    publisher = {Elsevier B.V.},
    doi = {10.1016/j.cej.2025.160780},
    issn = {13858947}
}

@incollection{Schrodt2020IntegratingScales,
    title = {{Integrating biodiversity, remote sensing, and auxiliary information for the study of ecosystem functioning and conservation at large spatial scales}},
    year = {2020},
    booktitle = {Remote Sensing of Plant Biodiversity},
    author = {Schrodt, Franziska and De La Barreda Bautista, Betsabe and Williams, Christopher and Boyd, Doreen S. and Schaepman-Strub, Gabriela and Santos, Maria J.},
    month = {1},
    pages = {449--484},
    publisher = {Springer International Publishing},
    isbn = {9783030331573},
    doi = {10.1007/978-3-030-33157-3{\_}17}
}

@misc{Samadzadegan2025AProspects,
    title = {{A critical review on multi-sensor and multi-platform remote sensing data fusion approaches: current status and prospects}},
    year = {2025},
    booktitle = {International Journal of Remote Sensing},
    author = {Samadzadegan, Farhad and Toosi, Ahmad and Dadrass Javan, Farzaneh},
    number = {3},
    pages = {1327--1402},
    volume = {46},
    publisher = {Taylor and Francis Ltd.},
    doi = {10.1080/01431161.2024.2429784},
    issn = {13665901}
}

@article{Ploton2020SpatialModels,
    title = {{Spatial validation reveals poor predictive performance of large-scale ecological mapping models}},
    year = {2020},
    journal = {Nature Communications},
    author = {Ploton, Pierre and Mortier, Frédéric and R{\'{e}}jou-M{\'{e}}chain, Maxime and Barbier, Nicolas and Picard, Nicolas and Rossi, Vivien and Dormann, Carsten and Cornu, Guillaume and Viennois, Gaëlle and Bayol, Nicolas and Lyapustin, Alexei and Gourlet-Fleury, Sylvie and P{\'{e}}lissier, Raphaël},
    number = {1},
    month = {12},
    volume = {11},
    publisher = {Nature Research},
    doi = {10.1038/s41467-020-18321-y},
    issn = {20411723},
    pmid = {32917875}
}

@article{Wolff2025InterannualCommunities,
    title = {{Interannual spectral consistency and spatial uncertainties in UAV-based detection of boreal and subarctic mire plant communities}},
    year = {2025},
    journal = {Remote Sensing in Ecology and Conservation},
    author = {Wolff, Franziska and Kolari, Tiina H.M. and R{\"{a}}s{\"{a}}nen, Aleksi and Tahvanainen, Teemu and Korpelainen, Pasi and Villoslada, Miguel and Verdonen, Mariana and Lotsari, Eliisa and Pang, Yuwen and Kumpula, Timo},
    number = {6},
    month = {12},
    pages = {719--739},
    volume = {11},
    publisher = {John Wiley and Sons Inc},
    doi = {10.1002/rse2.70017},
    issn = {20563485}
}

@article{Marti-Escofet2025Fine-tuneModels,
    title = {{Fine-tune Smarter, Not Harder: Parameter-Efficient Fine-Tuning for Geospatial Foundation Models}},
    year = {2025},
    journal = {arXiv preprint arXiv:2504.17397},
    author = {Marti-Escofet, Francesc and Blumenstiel, Benedikt and Scheibenreif, Linus and Fraccaro, Paolo and Schindler, Konrad},
    month = {6},
    url = {http://arxiv.org/abs/2504.17397},
    arxivId = {2504.17397}
}

@article{Chen2026LightFormer:Segmentation,
    title = {{LightFormer: A lightweight and efficient decoder for remote sensing image segmentation}},
    year = {2026},
    journal = {arXiv preprint arXiv:2504.10834},
    author = {Chen, Sihang and Yun, Lijun and Liu, Ze and Zhu, JianFeng and Chen, Jie and Wang, Hui and Nie, Yueping},
    month = {1},
    url = {http://arxiv.org/abs/2504.10834},
    arxivId = {2504.10834}
}

@article{Gorelick2017GoogleEveryone,
    title = {{Google Earth Engine: Planetary-scale geospatial analysis for everyone}},
    year = {2017},
    journal = {Remote Sensing of Environment},
    author = {Gorelick, Noel and Hancher, Matt and Dixon, Mike and Ilyushchenko, Simon and Thau, David and Moore, Rebecca},
    month = {12},
    pages = {18--27},
    volume = {202},
    doi = {10.1016/j.rse.2017.06.031},
    issn = {00344257}
}

@article{Vandeghen2026TrackMAE:Predict,
    title = {{TrackMAE: Video Representation Learning via Track Mask and Predict}},
    year = {2026},
    journal = {arXiv preprint arXiv:2603.27268},
    author = {Vandeghen, Renaud and Thoker, Fida Mohammad and Van Droogenbroeck, Marc and Ghanem, Bernard},
    month = {3},
    url = {http://arxiv.org/abs/2603.27268},
    arxivId = {2603.27268}
}

@article{Huang2023MaskedListen,
    title = {{Masked Autoencoders that Listen}},
    year = {2023},
    journal = {arXiv preprint arXiv:2207.06405},
    author = {Huang, Po-Yao and Xu, Hu and Li, Juncheng and Baevski, Alexei and Auli, Michael and Galuba, Wojciech and Metze, Florian and Feichtenhofer, Christoph},
    month = {1},
    url = {http://arxiv.org/abs/2207.06405},
    arxivId = {2207.06405}
}

@article{Gao2025SFA-Net:Matching,
    title = {{SFA-Net: A SAM-guided focused attention network for multimodal remote sensing image matching}},
    year = {2025},
    journal = {ISPRS Journal of Photogrammetry and Remote Sensing},
    author = {Gao, Tian and Lan, Chaozhen and Huang, Wenjun and Wang, Sheng},
    month = {5},
    pages = {188--206},
    volume = {223},
    publisher = {Elsevier B.V.},
    doi = {10.1016/j.isprsjprs.2025.02.032},
    issn = {09242716}
}

@article{Lee2021ScalableCapacity,
    title = {{Scalable deep learning to identify brick kilns and aid regulatory capacity}},
    year = {2021},
    journal = {Proceedings of the National Academy of Sciences},
    author = {Lee, Jihyeon and Brooks, Nina R and Tajwar, Fahim and Burke, Marshall and Ermon, Stefano and Lobell, David B and Biswas, Debashish and Luby, Stephen P},
    number = {17},
    volume = {118},
    url = {https://www.pnas.org/doi/abs/10.1073/pnas.2018863118},
    isbn = {2018863118},
    doi = {10.1073/pnas.2018863118}
}

@incollection{Cavender-Bares2020ApplyingScience,
    title = {{Applying remote sensing to biodiversity science}},
    year = {2020},
    booktitle = {Remote Sensing of Plant Biodiversity},
    author = {Cavender-Bares, Jeannine and Schweiger, Anna K. and Pinto-Ledezma, Jesús N. and Meireles, Jose Eduardo},
    month = {1},
    pages = {13--42},
    publisher = {Springer International Publishing},
    isbn = {9783030331573},
    doi = {10.1007/978-3-030-33157-3{\_}2}
}

@article{vanderLeeuw2024UsingModels,
    title = {{Using Fine Resolution Remotely Sensed Data-Derived Land Cover to Inform Dryland State and Transition Models}},
    year = {2024},
    journal = {Rangeland Ecology and Management},
    author = {van der Leeuw, Elisabeth and van Leeuwen, Willem J.D. and Marsh, Stuart E. and Archer, Steven R.},
    month = {9},
    publisher = {Elsevier Inc.},
    doi = {10.1016/j.rama.2024.06.003},
    issn = {15507424}
}

@article{Mai2023Sphere2Vec:Predictions,
    title = {{Sphere2Vec: A General-Purpose Location Representation Learning over a Spherical Surface for Large-Scale Geospatial Predictions}},
    journal = {arXiv preprint arXiv:2306.17624},
    year = {2023},
    author = {Mai, Gengchen and Xuan, Yao and Zuo, Wenyun and He, Yutong and Song, Jiaming and Ermon, Stefano and Janowicz, Krzysztof and Lao, Ni},
    month = {7},
    url = {http://arxiv.org/abs/2306.17624},
    arxivId = {2306.17624}
}

@article{Ruwurm2024GeographicNetworks,
    title = {{Geographic Location Encoding with Spherical Harmonics and Sinusoidal Representation Networks}},
    journal = {arXiv preprint arXiv:2310.06743},
    year = {2024},
    author = {Ru{\ss}wurm, Marc and Klemmer, Konstantin and Rolf, Esther and Zbinden, Robin and Tuia, Devis},
    month = {4},
    url = {http://arxiv.org/abs/2310.06743},
    arxivId = {2310.06743}
}

@article{Tseng2024LightweightTimeseries,
    title = {{Lightweight, Pre-trained Transformers for Remote Sensing Timeseries}},
    journal = {arXiv preprint arXiv:2304.14065},
    year = {2024},
    author = {Tseng, Gabriel and Cartuyvels, Ruben and Zvonkov, Ivan and Purohit, Mirali and Rolnick, David and Kerner, Hannah},
    month = {2},
    url = {http://arxiv.org/abs/2304.14065},
    arxivId = {2304.14065}
}

@article{Madan2024CL-MAE:Autoencoders,
    title = {{CL-MAE: Curriculum-Learned Masked Autoencoders}},
    year = {2024},
    journal = {arXiv preprint arXiv:2308.16572},
    author = {Madan, Neelu and Ristea, Nicolae-Catalin and Nasrollahi, Kamal and Moeslund, Thomas B. and Ionescu, Radu Tudor},
    month = {2},
    url = {http://arxiv.org/abs/2308.16572},
    arxivId = {2308.16572}
}

@article{Yoon2025CurriMAE:Classification,
    title = {{CurriMAE: curriculum learning based masked autoencoders for multi-labeled pediatric thoracic disease classification}},
    year = {2025},
    journal = {PeerJ Computer Science},
    author = {Yoon, Taeyoung and Kang, Daesung},
    doi = {http://dx.doi.org/10.7717/peerj-cs.3019}
}

@inproceedings{Astruc2025AnySat:Modalities,
    title = {{AnySat: One Earth Observation Model for Many Resolutions, Scales, and Modalities}},
    year = {2025},
    booktitle = {Proceedings of the IEEE/CVF Conference on Computer Vision and Pattern Recognition (CVPR)},
    author = {Astruc, Guillaume and Gonthier, Nicolas and Mallet, Clément and Landrieu, Loic},
    month = {6},
    pages = {19530--19540},
    url = {https://github.com/gastruc/AnySat.}
}

@article{Zhou2026RemoteMethod,
    title = {{Remote sensing meta modal representation for missing modality land cover mapping: From EarthMiss dataset to MetaRS method}},
    year = {2026},
    journal = {Remote Sensing of Environment},
    author = {Zhou, Yiheng and Ma, Ailong and Wang, Junjue and Chen, Zihang and Zhong, Yanfei},
    month = {1},
    volume = {333},
    publisher = {Elsevier Inc.},
    doi = {10.1016/j.rse.2025.115132},
    issn = {00344257}
}

@article{Wu2025AObservation,
    title = {{A semantic-enhanced multi-modal remote sensing foundation model for Earth observation}},
    year = {2025},
    journal = {Nature Machine Intelligence},
    author = {Wu, Kang and Zhang, Yingying and Ru, Lixiang and Dang, Bo and Lao, Jiangwei and Yu, Lei and Luo, Junwei and Zhu, Zifan and Sun, Yue and Zhang, Jiahao and Zhu, Qi and Wang, Jian and Yang, Ming and Chen, Jingdong and Zhang, Yongjun and Li, Yansheng},
    number = {8},
    month = {8},
    pages = {1235--1249},
    volume = {7},
    publisher = {Nature Research},
    doi = {10.1038/s42256-025-01078-8},
    issn = {25225839}
}

@article{Ji2026SpecAware:Mapping,
    title = {{SpecAware: a spectral-content aware foundation model for unifying multi-sensor learning in hyperspectral remote sensing mapping}},
    year = {2026},
    journal = {ISPRS Journal of Photogrammetry and Remote Sensing},
    author = {Ji, Renjie and Wang, Xue and Niu, Chao and Zhang, Wen and Mei, Yong and Tan, Kun},
    month = {4},
    pages = {242--260},
    volume = {234},
    publisher = {Elsevier B.V.},
    doi = {10.1016/j.isprsjprs.2026.02.024},
    issn = {09242716}
}

@inproceedings{Li2026Any-Optical-Model:Sensing,
    title = {{Any-Optical-Model: A Universal Foundation Model for Optical Remote Sensing}},
    year = {2026},
    booktitle = {Proceedings of the AAAI Conference on Artificial Intelligence},
    author = {Li, Xuyang and Li, Chenyu and Danfeng, Hong},
    pages = {6539--6547},
    url = {www.aaai.org},
    doi = {https://doi.org/10.1609/aaai.v40i8.37583}
}

@article{Dias2026EmergingModels,
    title = {{Emerging Flexible Designs for Geospatial Multimodal Foundation Models}},
    journal = {arXiv preprint arXiv:2606.12595},
    year = {2026},
    author = {Dias, Philipe and Abebe, Waqwoya and Potnis, Abhishek and Tsaris, Aristeidis and Lu, Dan and Wang, Xiao and Lunga, Dalton},
    month = {6},
    url = {http://arxiv.org/abs/2606.12595},
    arxivId = {2606.12595}
}

\end{document}